\documentclass{article}
\usepackage{amsmath, amsthm, amssymb}
\usepackage{graphicx}
\usepackage{verbatim}
\usepackage{natbib}
\usepackage{caption}
\usepackage{subcaption}
\usepackage{fancyvrb}
\usepackage{enumerate}
\usepackage{relsize}
\usepackage{multirow}
\usepackage[section]{placeins}

\usepackage{hyperref}
\usepackage[margin=1.4in]{geometry}
\hypersetup{colorlinks,citecolor=blue,urlcolor=blue,linkcolor=blue}
\usepackage[ruled,vlined]{algorithm2e}

\usepackage{floatrow}
\newfloatcommand{capbtabbox}{table}[][\FBwidth]

\usepackage{stefan_tex}
\graphicspath{{./figures/}}
\numberwithin{equation}{section}

\theoremstyle{plain}
\newtheorem{prop}{Proposition}

\newtheorem{coro}[prop]{Corollary}
\newtheorem{lemm}[prop]{Lemma}
\newtheorem{theo}[prop]{Theorem}

\theoremstyle{definition}
\newtheorem{exam}{Example}
\newtheorem{defi}[exam]{Definition}
\newtheorem{assumption}{Assumption}

\theoremstyle{remark}

\newcommand\blfootnote[1]{%
  \begingroup
  \renewcommand\thefootnote{}\footnote{#1}%
  \addtocounter{footnote}{-1}%
  \endgroup
}

\author{Roshni Sahoo\\
  \texttt{rsahoo@stanford.edu}
  \and
  Stefan Wager\\
  \texttt{swager@stanford.edu}
 }
\date{Stanford University \\}
\title{Policy Learning with Competing Agents}

\begin{document}

\maketitle

\begin{abstract}
    Decision\blfootnote{\hspace{-5.3mm}Draft version: \ifcase\month\or January\or February\or March\or April\or May\or June\or July\or August\or September\or October\or November\or December\fi \ \number \year. RS is supported by a NSF GRFP under grant DGE-1656518, Stanford Data Science Fellowship, and by NSF grant SES-2242876. Code available at \url{https://github.com/roshni714/policy-learning-competing-agents}.} makers often aim to learn a treatment assignment policy under a capacity constraint on the number of agents that they can treat. When agents can respond strategically to such policies, competition arises, complicating estimation of the optimal policy. In this paper, we study capacity-constrained treatment assignment in the presence of such interference. We consider a dynamic model where the decision maker allocates treatments at each time step and heterogeneous agents myopically best respond to the previous treatment assignment policy. When the number of agents is large but finite, we show that the threshold for receiving treatment under a given policy converges to the policy's mean-field equilibrium threshold. Based on this result, we develop a consistent estimator for the policy gradient. In a semi-synthetic experiment with data from the National Education Longitudinal Study of 1988, we demonstrate that this estimator can be used for learning capacity-constrained policies in the presence of strategic behavior.
\end{abstract}

\begin{section}{Introduction}
Decision makers often aim to learn policies for assigning treatments to human agents under capacity constraints \citep{athey2021policy, bhattacharya2012inferring, kitagawa2018should,  manski2004statistical}. These policies map an agent's observed characteristics to a treatment assignment. For example, \citet{bhattacharya2012inferring} learn a policy for allocating anti-malaria bed net subsidies to households in Kenya, where only a fraction of the population can receive the subsidy. To enforce the capacity constraint, they use a selection criterion, such as a machine learning model, to score households and assign the treatment to households that score above a threshold, given by a quantile of the score distribution. When the decision maker has a capacity constraint, an agent's treatment assignment depends on how their score ranks relative to others.

In policy learning, practitioners often assume that the data used for treatment choice is exogenous to the treatment assignment policy. Such an assumption is plausible if the treatment assignment policy is unknown to human agents or knowledge of the policy is unlikely to affect the agents' observed characteristics. For example, in the social program studied by \citet{bhattacharya2012inferring}, characteristics such as household wealth, household size, and whether or not the household has a child under the age of ten, are used to determine allocations; these attributes are unlikely to change even if households have knowledge of how the bed net subsidies are allocated. 

However, in many other applications, human agents may change their observed characteristics in response to the policy, violating the exogeneity assumption. For example, in college admissions, applicants, with knowledge that high test scores will improve their chances of admission, may enroll in test preparation services to improve their scores \citep{bound2009playing}. In job hiring, candidates may join intensive bootcamps to improve their career prospects \citep{thayer2017barriers}. A growing literature focuses on policy learning in the presence of strategic behavior \citep{bjorkegren2020manipulation, frankel2019improving, munro2020learning}, but these works implicitly assume that the decision maker does not have a capacity constraint, which means an agent's treatment assignment only depends on their own strategic behavior, unaffected by the behavior of others in the population. In contrast, when agents are strategic and treatment assignment is capacity-constrained, an agent's treatment assignment depends on the behavior of others and competition for the treatment arises. 

In this work, we study the problem of capacity-constrained treatment assignment in the presence of strategic behavior. We frame the problem in a dynamic setting, where the decision maker assigns treatments at each time step. Suppose a decision maker deploys a fixed selection criterion for all time. At time step $t$, agents report their covariates with knowledge of the policy from time step $t-1$, where the policy depends on the fixed criterion and the threshold for receiving treatment at time step $t-1$. Then, the decision maker assigns treatments to agents who score above the appropriate quantile of the score distribution at time step $t$, satisfying the capacity constraint. As a result, the threshold for receiving treatment at time step $t$ depends on agents' strategic behavior. At an equilibrium induced by the criterion, the threshold for receiving treatment is fixed over time.  The goal of the decision maker, and the main goal of this work, is to find a selection criterion that obtains high equilibrium policy value, which is the policy value obtained at the equilibrium induced by the selection criterion. 

The goal of maximizing the equilibrium policy value is motivated by prior works that estimate policy effects or treatment effects at equilibrium \citep{heckman1998general, munro2021treatment, wager2021experimenting}. \citet{heckman1998general} estimate the effect of a tuition subsidy program on college enrollment by accounting for the program's impact on the equilibrium college skill price. \citet{munro2021treatment} estimate the effect of a binary intervention in a marketplace setting by accounting for the impact of the intervention on the resulting supply-demand equilibrium. \citet{wager2021experimenting} estimate the effect of supply-side payments on a platform's utility in equilibrium. \citet{johari2022experimental} use a structural model of a marketplace and its associated mean field limit to analyze how marketplace interference affects the performance of different experimental designs and estimators.

In Section \ref{sec:model}, we outline a model for capacity-constrained treatment assignment in the presence of strategic behavior. We assume that agents are myopic, so the covariates they report to the decision maker at time step $t$ depend only on the state of the system in time step $t-1$. Also, as in the aggregative games literature \citep{acemoglu2010robust, acemoglu2015robust, corchon1994comparative}, we assume that agents respond to an aggregate of other agents' actions. In particular, at time step $t$, agents will react to the threshold for receiving treatment from time step $t-1$, which is an aggregate of agents' strategic behavior from time step $t-1$. Finally, as in, e.g., \cite{frankel2019improving, frankel2019muddled}, we assume that agents are heterogenous in their raw covariates (covariates prior to modification) and in their ability to deviate from their raw covariates in their reported covariates.

In Section \ref{sec:results_equilibrium}, we give conditions on our model that guarantee existence and uniqueness of equilibria in the mean-field regime, the limiting regime where at each time step, an infinite number of agents are considered for the treatment. Furthermore, we show that under additional conditions, the mean-field equilibrium arises via fixed-point iteration. In Section \ref{sec:finite_results}, we translate these results to the finite regime, where a finite number of agents, sampled i.i.d. at each time step, are considered for treatment. We show that as the number of agents grows large, the system converges to the equilibrium of the mean-field model in a stochastic version of fixed-point iteration.

In Section \ref{sec:results_learning}, we aim to learn the selection criterion that maximizes the equilibrium policy value. Adapting the approach of \citet{wager2021experimenting}, we estimate the gradient of equilibrium policy value via a local experimentation scheme and apply gradient-based optimization to optimize the equilibrium policy value. In Section \ref{sec:exp}, we demonstrate that this approach can be used to learn policies in presence of competition in a semi-synthetic experiment with data from the National Education Longitudinal Study of 1988 \citep{ingels1994national}.

\begin{subsection}{Related Work}
The problem of learning optimal treatment assignment policies has received attention in econometrics, statistics, and computer science \citep{athey2021policy, bhattacharya2012inferring, kallus2021minimax, kitagawa2018should, manski2004statistical}. Most related to our work, \cite{bhattacharya2012inferring} study optimal capacity-constrained treatment assignment, where the decision maker can only allocate treatments to $1-q$ proportion of the population for $q \in (0, 1)$. They show that the welfare-maximizing assignment policy is a threshold rule on the agents' scores, where agents who score above $q$-th quantile of the score distribution are allocated treatment. Our work differs from \cite{bhattacharya2012inferring} because we do not assume that the distribution of observed covariates is exogenous to the treatment assignment policy.
 
Recent references that consider learning in the presence of strategic behavior include 
\citet{bjorkegren2020manipulation}, \citet{frankel2019improving} and \citet{munro2020learning}.
\citet{bjorkegren2020manipulation} propose a structural model for manipulation and use data from a field experiment to estimate the optimal policy. \citet{frankel2019improving} demonstrate that optimal predictors that account for strategic behavior will underweight manipulable data. \citet{munro2020learning} studies the optimal unconstrained assignment of binary-valued treatments in the presence of strategic behavior, without parametric assumptions on agent behavior. The main difference between our work and these previous works is that we account for the equilibrium effect of strategic behavior that arises from competition.

Our work is also related to strategic classification \citep{ahmadi2022classification, bruckner2012static, chen2020learning, Dalvi04adversarialclassification, dong2018strategic, hardt2016strategic, jagadeesan2021alternative, kleinberg2020classifiers, levanon2022generalized} and performative prediction \citep{miller2021outside, perdomo2020performative}. These works model the interaction between a predictor and its environment (strategic agents), and develop methods that are robust to the distribution shift induced by the predictor. However, a key distinction between our work and these references is that we optimize decisions by explicitly considering utility from treatment assignment with strategic agents,
rather than optimizing a notion of classification or predictive accuracy.
Furthermore, the classification and prediction setups implicitly assume that an agent's label does not depend on the behavior of others in the population, limiting the applicability of these methods to settings with spillovers induced by capacity constraints.

To the best of our knowledge, \cite{liu2021strategic} is the only existing work that studies capacity-constrained allocation in the presence of strategic behavior. \cite{liu2021strategic} introduces the problem of strategic ranking, where agents' rewards depend on their ranks after investing effort in modifying their covariates. They consider a setting where agents are heterogenous in their raw covariates but homogenous in their ability to modify their covariates. Under these assumptions, the authors find that agents' post-effort ranking preserves their original ranking by raw covariates. However, the assumption of homogeneity in ability to modify covariates may not be credible in some applications; for example, in the context of college admissions, students with high socioeconomic status may be more readily able to improve their test scores by investing in tutoring services than students with low socioeconomic status. Our work differs from \cite{liu2021strategic} because we allow agents to be heterogenous in both their raw covariates and ability to modify their reported covariates, which fundamentally alters the nature of the resulting policy learning problem. 
 
The problem of estimating the effect of an intervention in a marketplace setting is also relevant to our work. Marketplace interventions can impact the resulting supply-demand equilibrium, introducing interference and complicating estimation of the intervention's effect \citep{blake2014marketplace, heckman1998general, johari2022experimental}. To estimate an intervention's effect without disturbing the market equilibrium, \cite{ munro2021treatment, wager2021experimenting} propose a local experimentation scheme, motivated by mean-field modeling. Methodologically, we adapt their mean-field modeling and estimation strategies to estimate the effect of a policy at its equilibrium threshold.

Finally, our model draws on many concepts from game theory. Our assumption that agents are myopic, or will take decisions based on information from short time horizons, is a standard heuristic used in many previous works \citep{cournot2020researches, kandori1993learning, monderer1996fictitious}. Our assumption that agents account for the behavior of others through an aggregate quantity of their actions is a paradigm borrowed from aggregative games \citep{acemoglu2010robust, acemoglu2015robust, corchon1994comparative}. Most related to our work, \citet{acemoglu2015robust} consider a dynamic setting where the market aggregate at time step $t$ is an aggregate function of all the agents' best responses from time step $t$, and an agent's best response at time step $t$ is selected from a constraint set determined by the market aggregate from time step $t-1$. Analogously, in our work, the ``market aggregate" is the threshold for receiving treatment. The threshold for receiving treatment is a particular quantile of the agents' score, so we can view it as a function of agents' reported covariates (agents' best responses). Furthermore, the covariates that agents report in time step $t$ depend on the value of the market aggregate, or the threshold for receiving treatment, in time step $t-1$.
\end{subsection}

\begin{subsection}{Motivating Example}
We consider college admissions as a running example. The agents are a population of students who are hetergeneous in their baseline test scores and grades and in their ability to invest effort to change them. The decision maker is a college with a fixed selection criterion for scoring students based on their test scores and grades and a capacity constraint that they can only accept $1-q$ proportion of the applicants where $q \in (0, 1)$. Each year $t$, the college scores students and accepts students who rank above the $q$-th quantile of the score distribution. In year $t$, students have knowledge of the criterion and the acceptance threshold from the previous year $t-1$. Some students may invest effort to improve their chances of admission by enrolling in test preparation services or taking more advanced classes to improve their chances of getting accepted \citep{bound2009playing,rosinger2021role}. Finally, students report their post-effort test scores and grades to the college. To ensure the capacity constraint is satisfied, the college sets the acceptance threshold in year $t$ to the $q$-th quantile of the score distribution in year $t$. The acceptance threshold may oscillate from year to year until an equilibrium arises and the acceptance threshold is fixed over time.

This example aims to capture the phenomenon that college admissions has become increasingly competitive since the 1980s; \citet{bound2009playing} demonstrates that between 1986-2003, the 75th percentile math SAT score of accepted students at the top 20 public universities, top 20 private colleges, and top 20 liberal arts colleges, steadily trended upward. In our model, the equilibrium acceptance threshold depends on students' strategic behavior and the decision maker's capacity constraint and selection criterion. Different selection criteria may induce different equilibrium thresholds. The equilibrium acceptance threshold affects the value of the policy because it determines which students are accepted or rejected. A selection criterion that induces a high acceptance threshold may not necessarily yield high value for the college.
\end{subsection}

\end{section}

\begin{section}{Model}
\label{sec:model}
We define a dynamic model for capacity-constrained treatment assignment in the presence of strategic behavior
and define the equilibrium policy value. We propose a model for agent behavior in terms
of myopic utility maximization and provide conditions under which the resulting best response functions vary smoothly in problem
parameters.

\subsection{Dynamic Model}
Let $q \in (0, 1).$ A decision maker allocates treatments to $1-q$ proportion of a population of agents $i=1, 2, \dots$ at each time step $t \in \{1, 2, \dots\}.$ Agents' observed covariates are denoted $X_{i}^{t} \in \mathbb{R}^{d}$. At time $t$, the decision maker observes covariates $X_{i}^{t}$ and assigns treatments $W_{i}^{t} \in \{0, 1\}$ using a policy $\pi: \mathbb{R}^{d} \rightarrow \{0, 1\}$. In our motivating example, the decision maker is the college and the agents are students applying to the college. The covariates are students' applications, consisting of test scores and grades, that they submit to the college. The treatment represents admission to the college, which is desirable for the students.

We consider linear threshold rules with coefficients $\beta \in \mathcal{B} = \mathbb{S}^{d-1}$ and a threshold $S^{t} \in \mathbb{R}$, i.e. $\pi(X; \beta, S^{t}) = \mathbb{I}(\beta^{T}X > S^{t}).$ The decision maker fixes the selection criterion $\beta$ for all time steps $t$, while the threshold $S^{t}$ varies with $t$ to ensure that the capacity constraint is satisfied at each time step. In time step $t$, agents will respond strategically to the coefficients $\beta$ and the threshold from the previous time step $t-1,$ which is $S^{t-1}$,
i.e., $X_{i}^{t} = X_{i}(\beta, S^{t-1})$. The function $X_{i}$ may be stochastic. In the context of college admissions, students, with knowledge of the selection criterion and previous acceptance threshold, can invest effort to change their baseline test scores and grades, to improve their chances of being admitted. Section \ref{sec:agent} provides additional structure for agent behavior. Let $P(\beta, s)(\cdot)$ denote the CDF over scores $\beta^{T}X_{i}(\beta, s)$ that results when agents report covariates in response to a policy with parameters $\beta$ and $s$. The distribution over scores at time step $t$ is given by $P^{t} = P(\beta, S^{t-1})$. Following \cite{bhattacharya2012inferring}, the threshold $S^{t}$ is set to $q(P^{t})$, which is the $q$-th quantile of $P^{t}$, ensuring that only $1-q$ proportion of agents are treated. In the context of college admissions, the college can only accept $1-q$ fraction of the applicant pool, so they admit students who rank above the $q$-th quantile of the score distribution. Accordingly, we set $S^{t} = q(P^{t})$ and $W_{i}^{t} = \pi(X_{i}^{t}; \beta, S^{t}).$
After treatment assignment, the decision maker observes individual outcomes $Y_{i}^{t}$ for each agent. These outcomes may depend on the treatment received. The outcome $Y_{i}(0)$ is observed if the agent is not assigned to treatment, and the outcome $Y_{i}(1)$ is observed when the agent is assigned to treatment, $Y_{i}^{t} = Y_{i}(W^{t}_{i})$. In the context of college admissions $Y_{i}$ may represent the number of months student $i$ enrolls in the college.
Let $V^{t}$ be the value of a policy at time step $t$. We define the policy value to be the mean outcome of the agents after treatments are allocated
\begin{align}
\label{eq:value_eq}
\begin{split}
    V^{t}(\beta) &= \EE[]{Y_{i}(W_{i}^{t})} = V(\beta, S^{t-1}, S^{t}),\\
    &\where V(\beta, s, r) = \EE[]{Y_{i}(\pi(X_{i}(\beta, s); \beta, r))}.
\end{split}
\end{align}

Note that the previous equation makes the dependence of $V^{t}$ on $S^{t-1}$ and $S^{t}$ explicit. The argument $s$ is the previous threshold, which agents react to in the current time step. The argument $r$ is the current threshold, which enforces the capacity constraint in the current time step.

Given a fixed choice of $\beta$, a policy may reach an equilibrium with a stable decision threshold.
When such an equilibrium threshold $s(\beta)$ exists and is unique, it is natural for the decision maker
to seek to maximize the induced equilibrium policy value $V_{\text{eq}}(\beta)$ as defined below.
We provide formal results on existence and uniqueness of equilibrium thresholds in Section \ref{sec:results_equilibrium}.
We note that the objective of optimizing equilibrium welfare is motivated by the observation that it may not be feasible for the
decision maker to change their selection criterion at each time step. Instead, the decision maker aims to select $\beta$ that
performs well when the system is at equilibrium.

\begin{defi}[Equilibrium Policy Value]
\label{def:eq_loss}
Given a fixed $\beta \in \mathcal{B}$, if there is a unique equilibrium threshold $s(\beta)$, then the equilibrium policy value is
$V_{\text{eq}}(\beta) = V(\beta, s(\beta), s(\beta))$,
where $V(\cdot)$ is as defined in \eqref{eq:value_eq}.
\end{defi}

\subsection{Agent Behavior}
\label{sec:agent}
We specify a model for agent behavior. In our model, agents are heterogenous in their raw covariates and ability to modify their covariates, myopic in that they choose their reported covariates based on the previous policy, and imperfect in that their reported covariates are subject to noise.

For every agent $i$, unobservables $(Z_{i}, c_{i}, \epsilon_{i}, Y_{i}(0), Y_{i}(1))$ are sampled i.i.d. from a distribution $F$, where $\epsilon_{i}$ is independent from the other unobservables and has distribution $N(0, \sigma^{2}\mathbf{I}_{d}).$ Motivated by \citet{frankel2019improving, frankel2019muddled}, we assume agents are heterogeneous in their raw covariates $Z_{i}$ and their ability to modify their covariates given by a cost function $c_{i}$. Let $Z_{i} \in \mathcal{X}$, where $\mathcal{X}$ is a convex, compact set in $\mathbb{R}^{d}$. Let $c_{i}(\cdot): \mathbb{R}^{d} \rightarrow \mathbb{R}$ capture agent $i$'s cost deviating from their raw covariates $Z_{i}$. Define $\tilde{X} = \{ x_{1}, x_{2} \in \mathcal{X} \mid x_{1} - x_{2} \}.$ We assume the cost functions $c_{i} \in \mathcal{C} \subset L^{2}(\tilde{\mathcal{X}}, dx)$, where $\mathcal{C}$ is a convex and compact set of functions. The agent is myopic in that they modify their covariates with knowledge of only the criterion $\beta \in \mathcal{B}$ and the previous threshold for receiving treatment $s \in \mathbb{R}$. 

In addition, we suppose the agent has imperfect control over the realized value of their modified covariates. In the context of college admissions, a student can influence their performance on an exam by changing the number of hours they study but cannot perfectly control their exam score. To capture this stochasticity, the agent's modified covariates are subject to noise $\epsilon_{i}.$ We express the agent's utility function as ${U_{i}(x; \beta, s) =  - c_{i}(x - Z_{i}) + \pi(x + \epsilon_{i}; \beta, s).}$ The left term is the cost to the agent of deviating from their raw covariates. The right term is the reward from receiving the treatment. Taking the expectation over the noise yields
\begin{equation}
    \label{eq:expected_utility function}
    \EE[\epsilon]{U_{i}(x; \beta, s)} = -c_{i}(x - Z_{i}) + 1 - \Phi_{\sigma}(s - \beta^{T}x),
\end{equation}
where we denote the Normal CDF with mean 0 and variance $\sigma^{2}$ as $\Phi_{\sigma}$. For example, the expected utility function with quadratic cost is given by \begin{align}
\label{eq:quadratic_utility}
\begin{split}
\EE[\epsilon]{U_{i}(x; \beta, s)} &= - (x - Z_{i})^{T}\text{Diag}(G_{i}) (x - Z_{i}) \\
&\indent+ 1 - \Phi_{\sigma}(s - \beta^{T}x),
\end{split}
\end{align} where $G_{i} \in (\mathbb{R}^{+})^{d}$ and $\text{Diag}(g)$ is a diagonal matrix in $\mathbb{R}^{d \times d}$ with diagonal equal to $g.$

The best response mapping yields the covariates $x \in \mathcal{X}$ that maximize the expected utility, and we define it as ${x^{*}_{i}(\beta, s) = \argmax_{x \in \mathcal{X}} \EE[\epsilon]{U_{i}(x; \beta, s)}}.$ The agent's reported covariates are given by $X_{i}(\beta, s) = x^{*}_{i}(\beta,s) + \epsilon_{i}$, which is the agent's best response subject to noise $\epsilon_{i}$.

\subsection{Properties of Agent Best Response}
\label{sec:best_response_prop}

We establish conditions on the variance $\sigma^{2}$ of the noise distribution that guarantee continuity and contraction properties of the agent best response. Recall that in the context of college admissions, the noise represents that students have imperfect control over their actions. When the noise is low, a student's efforts translate to their desired grades and test scores. In contrast, when the noise is high, there is more variability in a student's grades and test scores at any level of effort. 

We require the following assumption to provide structure to the agent's cost function $c_{i}(\cdot).$

\begin{assumption}
\label{assumption:strong_convexity}
The cost function $c_{i}: \mathbb{R}^{d} \rightarrow \mathbb{R}$ is twice continuously differentiable, $\alpha_{i}$-strongly convex for $\alpha_{i} > 0$, and minimized at the origin.
\end{assumption}

In the context of college admissions, Assumption \ref{assumption:strong_convexity} implies that a student must invest effort to deviate from their baseline test scores and grades. The amount of effort a student must invest to alter their test scores and grades changes smoothly with respect to the difference between modified and raw covariates. There are decreasing returns to deviating from their raw covariates because the cost of modifying their covariates grows at least quadratically. The lowest cost action is to not change their covariates, i.e. to stick with their baseline test scores and grades.

The following result summarizes how agents behave in our model. We demonstrate under Assumption \ref{assumption:strong_convexity} and sufficiently high $\sigma^{2}$, the agent best response exists, is unique, and is continuously differentiable in $(\beta, s)$. With a slightly stronger bound on the variance $\sigma^{2}$, we can additionally verify that the agent's expected score $\omega_{i}(s; \beta):= \beta^{T} x_{i}^{*}(\beta, s)$ is a contraction in $s$, i.e., for any fixed  $\beta \in \mathcal{B}$, there is a Lipschitz constant $\kappa \in (0, 1)$, so that $|\omega_{i}(s; \beta) - \omega_{i}(s'; \beta)| \leq \kappa |s - s'|$ for all $s, s' \in \mathbb{R}.$ 
This agent-level contraction property will be useful for later establishing a mechanism that gives rise to a mean-field (population-level) equilibrium in Section \ref{sec:results_equilibrium}.

\begin{prop}
\label{prop:br_all}
Let $\beta \in \mathcal{B}, s \in \mathbb{R}$, $\bar{\kappa} \in (0, 1].$ Under Assumption \ref{assumption:strong_convexity}, the best response $x^{*}_{i}(\beta, s)$ exists. If moreover $\sigma^{2} > \frac{1}{\alpha_{i} \cdot \sqrt{2\pi e}}$, then $x^{*}_{i}(\beta, s)$ is unique and continuously differentiable in $\beta, s$, whenever $x_{i}^{*}(\beta, s) \in \text{Int}(\mathcal{X})$. Finally, if $\sigma^{2} > \frac{1}{\alpha_{i} \cdot \sqrt{2\pi e}} \cdot \frac{\bar{\kappa} + 1}{\bar{\kappa}}$ additionally holds for some $\bar{\kappa} \in (0, 1]$, then the agent's expected score $\omega_{i}(s; \beta):=\beta^{T}x_{i}^{*}(\beta, s)$ is a contraction in $s$ with Lipschitz constant less than $\bar{\kappa}$ for any $\beta \in \mathcal{B}.$ 
\end{prop}

\begin{figure}
    \centering
    \includegraphics[width=\textwidth]{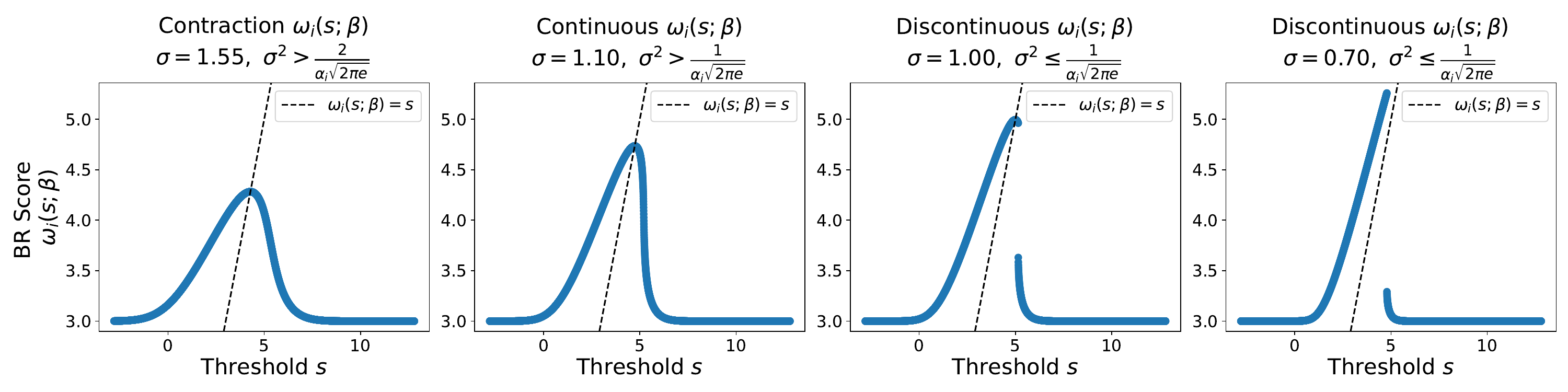}
    \caption{We plot $\omega_{i}(s;  \beta)$ vs. $s$ at different noise levels and observe its continuity and contraction properties.}

    \label{fig:discontinuities}
\end{figure}

We end this section by numerically investigating the role of noise on an agent's best response,
and verify that in the absence of sufficient noise, unstable behaviors may occur.
Qualitatively, in a zero-noise setting, instability arises because there are two modes of agent behavior.
In one mode, the agent defaults to their raw covariates, so $\omega_{i}(s; \beta) \approx \beta^{T}Z_{i}$, either because the threshold is low enough that the agent expects to receive the treatment without modifying their covariates or because the threshold is so high that the agent does not bother modifying their covariates at all. In the other mode, the threshold takes on intermediate values, so the agent will invest the minimum effort to ensure that they receive the treatment under the previous policy, meaning that $\omega_{i}(s; \beta) = s$. When covariate modification is no longer beneficial, the agent defaults to their raw covariates, creating a discontinuity in the best response. The presence of noise increases the agent's uncertainty in whether they will receive the treatment, which causes agents to be less reactive to the previous policy and smooths the agent best response.

Under different noise levels $\sigma^{2}$, we analyze the expected score of an agent's best response with a fixed selection criterion $\beta$ while the threshold $s$ varies. We consider agent $i$ with quadratic cost of covariate modification as in \eqref{eq:quadratic_utility}, where $Z_{i} = [3., 0.]^{T}$ and $G_{i}=[0.1, 1.]^{T}$. We suppose the decision maker's model is $\beta  =[1., 0.]^{T}$. In Figure \ref{fig:discontinuities}, we visualize the expected score $\omega_{i}(s; \beta)$, as a function of $s$, the previous threshold for receiving treatment. We plot $\omega_{i}(s; \beta) \text{ vs. } s$ at four different noise levels $\sigma^{2}$. In the left plot of Figure \ref{fig:discontinuities}, we consider $\sigma^{2} > \frac{1}{\alpha_{i} \cdot \sqrt{2\pi e}} \cdot \frac{\bar{\kappa} + 1}{\bar{\kappa}}$ where $\bar{\kappa}=1$, so the expected score is a contraction in $s$. In the middle left plot, we have that $\sigma^{2} > \frac{1}{\alpha_{i} \cdot \sqrt{2\pi e}}$, so the expected score is continuous. In the plots on the right of Figure \ref{fig:discontinuities}, we have that $\sigma^{2} \leq \frac{1}{\alpha_{i} \cdot \sqrt{2\pi e}}$. In such cases, the best response mapping may be discontinuous and may not necessarily have a fixed point.  

The lack of a fixed point in the expected score in low-noise regimes (rightmost plot, Figure \ref{fig:discontinuities}) implies that there are distributions over agent unobservables for which there is no equilibrium threshold in low-noise regimes. As a result, when the noise condition for continuity of best response does not hold for all agents, an equilibrium in our model may not exist. When we establish uniqueness and existence of the equilibrium in our model in Section \ref{sec:results_equilibrium}, we assume a noise condition that guarantees continuity of the agents' best response mappings.

\end{section}

\begin{section}{Mean-Field Results}
\label{sec:results_equilibrium}
Recall that the decision maker's objective, as outlined in Section \ref{sec:model}, is to find a selection criterion $\beta$ that maximizes the equilibrium policy value $V_{\text{eq}}(\beta)$. This is a sensible goal in settings where an equilibrium exists and is unique for each selection criterion $\beta$ in consideration. In this section, we characterize the equilibrium in the mean-field regime.

In the mean-field regime, an infinite population of agents with unobservables sampled from $F$ is considered for the treatment at each time step $t$. Let $\beta$ be the decision maker's fixed selection criterion. At time step $t$, agents report covariates with knowledge of criterion $\beta$ and previous threshold $S^{t-1}$, inducing a distribution over scores $P^{t}= P(\beta, S^{t-1}).$ Then, at time step $t$, the agents who score above threshold $S^{t} = q(P^{t})$ will receive treatment. The thresholds evolve in a deterministic fixed-point iteration process:
\begin{equation} 
\label{eq:fpi}
S^{t} = q(P(\beta, S^{t-1})) \quad t = 1, 2, \dots
\end{equation}
The system is at equilbrium if the threshold for receiving treatment is fixed over time. The equilibrium induced by $\beta$ is characterized by an \textit{equilibrium threshold} $s(\beta)$, which is the threshold $s \in \mathbb{R}$ that satisfies $s = q(P(\beta, s))$. To study equilibria, we make two further assumptions:

\begin{assumption}
\label{assumption:finite_types}
The $(Z, c)$-marginal of $F$ has finite support.
\end{assumption}

\begin{assumption}
\label{assumption:br_interior} 
For an agent $i$ with unobservables in the support of $F$, $x^{*}_{i}(\beta, s) \in \text{Int}(\mathcal{X})$.
\end{assumption}

In the context of college admission, Assumption \ref{assumption:finite_types} requires a finite number of student types. Realistically, this assumption may not hold because there could be an infinite number of types, but this assumption is made for technical convenience and we conjecture that similar results will hold when $F$ has continuous support.
Meanwhile, Assumption \ref{assumption:br_interior} excludes the scenario where agent best responses ``bunch” at the boundaries of $\mathcal{X}$. In the context of college admissions, we require that the space of possible test scores and grades is large enough that students' post-effort test scores and grades lie in the interior of the set. \citet{frankel2019muddled} argue that if this assumption is violated, it is possible to expand the covariate space to ensure that best responses lie in the interior, say by making an exam harder.

The following result establishes conditions under which a mean-field equilibrium threshold exists, is unique, and can be related to $\beta$ via a differentiable function $s(\beta).$
We use notation $\alpha_{*}(F) = \inf_{i \in \mathbb{N}} \alpha_{i}$, and note that under Assumption \ref{assumption:strong_convexity} and \ref{assumption:finite_types} $\alpha_{*}(F)$ is positive. We will omit the dependence of $\alpha_{*}(F)$ on $F$ when it is clear that there is only one distribution over unobservables of interest. 

\begin{theo}
\label{theo:equilibrium}
Under Assumptions \ref{assumption:strong_convexity},
\ref{assumption:finite_types}, and
\ref{assumption:br_interior}, and if $\sigma^{2} > \frac{1}{\alpha_{*} \cdot \sqrt{2\pi e}}$, then $q(P(\beta, s))$ is continuously differentiable in $\beta, s.$ Furthermore, for any $\beta \in \mathcal{B}$, there is a unique fixed point $s(\beta)$ satisfying $s(\beta) = q(P(\beta, s(\beta))),$ and $s(\beta)$ is continuously differentiable in $\beta$. If moreover $\sigma^{2} > \frac{1}{\alpha_{*} \cdot \sqrt{2\pi e}} \cdot \frac{\bar{\kappa} + 1}{\bar{\kappa}}$ for $\bar{\kappa} \in (0, 1]$, then, for any $\beta \in \mathcal{B}$, $q(P(\beta, s))$ is a contraction in $s$ with Lipschitz constant less than $\bar{\kappa}$ and the fixed-point iteration \eqref{eq:fpi} converges to $s(\beta)$. 
\end{theo}

The last part of Theorem \ref{theo:equilibrium} gives conditions for the population quantile mapping $q(P(\beta, \cdot))$ to be a contraction with Lipschitz constant less than $\bar{\kappa}$. 
Importantly, these conditions are sufficient, but not necessary. We find that the derivative of $q(P(\beta, \cdot))$ is a convex combination of the derivatives of the agents' expected scores $\omega_{i}(\cdot; \beta)$. Our conditions imply that these derivatives have magnitude less than $\bar{\kappa}$ for all agents, so $q(P(\beta, \cdot))$ must also have derivative with magnitude less than $\bar{\kappa}$ and thus be a contraction with Lipschitz constant less than $\bar{\kappa}$. Nevertheless, it is possible for $q(P(\beta, \cdot))$ to be a contraction even if $\omega_{i}(s; \beta)$ is not a contraction for every agent $i$. 


\end{section}

\begin{section}{Finite-Sample Approximation}
\label{sec:finite_results}
Understanding equilibrium behavior of our model in the finite regime is of interest because our ultimate goal is to learn optimal equilibrium policies in finite samples. In this section, we instantiate the model from Section \ref{sec:model} in the regime where a finite number of agents are considered for the treatment. A difficulty of the finite regime is that deterministic equilibria do not exist. Instead, we give conditions under which stochastic equilibria arise and show that, in large samples, these stochastic equilibria sharply approximate the mean-field limit derived above.

\begin{figure}[t]
    \centering
    \includegraphics[width=0.5\textwidth]{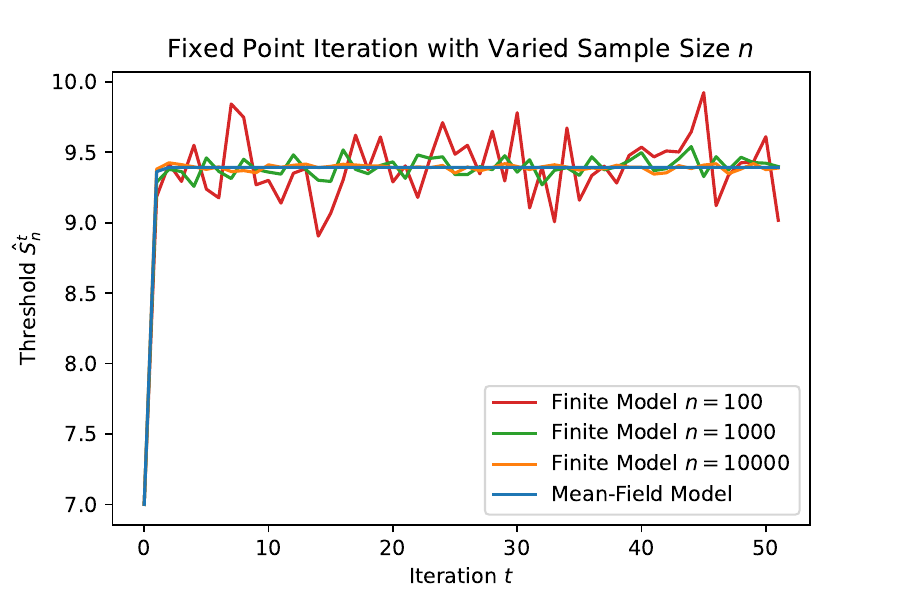}
    \caption{We plot iterates an example process \eqref{eq:emp_fpi}. For large $n$, iterates concentrate about the mean-field equilibrium. }
    \label{fig:fpi}
\end{figure}

Let $\beta$ be the decision maker's fixed selection criterion. At each time step, $n$ new agents with unobservables sampled i.i.d. from $F$ are considered for the treatment. In the context of college admissions, the sampled agents at each time step represent a cohort of students applying for admission each year. At time step $t$, the $n$ agents who are being considered for the treatment report their covariates with knowledge of the criterion $\beta$ and previous threshold $\hat{S}^{t-1}_{n}$. Let $P^{n}(\beta, s)$ be the empirical score distribution when $n$ agents best respond to a policy $\beta$ and threshold $s$. So, the distribution over scores at time step $t$ is given by $P^{n, t} = P^{n}(\beta, \hat{S}^{t-1}_{n}).$ Let  $q(P^{n, t})$ denote the $q$-th quantile of $P^{n, t}$. Then, agents who score above $\hat{S}^{t}_{n} = q(P^{n, t-1})$ will receive the treatment. The thresholds evolve in a stochastic fixed-point process
\begin{equation}
\label{eq:emp_fpi}
\hat{S}^{t}_{n} = q(P^{n, t}) = q(P^{n}(\beta, \hat{S}^{t-1}_{n})), \quad t=1, 2, \dots
\end{equation}
Since new agents are sampled at each time step, $q(P^{n}(\beta, \cdot))$ is a random operator. Iterating this operator given some initial threshold $\hat{S}^{0}_{n}$ yields a stochastic process $\{\hat{S}^{t}_{n}\}_{t \geq 0}$. Figure \ref{fig:fpi} visualizes an example of this process.

We note that for any fixed $\beta$, the random operator $q(P^{n}(\beta, \cdot))$ approximates the deterministic function $q(P(\beta, \cdot))$. The following result gives a finite-sample concentration inequality for the behavior of $q(P^{n}(\beta, s))$. 

\begin{lemm}
\label{lemm:constant_in_quantile_bound}
Fix $\beta \in \mathcal{B}$. For $\lambda > 0$, define \begin{align}
\label{eq:M_eps}
\begin{split}
    M_{\lambda} = \inf_{s \in \mathbb{R}} \min\{&P(\beta, s)(q(P(\beta, s)) + \lambda) - q,\\
    &\indent q - P(\beta, s)(q(P(\beta, s)) - \lambda)\}.
\end{split}
\end{align}
Under Assumption \ref{assumption:strong_convexity}, \ref{assumption:finite_types}, \ref{assumption:br_interior}, if $\sigma^{2} > \frac{1}{\alpha_{*} \sqrt{2 \pi e}},$ then $M_{\lambda} > 0$, $M_{\lambda}$ is approximately linear in $\lambda$ when $\lambda \rightarrow 0$, and
 \[ P(|q(P(\beta, s)) - q(P^{n}(\beta, s))| < \lambda) \geq 1 - 4e^{-2nM_{\lambda}^{2}}.\]
\end{lemm}

Notably, the bound in the concentration inequality does not depend on the particular choice of $s$. We use this lemma to characterize the behavior of the finite system for sufficiently large iterates $t$ and number of agents $n$. Theorem \ref{theo:emp_fpi_convergence} shows that under the same conditions that enable fixed-point iteration in the mean-field model \eqref{eq:fpi} to converge to the mean-field equilibrium threshold $s(\beta)$, sufficiently large iterates of the stochastic fixed-point iteration in the finite model \eqref{eq:emp_fpi} will lie in a small neighborhood about $s(\beta)$ with high probability. We can view these iterates as stochastic equilibria of the finite system.

\begin{theo}
\label{theo:emp_fpi_convergence}
Fix $\beta \in \mathcal{B}$. Suppose Assumptions \ref{assumption:strong_convexity}, \ref{assumption:finite_types}, \ref{assumption:br_interior} hold. Let $\lambda \in (0, 1), \delta \in (0, 1)$, $s(\beta)$ is the mean-field equilibrium threshold induced by $\beta$, and $M_{\lambda}$ is defined by \eqref{eq:M_eps}. Let $\kappa$ be the Lipschitz constant of $q(P(\beta, \cdot))$. Let $\lambda_{g} = \frac{\lambda (1 - \kappa)}{2}$ and $C = |\hat{S}^{0}_{n} - s(\beta)|.$  If $\sigma^{2} > \frac{2}{\alpha_{*}\sqrt{2\pi e}},$ then for $n, t$ such that $t \geq \Big\lceil \frac{\log(\frac{\lambda}{2C}) }{\log \kappa} \Big\rceil$ and $n \geq \frac{1}{2M_{\lambda_{g}}^{2}} \log (\frac{4t}{\delta}),$ then $P(|\hat{S}_{n}^{t} - s(\beta)| \geq \lambda ) \leq \delta.$
\end{theo}

Under our assumptions, we can apply Theorem \ref{theo:equilibrium} to see that $\kappa$, the Lipschitz constant of the population quantile mapping $q(P(\beta, \cdot))$, is a constant that is less than 1. Since $\lambda_{g}$ depends linearly on $(1- \kappa)$ and $M_{\lambda}$ is approximately linear in $\lambda$ for small $\lambda$, we find that the required sample complexity scales with $\frac{1}{(1 - \kappa)^{2}}.$

\begin{coro}
\label{coro:convergence_in_prob_threshold}
Fix $\beta \in \mathcal{B}$. Let $\{t_{n}\}$ be a sequence such that $t_{n} \uparrow \infty$ as $n \rightarrow \infty$ and $t_{n} \prec \exp(n)$ ($t_{n}$ grows slower than exponentially fast in $n$). Under the conditions of Theorem \ref{theo:emp_fpi_convergence}, $\hat{S}_{n}^{t_{n}} \xrightarrow{p} s(\beta),$ where $s(\beta)$ is the unique fixed point of $q(P(\beta, \cdot)).$
\end{coro}
\end{section}

\begin{section}{Learning Policies via Gradient-Based Optimization}
\label{sec:results_learning}
In this section, we define the policy gradient, the gradient of the equilibrium policy value with respect to the selection criterion, give an estimator of the policy gradient, and use the estimator to learn policies. We use a local unit-level experimentation scheme as in \cite{munro2021treatment, wager2021experimenting} to estimate the gradient and gradient-based optimization to learn policies.

\begin{lemm}
\label{lemm:loss_deriv}
Under the conditions of Theorem \ref{theo:equilibrium}, $V_{\text{eq}}(\beta)$ from Definition \ref{def:eq_loss} is continuously differentiable in $\beta$, and
\begin{align}
\label{eq:total_deriv}
\begin{split}
    \frac{dV_{\text{eq}}}{d \beta}(\beta) &=\frac{\partial V}{\partial \beta}(\beta, s(\beta), s(\beta)) \\
    &+ \Big(\frac{\partial V}{\partial s}(\beta, s(\beta), s(\beta)) \\
    &+  \frac{\partial V}{\partial r}(\beta, s(\beta), s(\beta))\Big) \cdot \frac{\partial s}{\partial \beta}(\beta).
\end{split} 
\end{align}
\end{lemm}

We decompose the total derivative of $V_{\text{eq}}(\beta)$, or the policy gradient, into two parts. The first term corresponds to the model gradient and the second term corresponds to the equilibrium gradient.
\begin{defi}[Model Gradient] The model gradient $\tau_{\text{MG}}$ is $\tau_{\text{MG}}(\beta) = \frac{\partial V}{\partial \beta}(\beta, s(\beta), s(\beta)).$
\end{defi}

The selection criterion $\beta$ impacts the policy value because the criterion is used to score the agents and agents modify their covariates in response to the criterion. Optimization of the selection criterion using the model gradient is a policy learning approach that accounts for agents' strategic behavior.

Due to the decision maker's capacity constraint, the equilibrium threshold for receiving treatment also depends on the selection criterion. So, we must also account for how policy value changes with respect to the equilibrium threshold and how the equilibrium threshold changes with respect to the selection criterion. Following notation from \eqref{eq:value_eq}, we write $\partial V / \partial s$ and $\partial V / \partial r$ for the partial derivatives of $V$ in its second and third arguments respectively.

\begin{defi}[Equilibrium Gradient] The equilibrium gradient $\tau_{\text{EG}}$ is $\tau_{\text{EG}}(\beta) =\Big( \frac{\partial V}{\partial s}(\beta, s(\beta), s(\beta)) + \frac{\partial V}{\partial r}(\beta, s(\beta), s(\beta)) \Big) \cdot \frac{\partial s}{\partial \beta}(\beta).$
\end{defi}
The previous threshold for receiving treatment $s$ impacts the policy value because agents modify their covariates in response to $s$. This influences the treatments that agents receive and thus the policy value. The current threshold for receiving treatment $r$ impacts the policy value because it determines agents' treatment assignments, which influences the policy value, as well. At equilibrium, we have that $s=r=s(\beta)$, so we can account for both of these effects simultaneously.

\begin{defi}[Policy Gradient]
The policy gradient $\tau_{\text{PG}}$ is $\tau_{\text{PG}}(\beta) = \tau_{\text{MG}}(\beta) + \tau_{\text{EG}}(\beta).$
\end{defi}

This decomposition is related to the decomposition theorem of \cite{hu2021average}. \cite{hu2021average} consider a Bernoulli trial, where treatments are allocated according to $W_{i} \sim \text{Bernoulli}(p_{i})$ for $p \in (0, 1)^{n}$, and they demonstrate that the effect of a policy intervention that infinitesimally increases treatment probabilities can be decomposed into the average direct and indirect effects. The average direct effect captures how the outcome $Y_{i}$ of a unit is affected by its own treatment $W_{i}$ on average. The average indirect effect is the term that captures the responsiveness of outcome $Y_{i}$ to the treatment of other units $j\neq i$ on average, measuring the effect of cross-unit interference. When there is no cross-unit interference, the average direct effect matches the usual average treatment effect \citep{hu2021average, savje2021average}. In our setting, the policy gradient consists of the model gradient, which captures how changes in the selection criterion directly impact the policy value through scoring the agents and agents' strategic behavior, and the equilibrium gradient, which captures how changes in the selection criterion indirectly impact the policy value by modulating the threshold for receiving treatment. In the absence of capacity constraints, the policy gradient is just the model gradient.

We derive estimators for the policy gradient through a unit-level randomized experiment in finite samples. In a system consisting of $n$ agents, we apply symmetric, mean-zero perturbations to the parameters of the policy that each agent responds to. Let $R$ represent the distribution of Rademacher random variables and let $R^{d}$ represent a distribution over $d$-dimensional Rademacher random variables. For agent $i$, we apply perturbations of magnitude $b > 0$ to the policy parameters. We set $\beta_{i} = \beta + b\zeta_{i}$ where $\zeta_{i} \sim R^{d}$ and $s_{i} = s + b\xi_{i}$ where $\xi_{i} \sim R.$
In practice, the perturbation to the selection criterion can be implemented by telling agent $i$ that they will be scored according to $\beta_{i}$ instead of $\beta$. The perturbation to the threshold $s$ can be implemented by publicly reporting the previous threshold $s$ but telling agent $i$ that a small shock of size $-b\xi_{i}$ will be added to the agent's score in the next time step. We extend our myopic agent model and assume that an agent $i$ will report covariates in response a policy $\pi(x; \beta_{i}, s_{i})$ as follows:
\begin{align}
\label{eq:perturbed_best_response}
\begin{split}
X_{i}(\beta_{i}, s_{i}) &= x_{i}^{*}(\beta_{i}, s_{i}) + \epsilon_{i}, \\
\where x_{i}^{*}(\beta_{i}, s_{i}) &= \argmax_{x \in \mathcal{X}} \EE[\epsilon]{U_{i}(x; \beta_{i}, s_{i})}.
\end{split}
\end{align}
The prescribed perturbations are applied to determine the agents' scores and treatment assignments. For clarity, we contrast the form of a score in the unperturbed setting with the perturbed setting. In the unperturbed setting, an agent with unobservables $(Z_{i}, c_{i})$ who best responds to $\beta, s$ will obtain a score $\beta^{T}X_{i}(\beta, s)$. In the perturbed setting, an agent with unobservables $(Z_{i}, c_{i})$ who best responds to a perturbed version of $\beta, s$ will obtain a score $\beta_{i}^{T}X_{i}(\beta_{i}, s_{i}) - b\xi_{i}.$ Let $P(\beta, s, b)(\cdot)$ denote the distribution over scores in the perturbed setting. The threshold for receiving treatment in the experiment is $r = q(P^{n}(\beta, s, b)).$ The agent-specific threshold for receiving treatment is given by $r_{i} = r + b\xi_{i}.$ So, in the experiment, the treatments are allocated as
$W_{i} = \pi(X_{i}(\beta_{i}, s_{i}); \beta_{i}, r_{i})$.
As in \cite{munro2021treatment, wager2021experimenting}, the purpose of applying these perturbations is so that we can recover relevant gradient terms without disturbing the equilibrium behavior of the system.

We compute the gradients by running regressions from the perturbations to quantities of interest, which include the observed outcomes and indicators for agents who score above the threshold $r$. To construct the estimators of the model and equilibrium gradients, we rely on gradient estimates of the policy value function $V(\beta, s, r)$ and gradient estimates of the complementary CDF of the score distribution $\Pi(\beta, s)(\cdot)$, which is defined as $\Pi(\beta, s)(\cdot) = 1 - P(\beta, s)(\cdot).$ 

In this experiment, we suppose that thresholds evolve by the stochastic fixed-point iteration process below. Note that it differs slightly from the process given in \eqref{eq:emp_fpi}.
\begin{equation} 
\label{eq:stochastic_fpi_with_perturbation} 
\hat{S}^{t}_{b, n} = \min(\max(q(P^{n}(\beta, \hat{S}^{t-1}_{b, n}), -D)), D).
\end{equation}
This process differs from \eqref{eq:emp_fpi} because it includes perturbations of size $b$ to the selection criterion and threshold and restricts the threshold to a bounded set $\mathcal{S}=[-D, D]$ where $D$ is sufficiently large constant so that $s(\beta) \in \mathcal{S}$. Such a set exists because it can be shown there exists $D > 0$ such that $|q(P(\beta, s))|< D$ for all $s \in \mathbb{R}$.

Analyzing the stochastic process $\{\hat{S}^{t}_{b, n}\}_{t \geq 0}$ generated by \eqref{eq:stochastic_fpi_with_perturbation} presents two technical challenges. First, \eqref{eq:stochastic_fpi_with_perturbation} truncates the threshold values so that they lie in $\mathcal{S}$, whereas the process analyzed in Section \ref{sec:finite_results} does not involve truncation. Nevertheless, the truncation is a contraction map to the equilibrium threshold, so the results of Section \ref{sec:finite_results} also apply to stochastic fixed point iteration with truncation. The other challenge is that the results from Section \ref{sec:results_equilibrium} and \ref{sec:finite_results} focus on the setting where all agents best respond to the same policy $\pi(x; \beta, s)$, whereas in \eqref{eq:stochastic_fpi_with_perturbation}, each agent $i$ best responds to a different perturbed policy $\pi(x; \beta_{i}, s_{i})$. 

To show that results from Section \ref{sec:finite_results} transfer to the setting with unit-level perturbations, we can define a new distribution over agent unobservables $\tilde{F}$ that are related to the original distribution over agent unobservables $F$. When agents with unobservables sampled from $\tilde{F}$ best respond to $\pi(x; \beta, s)$, the score distribution that results equals $P(\beta, s, b)$. 

We require the following assumption to guarantee that the transformed unobservables in $\text{supp}(\tilde{F})$ can be defined on the same spaces as the original unobservables in $\text{supp}(F).$
\begin{assumption}
\label{assumption:etas_in_interior}
For any $Z_{i} \in \text{supp}(F_{Z}),$ $Z_{i} \in \text{Int}(\mathcal{X}).$ In addition, $c_{i'} \in \mathcal{C}$, where $c_{i'}(y) = c_{i}(y) - \phi_{\sigma}(s - r)\beta^{T}y$ for any $c_{i} \in \text{supp}(F_{c}), \beta \in \mathcal{B}, s, r \in \mathcal{S}.$
\end{assumption}
In the context of college admissions, Assumption \ref{assumption:etas_in_interior} requires that $\mathcal{X}$ is large enough that agents' raw covariates do not ``bunch'' at the boundaries of $\mathcal{X}$, and that $\mathcal{C}$ is large enough that it contains cost functions that are linear offsets of the cost functions in $\text{supp}(F_{c})$. This assumption is also plausible in the context of college admissions; e.g., students would likely not get exactly 0 test score even if they do not try, and the set of possible cost functions is rich. Note that if this assumption is not satisfied for a given distribution $F$, we can likely enlarge the space of cost functions $\mathcal{C}$ and covariates $\mathcal{X}$. 

We define the model gradient estimator. Suppose $n$ agents are considered for the treatment. Let each row of $\mathbf{M}_{\beta} \in \mathbb{R}^{n \times d}$ correspond to $b\zeta_{i}^{T}$, the perturbation applied to the selection criterion $\beta$ observed by the $i$-th agent. Let each entry of $Y \in \mathbb{R}^{n}$ correspond to the observed outcome of the $i$-th agent, so $Y_{i}= Y_{i}(W_{i}).$ The regression coefficient obtained by running OLS of $Y$ on $\mathbf{M}_{\beta}$ is
\[ \hat{\Gamma}_{Y, \beta}^{b, n}(\beta, s, r) = b^{-1}\Big(\frac{1}{n} \sum_{i=1}^{n}\zeta_{i}\zeta_{i}^{T}\Big)^{-1}\Big(\frac{1}{n} \sum_{i=1}^{n}\zeta_{i}Y_{i}\Big).\]
The model gradient estimator with sample size $n$, perturbation size $b$, and iteration $t$ as \begin{equation}
    \label{eq:de_estimator}
    \hat{\tau}_{\text{MG},b, n}^{t}(\beta) = \hat{\Gamma}^{b, n}_{Y
   , \beta}(\beta, \hat{S}^{t-1}_{b, n}, \hat{S}^{t}_{b, n}), 
\end{equation}
where $\hat{S}^{t}_{b, n}$ is given by \eqref{eq:stochastic_fpi_with_perturbation}.

One of the challenges in analyzing this estimator is that $\hat{\Gamma}^{b, n}_{Y, \beta}$ is a stochastic function and its arguments are also stochastic. To demonstrate consistency of the overall estimator, we must establish stochastic equicontinuity for $\hat{\Gamma}^{b, n}_{Y, \beta}$. As a result, we require the following assumption on the outcomes $Y_{i}(w).$
\begin{assumption}
\label{assumption:nice_loss}
Let $m(z, c; w)= \EE[]{Y_{i}(w) \mid Z_{i}=z, c_{i}=c}.$ For $w \in \{0, 1\}$, the potential outcome $Y_{i}(w)$ can be decomposed as $Y_{i}(w) =  m(Z_{i}, c_{i}; w) + \rho_{i}$, where $\EE{\rho_{i} \mid Z_{i}, c_{i}} = 0$ and $m$ is continuous with respect to $(z, c) \in \mathcal{X} \times L^{2}(\tilde{\mathcal{X}}, dx)$ and bounded.
\end{assumption}
This assumption essentially states that agents with the same raw covariates and cost function for covariate modification, on average, have the same potential outcomes upon being accepted or rejected. This is plausible in the context of college admissions if space of raw features and cost functions is rich. In addition, we also require that the conditional mean outcome is continuous with respect to the raw covariates and cost function for covariate modification.

\begin{theo}
\label{theo:direct_effect}
Fix $\beta \in \mathcal{B}.$ Let $\{t_{n}\}$ be a sequence such that $t_{n} \uparrow \infty$ as $n \rightarrow \infty$ and $t_{n} \prec \exp(n).$ Let $\mathcal{S} = [-D, D]$ for a sufficiently large constant $D> 0$, so that the equilibrium threshold $s(\beta) \in \mathcal{S}.$ Under Assumptions \ref{assumption:strong_convexity}, 
\ref{assumption:finite_types},
\ref{assumption:br_interior},
\ref{assumption:etas_in_interior}, and
\ref{assumption:nice_loss},  if $\sigma^{2} > \frac{2}{\alpha_{*}\sqrt{2\pi e}},$ then there exists a sequence $\{b_{n}\}$ such that $b_{n} \rightarrow 0$ so that $\hat{\tau}_{\text{MG},b_{n}, n}^{t_{n}}(\beta) \xrightarrow{p} \tau_{\text{MG}}(\beta).$ 
\end{theo}

Second, we define the equilibrium gradient estimator. Although the same approach applies, estimating the equilibrium gradient is more complicated than estimating the model gradient. We estimate the equilibrium gradient by estimating the two components of the equilibrium gradient, $\frac{\partial V}{\partial s} + \frac{\partial V}{\partial r}$ and $\frac{\partial s}{\partial \beta}.$ 

Again, suppose $n$ agents are considered for the treatment. Let each row of $\mathbf{M}_{\beta} \in \mathbb{R}^{n \times d}$ and of $\mathbf{M}_{s} \in \mathbb{R}^{n \times 1}$ correspond to the perturbation applied to the selection criterion $\beta$ and threshold $s$, respectively for the $i$-th agent. Let each entry of $Y, I \in \mathbb{R}^{n}$ correspond to the following quantities for the $i$-th agent
\[Y_{i} = Y_{i}(W_{i}), \quad I_{i} = \pi(X_{i}(\beta_{i}, s_{i}); \beta_{i}, r).\]
The regression coefficients from running OLS of $Y$ on $\mathbf{M}_{s}$, $I$ on $\mathbf{M}_{\beta}$, and $I$ on $\mathbf{M}_{s}$ are given by
\begin{align*}
    \hat{\Gamma}_{Y, s, r}^{b, n}(\beta, s, r) &= b^{-1}\Big(\frac{1}{n} \sum_{i=1}^{n}\xi_{i}\xi_{i}^{T}\Big)^{-1}\Big(\frac{1}{n} \sum_{i=1}^{n}\xi_{i}Y_{i}\Big), \\
    \hat{\Gamma}_{\Pi, \beta}^{b, n}(\beta, s, r) &= b^{-1}\Big(\frac{1}{n} \sum_{i=1}^{n}\zeta_{i}\zeta_{i}^{T}\Big)^{-1}\Big(\frac{1}{n} \sum_{i=1}^{n}\zeta_{i}I_{i}\Big), \\
    \hat{\Gamma}_{\Pi, s}^{b, n}(\beta, s, r) &= b^{-1}\Big(\frac{1}{n} \sum_{i=1}^{n}\xi_{i}\xi_{i}^{T}\Big)^{-1}\Big(\frac{1}{n} \sum_{i=1}^{n}\xi_{i}I_{i}\Big).
\end{align*}
Let $\{h_{n}\}$ be a sequence such that $h_{n} \rightarrow 0$ and $nh_{n} \rightarrow \infty$. Let $p^{n}(\beta, s, b)(r)$ denote a kernel density estimate of $p(\beta, s, b)(r)$ with kernel function $k(z) = \mathbb{I}(z \in [- \frac{1}{2}, \frac{1}{2}))$ and bandwidth $h_{n}.$
We define the equilibrium gradient estimator with sample size $n$ and iteration $t$ as
\begin{align} 
\label{eq:ie_estimator}
\begin{split}
&\hat{\tau}_{\text{EG},b, n}^{t}(\beta) \\
&= \hat{\Gamma}^{b, n}_{Y, s, r}(\beta, \hat{S}^{t-1}_{b, n}, \hat{S}^{t}_{b, n}) \cdot \\
&\Bigg( \frac{1}{p^{n}(\beta, \hat{S}^{t}_{b, n}, b)(\hat{S}^{t}_{b, n}) - \hat{\Gamma}_{\Pi, s}^{b, n}(\beta, \hat{S}^{t}_{b, n}, \hat{S}^{t}_{b, n}) } \cdot \\
&\indent\hat{\Gamma}_{\Pi, \beta}^{b, n}(\beta, \hat{S}^{t}_{b, n}, \hat{S}^{t}_{b, n})\Bigg).
\end{split}
\end{align}
Theorem \ref{theo:indirect_effect} shows that this estimator is consistent for the equilibrium gradient.
\begin{theo}
\label{theo:indirect_effect}
Fix $\beta \in \mathcal{B}.$ Let $\{t_{n}\}$ be a sequence such that $t_{n} \uparrow \infty$ as $n \rightarrow \infty$ and $t_{n} \prec \exp(n).$ Let $\mathcal{S} = [-D, D]$ for a sufficiently large constant $D> 0$, so that the equilibrium threshold $s(\beta) \in \mathcal{S}.$ Under the conditions of Theorem \ref{theo:direct_effect}, there exists a sequence $\{b_{n}\}$ such that $b_{n} \rightarrow 0$ so that $\hat{\tau}_{\text{EG}, b_{n}, n}^{t_{n}}(\beta) \xrightarrow{p} \tau_{\text{EG}}(\beta).$ 
\end{theo}

To estimate the policy gradient, we sum the model and equilibrium gradient estimators.

\begin{coro}
\label{coro:total}
Fix $\beta \in \mathcal{B}.$ Let $\{t_{n}\}$ be a sequence such that $t_{n} \uparrow \infty$ as $n \rightarrow \infty$ and $t_{n} \prec \exp(n).$ Let $\mathcal{S} = [-D, D]$ for a sufficiently large constant $D> 0$, so that the equilibrium threshold $s(\beta) \in \mathcal{S}.$ We consider the sequence of approximate policy gradients given by
\[ \hat{\tau}_{\text{PG},b_{n}, n}^{t_{n}}(\beta) = \hat{\tau}_{\text{MG},b_{n}, n}^{t_{n}}(\beta) + \hat{\tau}_{\text{EG},b_{n}, n}^{t_{n}}(\beta).\]
Under the conditions of Theorem \ref{theo:direct_effect}, there exists a sequence $\{b_{n}\}$ such that $b_{n} \rightarrow 0$ so that $\hat{\tau}_{\text{PG}, b_{n}, n}^{t_{n}}(\beta) \xrightarrow{p} \tau_{\text{PG}}(\beta).$  
\end{coro}

These policy gradients can then be used for learning an optimal policy.
Following \citet{wager2021experimenting},we first learn equilibrium-adjusted gradients of the
policy value as discussed above and then update the selection criterion with the gradient to maximize the policy value;
see Algorithm \ref{alg:method} in the appendix for details.
In this paper, we will only investigate
empirical properties of this approach and refer to \citet{wager2021experimenting}
for formal results for this type of gradient-based learning.

The decision maker runs the algorithm for $J$ epochs. In Section \ref{sec:model}, we describe that it may be infeasible for the decision maker to update the selection criterion at each time step. This algorithm requires the decision maker to deploy an updated selection criterion at each epoch $j$. In other words, updates to the selection criterion are necessary but infrequent. We emphasize that deploying different selection criteria is only necessary for the learning procedure, and ultimately, we aim to learn a fixed selection criterion that maximizes the equilibrium policy value.
 
 In epoch $j$, the decision maker deploys criterion $\beta^{j}$. By iterating \eqref{eq:stochastic_fpi_with_perturbation}, a stochastic equilibrium induced by $\beta^{j}$ emerges, yielding the equilibrium threshold $s^{j}$. Each agent best responds to their perturbed policy and the decision maker observes their reported covariates. Following the procedure given above, the decision maker can then estimate the policy gradient of $\beta^{j}$ on the equilibrium policy value (Algorithm \ref{algorithm:gradient_estimates}). The decision maker can set $\beta^{j+1}$ by taking a gradient step from $\beta^{j}$ using the policy gradient estimator as the gradient. Note that because we aim to maximize the policy value, we update $\beta$ by moving in the direction of the gradient.

\end{section}

\begin{section}{Numerical Evaluation}
\label{sec:exp}
We demonstrate that the policy gradient estimator defined in Section \ref{sec:results_learning} can be used to learn a policy that achieves higher equilibrium policy value compared to approaches that only account for capacity constraints or only account for strategic behavior. In a semi-synthetic experiment using data from the National Education Longitudinal Study of 1988 (NELS)
\citep{ingels1994national}, we also discuss how the learned policy impacts the distribution of accepted students.

We evaluate our approach along with two baselines, as described below. The first baseline, capacity-aware policy learning, considers capacity constraints but ignores strategic behavior, and the second baseline, strategy-aware gradient-based optimization, accounts for strategic behavior but ignores capacity constraints. Our proposed method accounts for both strategic behavior and capacity constraints. All methods only use sampled data and do not have access to unobservables.
\paragraph{Capacity-Aware Policy Learning} Following \cite{bhattacharya2012inferring}, the decision maker runs a randomized controlled trial (RCT) to estimate the conditional average treatment effect (CATE) $\tau(x) = \EE[]{Y_{i}(1) - Y_{i}(0) \mid X=x}$ and at deployment, assigns treatment to students with CATE estimate above the $q$-th quantile. Note that students are not strategic in the RCT because treatment assignment is random but will be strategic at deployment. In our implementation, we obtain a CATE estimate of the form $\beta_{1}x + \beta_{0}$ by estimating the conditional mean outcomes via linear regression and subtracting the models. We refer to this method's learned policy as $\beta_{\text{cap}} = \text{Proj}_{\mathcal{B}}(\beta_{1}),$ which is a projection of the parameters of the CATE onto the allowed policy class $\mathcal{B}$.
\paragraph{Strategy-Aware Gradient-Based Optimization} The decision maker runs a unit-level experiment each epoch to obtain a consistent estimate the model gradient (Section \ref{sec:results_learning}). The decision maker then updates the selection criterion $\beta$ via projected gradient descent. Recall that the model gradient accounts for students' strategic behavior but does not account for the equilibrium effect. We refer to this method's learned solution as $\beta_{\text{strat}}$.
\paragraph{Competition-Aware Gradient-Based Optimization} In our proposed approach, the decision maker runs a unit-level experiment each epoch to obtain a consistent estimate of the policy gradient (Section \ref{sec:results_learning}). The decision maker then updates the selection criterion $\beta$ via projected gradient descent. We refer to the learned solution of this method as $\beta_{\text{comp}}.$ 

We evaluate these methods on a semi-synthetic policy learning experiment for college admissions. Using NELS data, we construct a realistic distribution over agent unobservables.
NELS is a nationally representative, longitudinal study that followed eighth graders in 1988 throughout their secondary and postsecondary years. NELS includes socioeconomic status (SES), twelfth grade standardized test scores in reading, math, science, and history and average grades in English, math, science, social studies, and foreign language, and post-secondary outcomes for $n=14915$ students.

We construct a simulated distribution over agent unobservables using the NELS data.\footnote{Due to computational constraints of the data generation, we construct a distribution over $K = 8$ representative agent types instead of using the empirical distribution over all $m$ unobservables. Additional details on the dataset generation are provided in Appendix \ref{subsec:constr}.} We assume a student's $d$-dimensional ($d=9$) best response $x_{i}^{*}$ corresponds to reported test scores and grades from the NELS dataset. We use the subscripts ``test scores" and ``grades'' to specify the covariates that correspond to test scores and grades, respectively.
We also assume that students have a quadratic cost of modifying covariates (see \eqref{eq:quadratic_utility}), where the cost of modifying test scores is low and is the same for all students and the cost of modifying grades is high and inversely proportional to SES. So, all students can easily modify their test scores but high SES students can more easily modify their grades compared to low SES students. Under these assumptions, we estimate the students' raw covariates $Z_{i}.$ We consider a decision maker who can only accept 30\% of the student population.

We consider three different outcome variables. Let $Y_{i, 1}(1)$ be the number of months student $i$ will enrolled in postsecondary school from June 1992 - August 1994 (reported in NELS) if accepted, $Y_{i, 2}(1)$ be a proxy for the raw academic performance of student $i$ given by $\bar{Z}_{i, \text{test scores}}$ if accepted, and $Y_{i, 3}(1)$ is proportional to the inverse of the SES of student $i$ if accepted. Note that $Y_{i, j}(0)=0$ for all $i, j$.
We define $V_{\text{eq}, j}(\beta)$ as the equilibrium policy value when the policy value is defined in terms of $Y_{\cdot, j}$ (Definition \ref{def:eq_loss}). We evaluate how well the methods maximize $V_{\text{eq}, j}(\beta)$ for $j=1,2,3.$

\begin{figure}
    \centering
    \includegraphics[width=\textwidth]{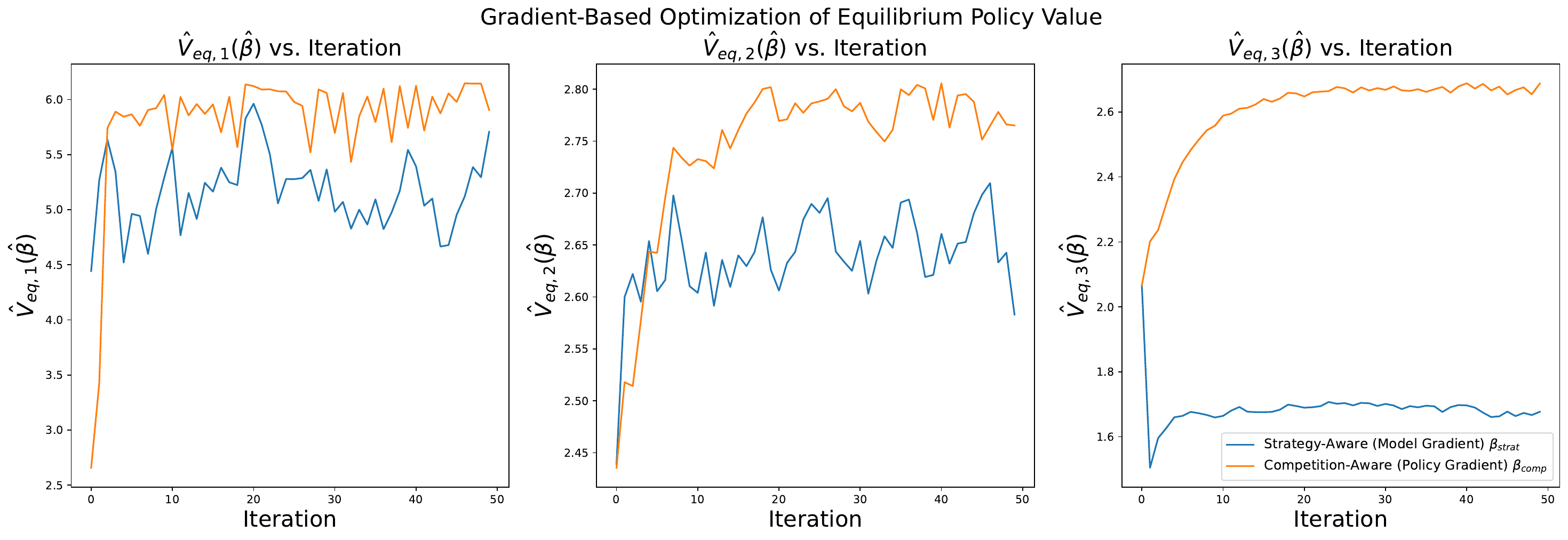}
    \caption{We plot the equilibrium policy value obtained from iterates of strategy-aware and competition-aware methods.}
    \label{fig:nels_learning}
\end{figure}

\begin{table}[t]
    \centering
    \begin{tabular}{c|c|c|c}
         Method &$\hat{V}_{\text{eq}, 1}(\hat{\beta})$  & $\hat{V}_{\text{eq}, 2}(\hat{\beta})$ & $\hat{V}_{\text{eq}, 3}(\hat{\beta})$  \\



\hline
        Capacity-Aware $\beta_{\text{cap}}$ & 3.87 $\pm$ 0.03 & 2.51 $\pm$ 0.15 & 1.62 $\pm$ 0.01

 \\
        Strategy-Aware (Model Gradient) $\beta_{\text{strat}}$ & 5.32 $\pm$ 0.22 & 2.63 $\pm$ 0.04 & 1.70 $\pm$ 0.02

 \\
        Competition-Aware (Policy Gradient) $\beta_{\text{comp}}$ & 5.87 $\pm$ 0.27  & 2.77 $\pm$ 0.03 & 2.38 $\pm$ 0.35
    \end{tabular}
    \caption{The competition-aware method outperforms strategy-aware and capacity-aware baselines. A one-sided paired $t$ test yields a $p$-value of less than 9e-4 for $j=1, 2, 3$. 
    }
    \label{tab:nels}
\end{table}




\begin{figure}
\centering
\includegraphics[width=\textwidth]{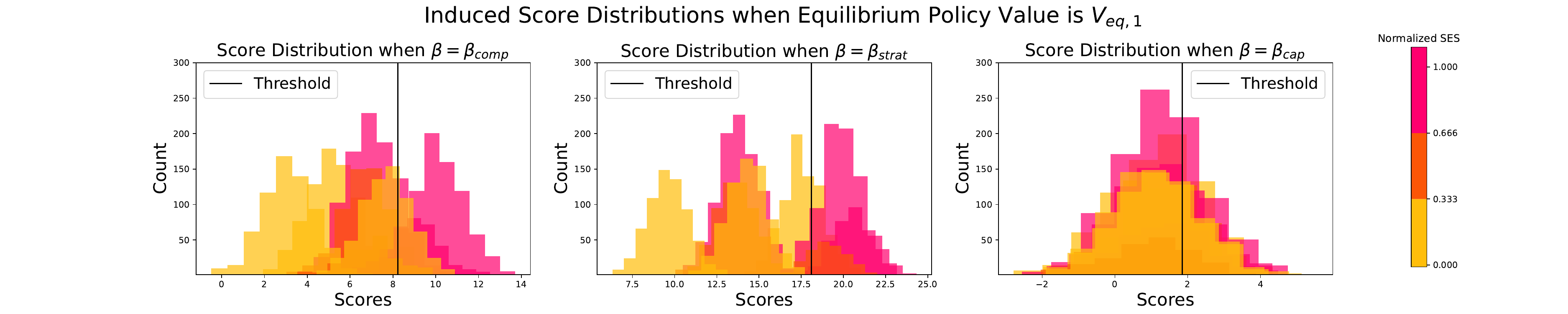}
\includegraphics[width=\textwidth]{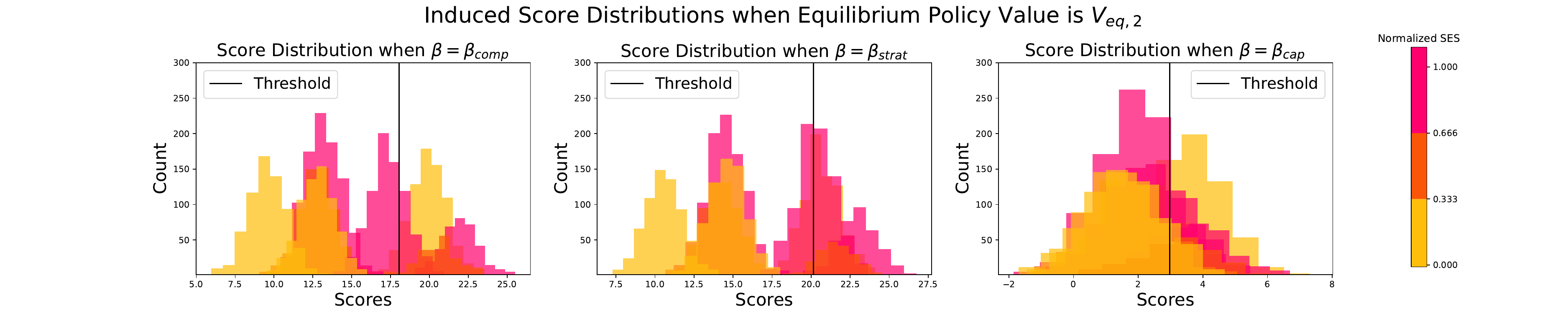}
\includegraphics[width=\textwidth]{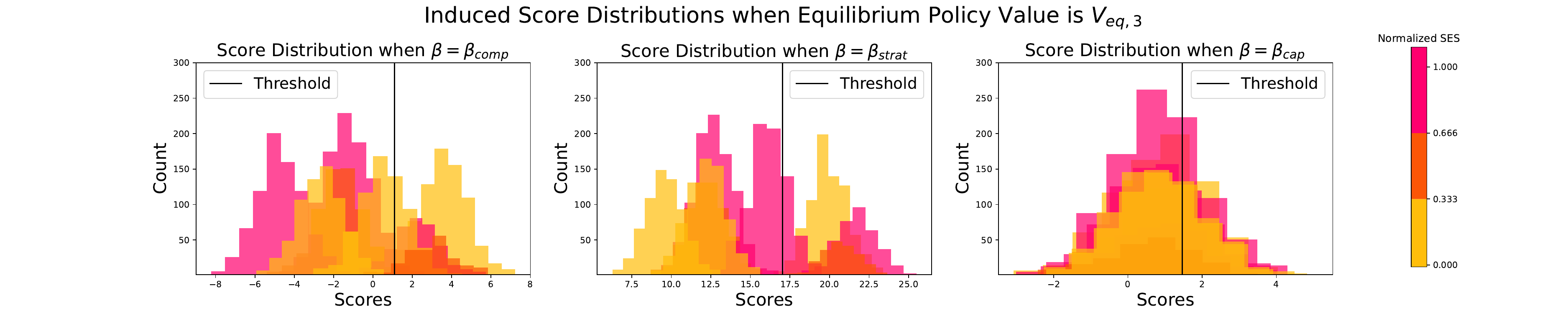}
\caption{We plot the score distributions that are induced by $\beta_{\text{comp}}$, $\beta_{\text{strat}}$, $\beta_{\text{cap}}$ to maximize $V_{\text{eq}, j}(\beta)$ for $j=1, 2, 3$. We plot a histogram of scores for each agent with a distinct unobservable $(Z_{i}, c_{i})$. Agents are color-coded on according to low (yellow), medium (orange), and high (pink) relative SES. The selection criterion $\beta_{\text{comp}}$ accepts students with varying SES.}
\label{fig:outcomes_plot}
\end{figure}


We generate a dataset of $n=14915$ students by sampling from our simulated distribution. For the capacity-aware baseline, we consider an RCT where treatments are assigned to $n$ students in order to estimate the CATE. For the gradient-based methods, we randomly initialize the policy and optimize $\beta$ via projected stochastic gradient descent (in our case, ascent because we aim to maximize the policy value) using Algorithm \ref{alg:method}. We assume that $n$ students are observed by the decision maker at each iteration, and we run the algorithm for 50 iterations. We run 10 random trials, where the randomness is over the initialization and sampled students. 

Across our experiments, the competition-aware policy is the most performant, and the strategy-aware policy is strong baseline (Table \ref{tab:nels}). Figure \ref{fig:nels_learning} visualizes iterates of the gradient-based methods for maximizing $V_{\text{eq},j}(\beta)$ for $j=1, 2, 3.$ We note that the competition-aware method, which relies on the policy gradient estimator, can exhibit higher variance than the strategy-aware method, which relies on the model gradient estimator. While the policy gradient estimator consistently estimates the policy gradient, it targets a more complicated estimand and can have higher variance than the model gradient estimator, which does not consistently estimate the policy gradient. This is analogous to the related literature on estimating treatment effects under interference, where indirect effect estimators often have higher variance than direct effect estimators \citep{li2022random}.

When the objective is $V_{\text{eq}, 2}$, the competition-aware policy only yields a modest improvement over the baselines, despite that the three approaches admit different groups of students (Figure \ref{fig:outcomes_plot}b). We attribute this to the observation that all students have similar values of $Y_{i, 2}(1)$. Among the students, the range of $Y_{i, 2}(1)$ is $5.56$, while the range of $Y_{i, 1}(1)$ is $23.5$, and the range of $Y_{i, 3}(1)$ is $10$. All three methods obtaining similar policy value when optimizing $V_{\text{eq}, 2}$ is an artefact of the fact that all students have $Y_{i, 2}(1)$ that are close in value.

The learned policies for each combination of method and objective are given in Appendix \ref{subsec:more_res}. Qualitatively, we find that when the objective is $V_{\text{eq}, 1}$, the most performant selection criterion favors students with high SES (Figure \ref{fig:outcomes_plot}a, left). We observe that $Y_{i, 1}(1)$ is positively correlated with SES; the correlation between these variables is $0.44.$ When optimizing $V_{\text{eq}, 1}$, we find that $\beta_{\text{comp}}$ places high weight on most of $X_{i, \text{grades}}$ (except foreign language), which are features that students with high SES can manipulate. In contrast, when the objective is $V_{\text{eq}, 2},$ we find that the most performant selection criterion accepts students with varied socioeconomic background (Figure \ref{fig:outcomes_plot}b left), and we find that the correlation between the outcome $Y_{i, 2}(1)$ and SES is $0.07$, so they are not strongly correlated. Finally, we find that when the objective is $V_{\text{eq}, 3},$ the most performant selection criterion accepts students with low socioeconomic background (Figure \ref{fig:outcomes_plot}c, left), and we find that the correlation between $Y_{i, 3}(1)$ and SES is $-0.82,$ a strong negative correlation. The most performant policy for maximizing $V_{\text{eq}, 3}$ places negative or low magnitude weights on $X_{i, \text{grades}}$, which students with high SES can more easily modify.

Here, we have illustrated that our policy gradient estimator can be used to optimize different possible objectives, and the distribution of students that are treated under the different policies will vary depending on the correlation between the outcome of interest with the students' unobservable types $(Z_{i}, c_{i})$. An additional insight from our empirical analysis is that the choice of outcome has a large impact on the learned policy. 
\end{section}

\begin{section}{Discussion}
\label{sec:conclusion}
While there has been considerable recent interest on learning in the presence of strategic behavior,
settings that also include capacity constraints have received less attention to date.
The interplay of strategic behavior and capacity constraints can lead
to effective competition between agents, thus leading to qualitatively different challenges in policy learning relative to the unconstrained setting.
Furthermore, many motivating applications
for learning with strategic behavior, such as college admissions and hiring, are precisely settings where the
decision maker is capacity-constrained---thus highlighting the need for further work in this area.

We adopt a flexible model where agents are heterogenous in their raw covariates and their ability to modify them. Depending on the context, strategic behavior may be harmful, beneficial, or neutral for the decision maker. In some applications, strategic behavior may be a form of ``gaming the system,'' e.g., cheating on exams in the context of college admissions, and the decision maker may not want to assign treatment to agents who have high ability to modify their covariates. In other applications, the decision maker may want to accept such agents because the agents who would benefit the most from the treatment are those who can invest effort to make themselves look desirable. Lastly, as demonstrated by \cite{liu2021strategic}, when all agents have identical ability to modify their covariates, the strategic behavior may be neutral for the decision maker because it does not affect which agents are assigned treatment. Our model permits all of these interpretations because we allow for potential outcomes to be flexibly related to the agent's type.

A number of questions remain open for further research.
First, our model assumes that the decision maker's policy is fixed over time. Considering
dynamic treatment rules, i.e., where the policy is time-varying, would extend this work and
would likely lead to new types of equilibria. Second, we here considered linear policies because
they are relevant to many real-world applications, such as the Chilean college admissions system
\citep{santelices2019institution}; but more flexible policies are possible. Third, we here assumed that agents are myopic.
Future work may consider agents who respond to a history of thresholds $\{ s_{j} \}_{j \in [t-k, t]}$ instead of just $s_{t}$. Many models of agent behavior are possible in this case, e.g., an agent could respond to the mean of the thresholds or use the trend of the history to predict the next threshold and respond to the prediction. These different assumptions may yield different dynamics than the ones that we study.

On the more technical side, one limitation of our model is that it does not permit an agent's post-effort covariates $X_{i}$ to exactly equal their raw covariates $Z_{i}$. For $X_{i}$ to be exactly equal to $Z_{i},$ the cost of covariate modification must be infinite for any deviation from $Z_{i}$. Our work here only considers cost functions that are twice-differentiable and lie in an $L^{2}$-space, but we expect that a more general proof strategy can relax these requirements. Relatedly, our model heavily relies on some form of noise being present in the system. If agents have a noisy understanding of the policy parameters \citep{jagadeesan2021alternative}, or agents best respond imperfectly (as in our work), or exogenous noise affects how decisions are made \citep{kleinberg2018human}, then best responses will be continuous. We expect our results to hold as long as there is sufficient exogenous noise in the system to guarantee that best responses are continuous--the source of the noise itself is not especially crucial. Here, for technical convenience we assume Gaussian noise, but we expect our results to apply to more generic noise distributions, provided they are mean-zero, twice-differentiable, and have bounded second derivative. Possible modifications of our model that could still allow for tractable equilibrium modeling include considering stochastic policies instead of deterministic ones, generic noise distributions, or the noisy response model of \citet{jagadeesan2021alternative}.
On the other hand, if there is no noise in the system, then the agents can strategize perfectly, yielding a discontinuous best response function. In some practical scenarios, such discontinuities are unnatural; see \citet{jagadeesan2021alternative} for a number of examples. Nevertheless, it may still be interesting to develop procedures that perform well despite these discontinuities.

Finally, it would be of considerable interest to study how our proposed approach behaves under misspecification of our model---and
to develop simple verifiable heuristics for assessing whether our proposed method will find a solution that obtains significant
gains over baselines that are oblivious to competition. While our policy gradient estimator is consistent in the absence of competition
and strategic behavior, it may not be the simplest strategy if competition and strategic behavior are not first-order concerns.
Thus, rules of thumb for detecting whether agents are strategic and whether competition effects are material to optimal decision
making would be useful when considering practical application of the method.


\end{section}

\bibliographystyle{plainnat}
\bibliography{ref}

\appendix

\begin{section}{NELS Experiment Details}
\label{sec:exp_details}

\begin{subsection}{Dataset Information}
National Education Longitudinal Study of 1988 (NELS:88) is nationally representative, longitudinal study of eighth graders in 1988. The cohort of students is followed throughout secondary and postsecondary years; the first followup occurs in 1990 (tenth grade for students who continue school), the second followup occurs in 1992 (twelfth grade for students who continue school), and the third occurs in 1994 (postsecondary years). 

In this experiment, we aim to simulate a college admissions process, so we focus on the data collected in 1992 followup, where members of the initial cohort who remained in school are in 12th grade. We use the following publicly-available variables in Table \ref{tab:vars} from the NELS dataset to construct the agent types, the agent covariates, and the agents' outcomes based on whether they are admitted to the college or not.

\begin{table}[ht]
    \centering
    \begin{tabular}{l|l|r|l}
    Variable & Meaning & Imputed Value & Range \\
    \hline 
    F2SES1 & Socio-economic status composite & -0.088 & [-3.243, 2.743]\\
    F22XRSTD & Reading standardized score & 63.81 & [29.01, 68.35] \\
F22XMSTD & Mathematics standardized score & 63.96 & [29.63, 71.37] \\
F22XSSTD & Science standardized score & 64.01  & [29.70, 70.81]  \\
F22XHSTD & History standardized score & 64.30  & [25.35, 70.26] \\
F2RHENG2 & Average grade in English & 7.07 & [1., 13.] \\
F2RHMAG2 & Average grade in mathematics & 7.61 & [1., 13.]\\
F2RHSCG2 & Average grade in science & 7.43 & [1., 13.]\\
F2RHSOG2 & Average grade in social studies & 7.01 & [1., 13.]\\
F2RHFOG2 & Average grade in foreign language & 6.58 & [1., 13.]\\
\multirow{2}{*}{F3ATTEND} & Number of months attended postsecondary  & \multirow{2}{*}{19.21} & \multirow{2}{*}{[1., 27.]}  \\
&  institutions 06/1992-08/1994 & 
\end{tabular}
\caption{NELS:88 variables used in semi-synthetic experiment.}
\label{tab:vars}
\end{table}

\end{subsection}
Note that for the grades, 1. represents the highest grade (A+) and 13. represents the lowest grade. For simplicity, we negate these quantities in our preprocessing step, so that a higher score is more desirable.

\begin{subsection}{Estimating Agent Unobservables from NELS.}
\label{subsec:constr}

\begin{figure}
    \centering
    \includegraphics[width=0.5\textwidth]{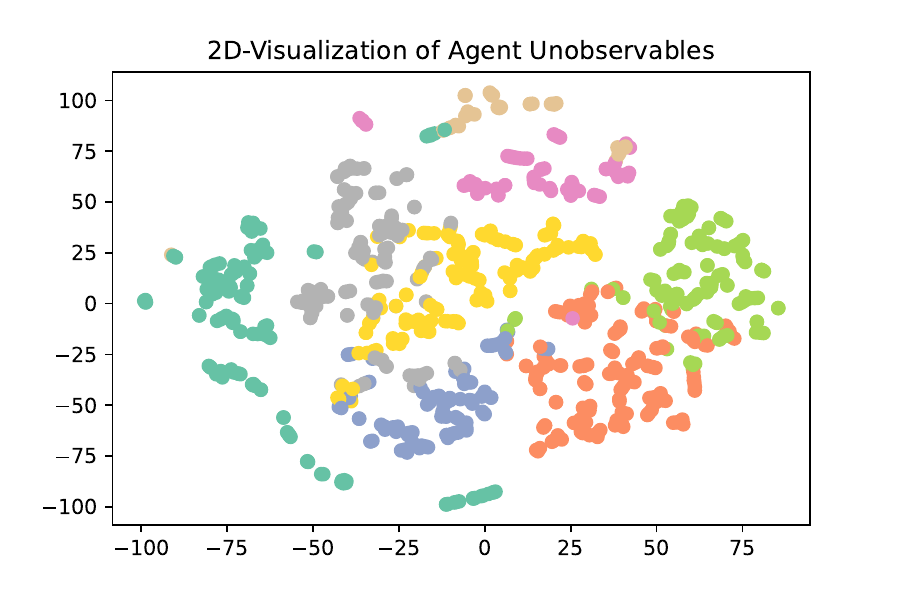}
    \caption{We cluster the agent unobservables computed from the NELS dataset into $K=8$ clusters with $K$-means clustering. Using t-SNE, we visualize two-dimensional embeddings of unobservables $\{( Z_{i}, G_{i}, Y_{i, 1}(1), Y_{i, 2}(1)), Y_{i, 3}(1)\}_{i=1}^{n}$ from the NELS data. Each point represents an two-dimensional embedding of $(Z_{i}, G_{i}, Y_{i, 1}(1), Y_{i, 2}(1), Y_{i, 3}(1))$ and the color of the point corresponds to the cluster that the agent belongs to.}
    \label{fig:agent_type_clusters}
\end{figure}

The student covariates $X_{i}$ consist of twelfth grade standardized test scores in reading, math, science, and history and average grades in English, math, science, social studies, and foreign language. We assume that the reported test scores and grades from the NELS reflect student $i$'s best response $x_{i}^{*}$. So, $x_{i}^{*} \in \mathbb{R}^{d}$ for $d=9$. We use the subscripts ``test scores" and ``grades'' to specify the covariates and unobservables that correspond to test scores and grades, respectively. We assume a quadratic cost of modifying features (see \eqref{eq:quadratic_utility}), where each student's cost function $c_{i}$ is parameterized by
\begin{align*}
\label{eq:nels_gamma}
    G_{i, \text{test scores}} = 0.1,  \quad G_{i, \text{grades}} \propto 1/\gamma_{i},
\end{align*}
where $\gamma_{i}$ is student $i$'s socioeconmic status (SES) percentile reported in NELS. So, all students have the same cost to improving their standardized test scores, but students with high SES can more easily improve their grades compared to students with low SES. In addition, we assume that $x_{i}^{*}$ is the best response to a particular policy with parameters $(\bar{\beta}, \bar{s})$, where $\bar{\beta} = [\frac{1}{\sqrt{d}}, \frac{1}{\sqrt{d}}, \dots \frac{1}{\sqrt{d}}]^{T}, \quad \bar{s}=19.5.$ Lastly, we set the variance of the noise distribution $\sigma^{2}$ is set to ensure that students' best responses are continuous; we set $\sigma=1.20$.
With this information, we can estimate $Z_{i}$ for each student $i$ that are consistent with the NELS data.  Under these assumptions, we can compute each student's raw covariates
 \[Z_{i, j} = x^{*}_{i, j} - \frac{1}{2G_{i, j}} \phi_{\sigma}(\bar{s} - \bar{\beta}^{T}x^{*}_{i}) \bar{\beta}_{j}\]
 for $j \in \{1, 2, \dots d\}$ where $\phi_{\sigma}$ is the Normal PDF with mean 0 and variance $\sigma^{2}.$

To simulate agent best responses to policy parameters, we run Newton's method to compute the best response for each unique choice of agent unobservables and policy parameters because $x_{i}^{*}(\beta, s)$ does not have a closed form. Due to computational constraints of the data generation, we construct a distribution over $K$ representative unobservables instead of the direct approach of using the empirical distribution over all $n$ unobservables. We cluster the dataset $\{(Z_{i}, G_{i}, Y_{i, 1}(1), Y_{i, 2}(1), Y_{i,3}(1))\}_{i=1}^{n}$ into $K=8$ clusters via $K$-means clustering to determine the representative unobservables $\{(Z_{k}, G_{k}, Y_{k, 1}(1), Y_{k, 2}(1), Y_{k, 3}(1))\}_{k=1}^{K}$. A visualization of this clustering is provided in Figure \ref{fig:agent_type_clusters}. After that, we compute the proportion $p_{k}$ of unobservables in each cluster $k$ for $k=1, 2, \dots K.$ Under our constructed distribution, a student with unobservables $(Z_{k}, G_{k}, Y_{k, 1}(1), Y_{k, 2}(1), Y_{k, 3}(1))$, is sampled with probability $p_{k}.$ Note that $Y_{k, 1}(0) =Y_{k, 2}(0) = Y_{k, 3}(0)=0$ for $k =1, 2, \dots K$. 

\end{subsection}

\begin{subsection}{Learning}
We use a learning rate of $a=0.025$ in projected SGD. We use a perturbation size $b=0.025$ for $\beta$ and $b=0.2$ for $s$.
\end{subsection}

\begin{subsection}{Additional Results}
\label{subsec:more_res}
Recall that first four coordinates of $\beta$ are the coefficients of $X_{\text{test scores}}$ and the last five are the coefficients of $X_{\text{grades}}.$ We examine the learned policies from a single trial of our semi-synthetic experiment. When optimizing $V_{\text{eq}, 1}(\beta)$, we find that
\begin{enumerate}
	\item $\beta_{\text{cap}} = [-0.003, -0.123, 0.099, 0.038, -0.584, -0.23, -0.067, 0.748, 0.124]^{T}.
$
	\item $\beta_{\text{strat}} = [0.295, 0.217, 0.202, 0.285, 0.227, 0.385, 0.438, 0.454, 0.383]^{T}
.$
	\item $\beta_{\text{comp}} = [0.364, -0.095, -0.159, 0.011, 0.218, 0.142, 0.56, 0.609, -0.284]^{T}
.$
\end{enumerate}
We find that the most performant policy $\hat{\beta}_{\text{comp}}$ places high/positive weight on most grades and close to zero weight on most test scores. Interestingly, it places negative weight on the last grade (foreign language).

When optimizing $V_{\text{eq}, 2}(\beta)$, we find that
\begin{enumerate}
\item $\beta_{\text{cap}} = [-0.535, 0.077, 0.024, 0.706, -0.189, 0.311, -0.205, 0.163, -0.085]^{T}.$
\item $\beta_{\text{strat}} = [0.347, 0.35, 0.368, 0.394, 0.345, 0.294, 0.32, 0.271, 0.289]^{T}
.$
\item $\beta_{\text{comp}} = [0.384, 0.409, 0.397, 0.485, 0.265, 0.12, 0.314, 0.327, -0.052]^{T}.$
\end{enumerate}
We find that the most performant policy $\hat{\beta}_{\text{comp}}$ place similar weights on all grades and test scores.

When optimizing $V_{\text{eq}, 3}(\beta)$, we find that
\begin{enumerate}
\item $\beta_{\text{cap}} = [-0.0, 0.107, -0.006, -0.112, 0.644, 0.127, 0.06, -0.731, -0.085]^{T}.$
\item $\beta_{\text{strat}} = [0.391, 0.319, 0.505, 0.554, 0.343, 0.09, 0.153, 0.184, -0.01]^{T}
.$
\item $\beta_{\text{comp}} = [0.315, 0.13, 0.121, 0.361, -0.058, -0.618, 0.027, -0.154, -0.573]^{T}.$
\end{enumerate}
We find that the most performant policy $\beta_{\text{comp}}$ places close to positive weight on the test scores and negative weights on most grades.

In Figure \ref{fig:mag}, we also examine the magnitude of the equilibrium gradient, model gradient, and policy gradient, while optimizing the equilibrium policy value using the policy gradient.  We find that the magnitude of the policy gradient is much smaller than the magnitude of the model gradient or equilibrium gradient, so the components of the gradients often have components with opposing signs. 

\begin{figure}
\includegraphics[width=\textwidth]{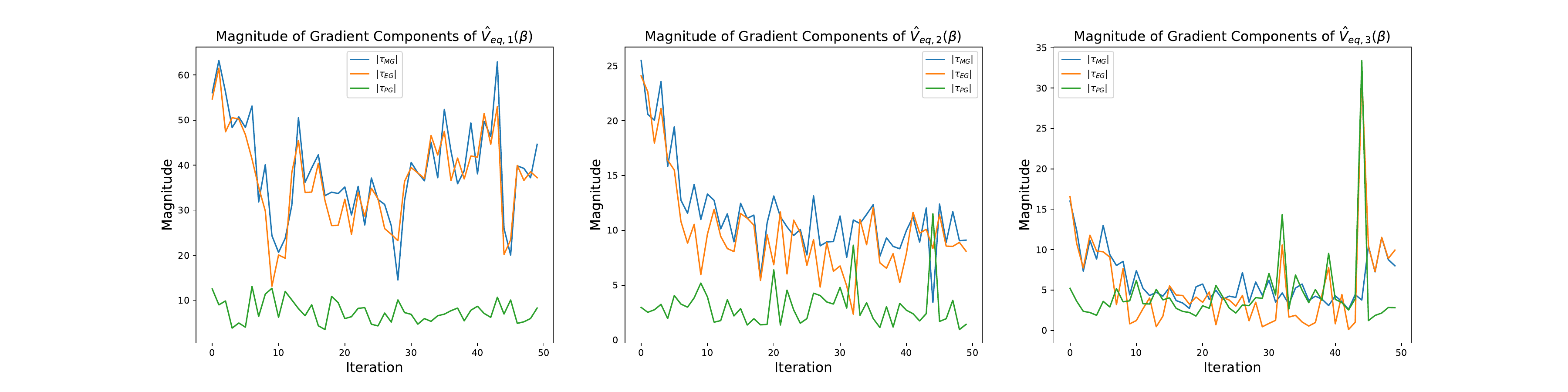}
\caption{The magnitude of the equilibrium and model gradients are close across iterations but the magnitude of the policy gradient is much smaller.}
\label{fig:mag}
\end{figure}

\end{subsection}
\end{section}

\begin{section}{Additional Experiments}
\label{sec:add_exp}
\begin{subsection}{Toy Example}
\label{subsec:toy}
\begin{figure}[t]
    \centering
    \includegraphics[width=\textwidth]{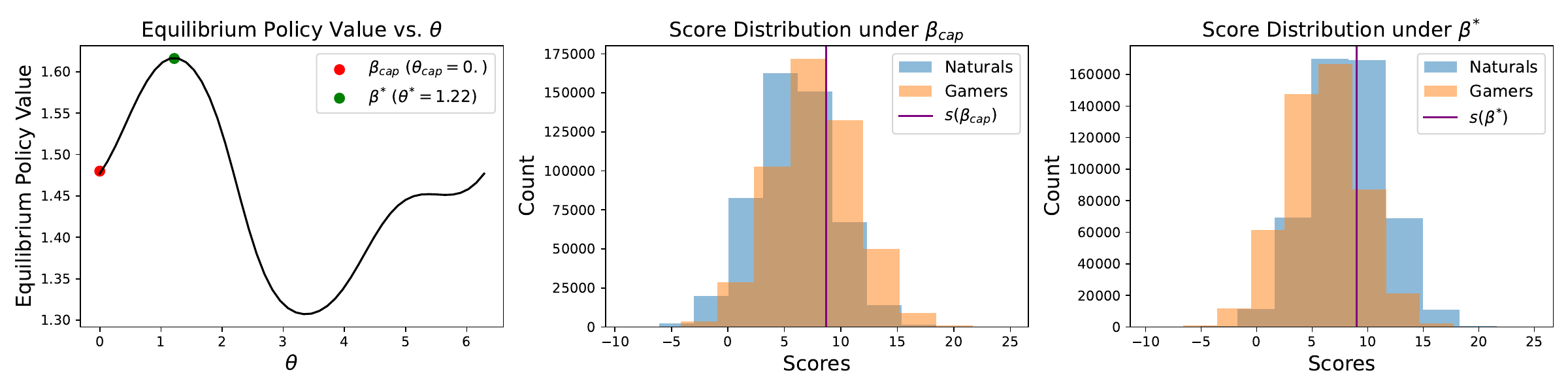}
    \caption{\textbf{Left}: We plot the expected equilibrium policy value across $\theta = \arctan(\beta_{1}/ \beta_{0}).$ Deploying $\beta_{\text{cap}} \text{ (equivalently, }\theta_{\text{cap}} = 0.)$ does not yield an optimal policy value. We note that the equilibrium policy value has a global minimum $\beta^{*} = [0.345, 0.938]^{T} (\text{equivalently, }\theta^{*} =1.22)$. \textbf{Middle}: When the selection criterion is $\beta_{\text{cap}}$, the naturals make up only 35\% of agents who score above the threshold. \textbf{Right}: When the selection criterion is $\beta^{*},$ the naturals make up 69\% of agents who score above the threshold.}
    \label{fig:cross_type}
\end{figure}

For a one-dimensional example, we consider policies with the parametrization $\beta = [\cos \theta, \sin \theta]^{T}, \text{ where } \theta \in [0, 2\pi).$ We suppose that the capacity constraint limits the decision maker to accept only 30\% of the agent population. We define $Y_{i}(1) = Z_{i, 1}$ and $Y_{i}(0) = 0.$ The decision maker's equilibrium policy value $V_{\text{eq}}(\beta)$ is given by Definition \ref{def:eq_loss}.

We consider an agent distribution where agents are heterogeneous in their raw covariates and ability to modify their observed covariates and optimize the quadratic utility function \eqref{eq:quadratic_utility}. So, the entry $G_{i, j}$ quantifies the cost to agent $i$ of deviating from $Z_{i, j}$ in their reported covariate $X_{i, j}$. We suppose that $\mathcal{X} = [0, 10]^{2}$ and $G_{i} \in [0.01, 20]^{2}.$ Motivated by \cite{frankel2019muddled}, we consider an agent distribution with two groups of agents in the population of equal proportion, \textit{naturals} and \textit{gamers}. The naturals have $Z_{i, 1}, Z_{i, 2} \sim \text{Uniform}[5, 7]$ and $G_{i, 1}, G_{i, 2} \sim \text{Uniform}[10, 20].$ In contrast, the gamers have $Z_{i, 1}, Z_{i, 2} \sim \text{Uniform}[3, 5], G_{i, 1} \sim \text{Uniform}[0.01, 0.02],$ and $G_{i, 2} \sim \text{Uniform}[10, 20].$ We suppose $\text{supp}(F_{Z, c})$ has 5 naturals and 5 gamers. The variance of the noise distribution $\sigma^{2}$ is set to ensure continuity of agent best responses; we set $\sigma=3.30$.

Note that naturals have higher values of $Z_{i, 1}, Z_{i, 2}$ compared to gamers, and gamers have lower cost to modifying $Z_{i, 1}$ compared to naturals. Accepting a natural yields a higher policy value compared to a gamer because naturals have higher $Z_{i, 1}$.

If the decision maker ignores the presence of strategic behavior, they may use the first baseline of capacity-aware policy learning, estimating the criterion $\beta_{\text{cap}}$ using data from an RCT. Since agents are not strategic in the RCT, $\beta_{\text{cap}}$ likely places substantial weight on the first covariate. However, this criterion yields suboptimal policy value at deployment because gamers have high ability to deviate from $Z_{i, 1}$ when reporting $X_{i, 1}.$ Intuitively, there should exist a better criterion. We note that all agents are relatively homogenous in their ability to deviate from $Z_{i, 2}$ when reporting $X_{i, 2}$ because $G_{i, 2} \sim \text{Uniform}[10, 20]$ for all agents. At the same time, $Z_{i, 2}$ is correlated with $Z_{i, 1}$. So, a criterion that places high weight on the second covariate should yield higher policy value by accepting more naturals.

We plot the equilibrium policy value of decision maker as a function of the polar-coordinate representaiton of the criterion $\theta = \arctan(\beta_{1}/\beta_{0})$ (Figure \ref{fig:cross_type}, left plot). As expected, deploying $\beta_{\text{cap}}$, which corresponds to $\theta_{\text{cap}}$, is suboptimal for maximizing the equilibrium policy value. Under $\beta_{\text{cap}}$, only 35\% of agents who score above the threshold for receiving treatment under $\beta_{\text{cap}}$ are naturals (Figure \ref{fig:cross_type}, middle plot). The policy $\beta^{*} = [0.345, 0.938]^{T}$ achieves the optimal equilibrium policy value, and as expected, it places considerable weight on the second covariate. When $\beta = \beta^{*}, $ we observe that  69\% of agents who score above the threshold are naturals (Figure \ref{fig:cross_type}, right plot). 

 We evaluate the solutions learned by the two baselines and our proposed approach. For the capacity-aware baseline, we learn $\theta_{\text{cap}}$ using data from an RCT where treatment is allocated randomly among $n=1000000$ agents. For the gradient-based methods, we run stochastic gradient descent (in our case, ascent to maximize the policy value) on $\theta$, the polar-coordinate representation of $\beta$, initialized at $\theta=0$ for 100 iterations. We assume that $n=1000000$ agents are observed by the decision maker at each iteration. We use a learning rate of $a=0.5$ in vanilla SGD with for the competition-aware policy gradient approach. We use a learning rate of $a=0.25$ in vanilla SGD with strategy-aware model gradient approach. We use a perturbation size $b=0.025$ for $\beta$ and $b=0.2$ for $s$.

Across 10 random trials (where the randomness is over the sampled unobservable values and sampled agents), we observe that $\theta_{\text{comp}}=\theta^{*}$, achieving optimal equilibrium policy value (Table \ref{tab:toy}). This demonstrates that accounting for competition is beneficial. Meanwhile, $\theta_{\text{cap}}$ and $\theta_{\text{strat}}$ obtain suboptimal equilibrium policy value (Table \ref{tab:toy}). Nevertheless, strategy-aware gradient-based optimization is a relatively strong baseline because it accounts for the impact of the strategic behavior due to agents' knowledge of the policy on the policy value. 

\begin{table}[t]
    \centering
    \begin{tabular}{c|c|c}
        Method & $|V_{\text{eq}}(\hat{\beta}) - V_{\text{eq}}(\beta^{*})|$ \\
        \hline 
        Capacity-Aware $(\theta_{\text{cap}})$ & 0.19 $\pm$ 0.04   \\
        Strategy-Aware $(\theta_{\text{strat}})$ & 0.04 $\pm$ 0.05 \\
        Competition-Aware $(\theta_{\text{comp}})$ & 0.00 $\pm$ 0.00  \\
    \end{tabular}
    \caption{In our toy experiment, we observe that $\theta_{\text{comp}}$ converges to the optimal policy $\theta^{*}$. However, $\theta_{\text{strat}}$ and $\theta_{\text{cap}}$ are suboptimal.}
    \label{tab:toy}
\end{table}

\end{subsection}

\begin{subsection}{High-Dimensional Simulation}
\label{subsec:exp_high_dim}

\begin{table}[t]
    \centering
    \begin{tabular}{c|c}
        Method & Equilibrium Policy Value $\hat{V}_{\text{eq}}(\hat{\beta})$  \\
        \hline 
        Capacity-Aware $(\beta_{\text{cap}})$ & 5.832 $\pm$ 0.14
 \\
        Strategy-Aware $(\beta_{\text{strat}})$ & 6.119 $\pm$ 0.129
\\
        Competition-Aware $(\beta_{\text{comp}})$ & 6.151 $\pm$ 0.14
 \\
    \end{tabular}
    \caption{Across 10 random trials, we find that competition-aware gradient-based optimization outperforms the other baselines $(d=10).$ A one-sided paired $t$-test, where we compare the final policy value of $\beta_{\text{comp}}$ and the final policy value of $\beta_{\text{strat}}$ with the same random seed, yields a $p$-value of 7e-4.
    }

    \label{tab:high_dim}
\end{table}
We evaluate the solutions learned by the two baselines and our proposed approach. For $d=10,$ we consider policies $\beta \in \mathcal{B} = \mathbb{S}^{d-1}$. We suppose the capacity constraint only allows the decision maker to accept 30\% of the agent population. We define $Y_{i}(1) = \sum_{j=1}^{d/2} Z_{i, j}$ and $Y_{i}(0)=0.$ The decision maker's equilibrium policy value $V_{\text{eq}}(\beta)$ is given by Definition \ref{def:eq_loss}. Agents optimize , and $G_{i, j}$ quantifies the cost of modifying agent $i$'s $j$-th covariate $X_{i, j}$ from $Z_{i, j}$. We suppose that $F$ is supported on 10 points $(Z_{i}, c_{i})$. We consider agents with a quadratic utility function \eqref{eq:quadratic_utility} where $G_{i}$ specifies the cost of covariate modification. We again consider gamers and naturals. The naturals have $Z_{i, j} \sim \text{Uniform}[5, 7]$ and $G_{i, j} \sim \text{Uniform}[1, 2]$ for  $j \in \{1, \dots d\}.$ The gamers have $Z_{i, j} \sim \text{Uniform}[3, 5]$ and $G_{i, j} \sim \text{Uniform}[0.1, 0.2]$ for $j \in \{1, \dots d/2\}$ and $G_{i, j} \sim \text{Uniform}[1., 2.]$ for $j \in \{ d/2 + 1, \dots d\}.$ The variance of the noise distribution $\sigma^{2}$ is set to ensure the continuity of agent best responses; we set $\sigma=1.10$.

We compare the learned solutions of the three methods. For the capacity-aware baseline, we learn $\theta_{\text{cap}}$ using data from an RCT where treatment is allocated randomly among $n=1000000$ agents. For the strategy-aware and competition-aware methods, we optimize $\beta$ via projected stochastic gradient descent (in our case, ascent because we aim to maximize the policy value), initialized at $\beta = [1., 1., \dots 1.]/\sqrt{d}$ for 100 iterations. We assume that $n=1000000$ agents are observed by the decision maker at each iteration. We use a learning rate of $a=0.5$ in projected SGD for all three baselines. For the gradient-based approaches, we use a perturbation size $b=0.025$ for $\beta$ and $b=0.2$ for $s$.

Across 10 random trials (where the randomness is over the sampled unobservable values and the sampled agents), we observe that $\beta_{\text{comp}}$ finds a solution with higher equilibrium policy value than $\beta_{\text{strat}}$ and $\beta_{\text{cap}}$ (Table \ref{tab:high_dim}). 
\end{subsection}

\end{section}

\begin{section}{Algorithms}
\label{sec:alg}
\begin{algorithm}{}
\SetAlgoLined

Run OLS of $Y^{j}$ on $\mathbf{M}_{\beta}^{j}$: $\Gamma_{Y, \beta}^{j} \leftarrow ((\mathbf{M}_{\beta}^{j})^{T}\mathbf{M}_{\beta}^{j})^{-1}(\mathbf{M}_{\beta}^{j})^{T} Y^{j})$ \;
Run OLS of $Y^{j}$ on $\mathbf{M}_{s}^{j}$: $\Gamma_{Y, s, r}^{j} \leftarrow ((\mathbf{M}_{s}^{j})^{T}\mathbf{M}_{s}^{j})^{-1}((\mathbf{M}_{s}^{j})^{T} Y^{j})$ \;

Run OLS of $I^{j}$ on $\mathbf{M}_{\beta}^{j}$: $\Gamma_{\Pi, \beta }^{j} \leftarrow ((\mathbf{M}_{\beta}^{j})^{T}\mathbf{M}_{\beta}^{j})^{-1}((\mathbf{M}_{\beta}^{j})^{T} I^{j})$ \;

Run OLS of $I^{j}$ on $\mathbf{M}_{s}^{j}$: $\Gamma_{\Pi, s }^{j} \leftarrow ((\mathbf{M}_{s}^{j})^{T}\mathbf{M}_{s}^{j})^{-1}((\mathbf{M}_{s}^{j})^{T}I^{j})$ \;

$\Gamma_{s, \beta}^{j} \leftarrow  \frac{1}{\rho^{j} - \Gamma_{\Pi, s}^{j}} \cdot \Gamma_{\Pi, \beta}^{j}.$

$\Gamma^{j} \leftarrow \Gamma_{Y, \beta}^{j} + \Gamma_{Y, s, r}^{j} \cdot \Gamma_{s, \beta}^{j} $

\caption{Construct gradient estimates}
\label{algorithm:gradient_estimates}
\end{algorithm}

\begin{algorithm}{}
\SetAlgoLined
\While{$j \leq J$}{
Decision maker deploys $\beta^{j}$ \;
Equilibrium threshold $s^{j}$ arises. \\
\For {$i \in \{1 \dots n\}$} {
Sample random perturbation $\zeta_{i} \sim R^{d}$ and  $\xi_{i} \sim R$\;
$\beta_{i}^{j} \leftarrow \beta^{j} + b_{n}\zeta_{i}$\; 
$s_{i}^{j} \leftarrow s^{j} + b_{n} \xi_{i}$ \;
Agent $i$ best responds to $\beta_{i}^{j}, s_{i}^{j}$ \;
Decision maker observes best response $x_{i}^{j}$ \;
}

Compute $q^{j}$, $q$-th quantile of scores $\{\beta_{i}^{j}x_{i}^{j} - b_{n}\xi_{i}^{j} \}_{i=1}^{n}$ \;
Compute $\rho^{j}$, density of scores at $\rho^{j}$\;

\For{$i \in \{1 \dots n\}$}{
Observes outcome $Y_{i}^{j}$ and measures \\
$I_{i}^{j} \leftarrow \mathbb{I}((\beta_{i}^{j})^{T} x_{i}^{j} > s^{j})$\;
}

$\mathbf{M}_{\beta}^{j} \leftarrow b_{n}\zeta^{j}$ is the $n \times d$ matrix of perturbations $\zeta$\;
$\mathbf{M}_{s}^{j} \leftarrow b_{n} \xi^{j}$ is the $n \times 1$ matrix of perturbations $\xi$\;
$Y^{j}$ is the $n$-length vector of outcomes $Y_{i}^{j}$ \;
$I^{j}$ is the $n$-length vector of indicators $I_{i}^{j}$\;

Construct gradient estimate $\Gamma^{j}$ from $\mathbf{M}_{\beta}^{j}, \mathbf{M}_{s}^{j}, Y^{j}, I^{j}, \rho^{j}$ (See Algorithm \ref{algorithm:gradient_estimates})\;

Take a projected gradient step\\
$\beta^{j+1} \leftarrow \text{Proj}_{\mathcal{B}}( \beta^{j} + a \cdot  \Gamma^{j})$ \;
}
\caption{Learning with Policy Gradient Estimator}
\label{alg:method}
\end{algorithm}
\FloatBarrier
\end{section}

\begin{section}{Standard Results}
\label{sec:standard_results}



\begin{lemm}
\label{lemm:unique_maximizer}
Let $\mathcal{X} \subset \mathbb{R}^{d}$ is a convex set. Let $f: \mathcal{X} \rightarrow \mathbb{R}$ be a strictly concave function. If $f$ has a global maximizer, then the maximizer is unique (\cite{boyd2004convex}).
\end{lemm}


\begin{lemm}
\label{lemm:maximizer_characterization}
Let $f : \mathcal{X} \rightarrow \mathbb{R}$, where $\mathcal{X} \subset \mathbb{R}^{d}$ is a convex set, be a twice-differentiable function. If $f$ is a strictly concave function and $x^{*}$ is in the interior of $\mathcal{X}$, then $x^{*}$ is the unique global maximizer of $f$ on $\mathcal{X}$ if and only if $\nabla f(x^{*}) = 0$ (\cite{boyd2004convex}).
\end{lemm}

\begin{theo}[Implicit Function Theorem]
\label{theo:implicit_function_theorem}
Suppose $\mathbf{f}: \mathbb{R}^{n} \times \mathbb{R}^{m} \rightarrow \mathbb{R}^{m}$ is continuously differentiable in an open set containing $(x_{0}, y_{0})$ and $\mathbf{f}(x_{0}, y_{0}) = 0.$ Let $\mathbf{M}$ be the $m \times m$ matrix \[D_{n+j}\mathbf{f}^{i}(x, y) \quad 1 \leq i, j \leq m.\]
If $\text{det}(\mathbf{M}) \neq 0,$ then there is an open set $X \subset \mathbb{R}^{n}$ containing $x_{0}$ and an open set $Y \subset \mathbb{R}^{m}$ containing $y_{0}$, with the following property: for each $x \in X$ there is a unique $\mathbf{g}(x) \in Y$ such that $\mathbf{f}(x, \mathbf{g}(x)) = 0.$ The function $g$ is continuously differentiable.
\end{theo}

\begin{theo}[Sherman-Morrison Formula]
\label{theo:sherman}
Suppose $\mathbf{A} \in \mathbb{R}^{d \times d}$ is an invertible square matrix, $\mathbf{u}, \mathbf{v} \in \mathbb{R}^{d}$. Then $\mathbf{A} + \mathbf{u}\mathbf{v}^{T}$ is invertible iff $1 + \mathbf{v}^{T} \mathbf{A}^{-1} \mathbf{u} \neq 0.$ In this case,

\[(\mathbf{A} + \mathbf{uv}^{T})^{-1} = \mathbf{A}^{-1} - \frac{\mathbf{A}^{-1} \mathbf{uv}^{T}\mathbf{A}^{-1}}{1 + \mathbf{v}^{T} \mathbf{A}^{-1} \mathbf{u}}.\]
\end{theo}

\begin{lemm}
\label{lemm:pd_difference}
Suppose $\mathbf{A}, \mathbf{B} \in \mathbb{R}^{d \times d}$ are positive definite matrices. If $\mathbf{A} - \mathbf{B}$ is positive semidefinite, then $\mathbf{B}^{-1} - \mathbf{A}^{-1}$ is positive semidefinite (\cite{dhrymes1978mathematics}).
\end{lemm}



\begin{lemm}
\label{lemm:contraction_derivative}
Let $f: \mathbb{R} \rightarrow \mathbb{R}$ be a differentiable real function. The function $f$ is a contraction with modulus $\kappa \in (0, 1)$ if and only if $|f'(x)| \leq \kappa$ for all $x \in \mathbb{R}$ (\cite{ortega1990numerical}).
\end{lemm}

\begin{theo}[Banach's Fixed-Point Theorem]
\label{theo:banach_fixed_point}
 Let $(X,d)$ be a non-empty complete metric space with a contraction mapping $T:X \rightarrow X$. Then $T$ admits a unique fixed-point $x^{*}$ such that $T(x^{*}) = x^{*}.$ Furthermore, for any number $x_{0} \in X$, the sequence defined by $x_{n} = T(x_{n-1}), n \geq 1$ converges to the unique fixed point $x^{*}$.
\end{theo}


\begin{theo}
\label{theo:concentration_quantiles}
Let $X_{1}, X_{2}, \dots, X_{n}$ be i.i.d. random variables from a CDF $F$. Let $\theta_{p}$ be the $p$-th quantile of $F$ and let $\hat{\theta}_{p}$ be the $p$-th quantile of $F_{n}.$ Suppose $F$ satisfies $p < F(\theta_{p} + \epsilon)$ for any $\epsilon > 0.$ Then for every $\epsilon > 0,$ then 
\[ P(|\hat{\theta}_{p} - \theta_{p}| > \epsilon ) \leq 4e^{-2nM_{\epsilon}^{2}}, \]
where $M_{\epsilon}=\min\{F(\theta_{p} + \epsilon ) - p, p - F(\theta_{p}- \epsilon)\}$ (Theorem 5.9, \cite{shao2003mathematical}).
\end{theo}

\begin{theo}[Bernoulli's Inequality]
\label{theo:bernoulli_inequality}
For every $r \geq 0$ and $x \geq -1$, $(1 + x)^{r} \geq 1 + rx.$
\end{theo}

\begin{lemm}
\label{lemm:uniform_convergence_sum}
If the $u_{i}$ i.i.d., $\Theta$ is compact, $a(\cdot, \theta)$ is continuous at each $\theta \in \Theta$ with probability
one, and there is $d(u)$ with $||a(u,\theta) || \leq d(u)$ for all $\theta \in \Theta$ and $\EE{d(u)} < \infty$, then
$\EE{a(u, \theta)}$ is continuous and \[\sup_{\theta \in \Theta} \Big| \frac{1}{n} \sum_{i=1}^{n} a(u_{i},\theta) - \EE{a(u, \theta)} \Big| \xrightarrow{p} 0\]
(Lemma 2.4, \cite{newey1994large}).
\end{lemm}

\begin{lemm}
\label{lemm:uniform_convergence_implies_stochastic_equicontinuity}
Suppose $\Theta$ is compact and $f(\theta)$ is continuous. Then $\sup_{\theta \in \Theta} |\hat{f}_{n}(\theta) - f(\theta)| \rightarrow 0$ if and only if $\hat{f}_{n}(\theta) \xrightarrow{p} f(\theta)$ for all $\theta \in \Theta$ and $\{\hat{f}_{n}(\theta)\}$ is stochastically equicontinuous (Lemma 2.8, \cite{newey1994large}).
\end{lemm}

\begin{lemm}
\label{lemm:stochastic_equicontinuity_convergence_in_prob}
Suppose $\{Z_{n}(t)\}$ is a collection of stochastic processes indexed by $t \in \mathcal{T}.$ Suppose $\{Z_{n}(t)\}$ is stochastically equicontinuous at $t_{0} \in \mathcal{T}.$ Let $\tau_{n}$ be a sequence of random elements of $\mathcal{T}$ known to satisfy $\tau_{n} \xrightarrow{p} t_{0}$. It follows that $Z_{n}(\tau_{n}) - Z_{n}(t_{0}) \xrightarrow{p}0,$ (\cite{pollard2012convergence}).
\end{lemm}

\begin{lemm}
\label{lemm:uniform_convergence_monotonic}
Let $f_{n}: \mathcal{X} \rightarrow \mathbb{R}$ where $\mathcal{X} \subset \mathbb{R}$ is a compact set. Let $\{f_{n}\}$ be a sequence of continuous, monotonic functions that converge pointwise to a continuous function $f$. Then $f_{n} \rightarrow f$ uniformly (\cite{buchanan1908note}).
\end{lemm}

\begin{lemm}
\label{lemm:uniform_convergence_concave}
Let $f_{n}: \mathcal{X} \rightarrow \mathbb{R}$ where $\mathcal{X} \subset \mathbb{R}^{d}$ is a compact set. Let $\{f_{n}\}$ be a sequence of continuous, concave functions that converge pointwise to $f$. Furthermore, assume that $f$ is continuous. Then $f_{n} \rightarrow f$ uniformly (\cite{rockafellar1970convex}).
\end{lemm}

\begin{lemm}
\label{lemm:uniform_convergence_root}
Let $f_{n}: \mathcal{X} \rightarrow \mathbb{R}$ where $\mathcal{X} \subset \mathbb{R}$ is a compact set. Let $\{f_{n}\}$ be a sequence of continuous functions that converge uniformly to $f$. Suppose each $f_{n}$ has exactly one root $x_{n} \in \mathcal{X}$ and $f$ has exactly one root $x^{*} \in \mathcal{X}$. Then $x_{n} \rightarrow x^{*}.$ \hyperref[subsec:uniform_convergence_root]{Proof in Appendix \ref{subsec:uniform_convergence_root}}.
\end{lemm}

\begin{lemm}
\label{lemm:uniform_convergence_maximizer}
Let $f_{n}: \mathcal{X} \rightarrow \mathbb{R}$ where $\mathcal{X}\subset \mathbb{R}^{d}$ is a compact set. Let $\{f_{n}\}$ be a sequence of continuous functions that converge uniformly to $f$. Suppose each $f_{n}$ has exactly one maximizer $x_{n} \in \mathcal{X}$ and $f$ has exactly one maximizer $x^{*} \in \mathcal{X}.$ Then $x_{n} \rightarrow x^{*}.$
\hyperref[subsec:uniform_convergence_maximizer]{Proof in Appendix \ref{subsec:uniform_convergence_maximizer}}.
\end{lemm}

\begin{theo}
\label{theo:parzen_kernel_density}
Let us assume the following: \begin{enumerate}
    \item $K$ vanishes at infinity, and $\int_{-\infty}^{\infty} K^{2}(x) dx < \infty,$
    \item $h_{n} \rightarrow 0$ as $n \rightarrow \infty,$
    \item $nh_{n} \rightarrow \infty$ as $n \rightarrow \infty$.
\end{enumerate}
Let $f^{n}(x)$ be a kernel density estimate of the density function $f$ with $n$ samples, kernel $K$, and bandwidth $h_{n}.$ Then $f^{n}(x) \xrightarrow{p} f(x),$ as $n \rightarrow \infty$ (\cite{parzen1962estimation}).
\end{theo}


\end{section}

\begin{section}{Proofs of Agent Results}
\label{sec:agent_proofs}
We state technical lemmas that will be used in the proof of Proposition \ref{prop:br_all} and Theorem \ref{theo:equilibrium}.

\begin{lemm}
\label{lemm:diff_concavity_utility}
Under Assumption \ref{assumption:strong_convexity}, $\EE[\epsilon]{U_{i}(x; \beta, s)}$ is twice continuously differentiable in $x, \beta, s$. Moreover, if $\sigma^{2} > \frac{1}{\alpha_{i}\sqrt{2\pi e}}$, then $\EE[\epsilon]{U_{i}(x; \beta, s)}$ is strictly concave in $x$. Finally, if additionally $x \in \text{Int}(\mathcal{X})$, then $x = x^{*}_{i}(\beta, s)$ if and only if $\nabla_{x} \EE[\epsilon] {U_{i}(x; \beta, s)} = 0.$
\hyperref[subsec:diff_concavity_utility]{Proof in Appendix \ref{subsec:diff_concavity_utility}}.
\end{lemm}

\begin{lemm}
\label{lemm:score_gradient}
Let $\phi_{\sigma}$ denote the Normal p.d.f. with mean 0 and variance $\sigma^{2}$. Under Assumption \ref{assumption:strong_convexity}, if $\sigma^{2} > \frac{1}{\alpha_{i}\sqrt{2\pi e}}$ and $x_{i}^{*}(\beta, s) \in \text{Int}(\mathcal{X})$, then 
\begin{equation} 
\label{eq:score_gradient}
\nabla_{s} \omega_{i}(s; \beta)  =  \frac{\phi'_{\sigma}(s - \omega_{i}(s; \beta)) \beta^{T} \bfH^{-1} \beta}{1 + \phi'_{\sigma}(s - \omega_{i}(s; \beta)) \beta^{T} \bfH^{-1} \beta},
\end{equation}
where $\bfH:= \nabla^{2}c_{i}(x_{i}^{*}(\beta, s) - Z_{i}).$ Moreover, the function $h_{i}(s; \beta) := s - \omega_{i}(s; \beta)$ is strictly increasing in $s$. \hyperref[subsec:score_gradient]{Proof in Appendix \ref{subsec:score_gradient}}.
\end{lemm}


\begin{lemm}
\label{lemm:limits_br}
Under Assumption \ref{assumption:strong_convexity}, if $\sigma^{2} > \frac{1}{\alpha_{i}\sqrt{2\pi e}}$ and $x_{i}^{*}(\beta, s) \in \text{Int}(\mathcal{X}),$
\begin{align}
    \lim_{s \rightarrow \infty} \omega_{i}(s; \beta ) &= \beta^{T}Z_{i}. \label{eq:up} \\
    \lim_{s \rightarrow -\infty} \omega_{i}(s; \beta) &= \beta^{T}Z_{i}. \label{eq:down}.
\end{align}
Moreover, $\omega_{i}(s; \beta)$ is maximized at a point $s^{*}$, $\omega_{i}(s; \beta)$ is increasing when $s < s^{*}$ and is decreasing on $s > s^{*}.$
\hyperref[subsec:limits_br]{Proof in Appendix \ref{subsec:limits_br}}.
\end{lemm}

\begin{subsection}{Proof of Proposition \ref{prop:br_all}}
\label{subsec:br_all}

This proof has three parts. First, we establish existence and uniqueness of the agent best response $x_{i}^{*}(\beta, s).$ Second, we demonstrate that $x_{i}^{*}(\beta, s)$ is continuously differentiable in $(\beta, s)$. Third, we establish the contraction property of $\omega_{i}(s; \beta).$ 

\begin{subsubsection}{Existence and Uniqueness of Best Response}

 We can apply Lemma \ref{lemm:diff_concavity_utility} to show that the expected utility \eqref{eq:expected_utility function} is twice continuously differentiable and strictly concave in $x$, and thus continuous in $x$. Since $\mathcal{X}$ is compact, the expected utility attains a maximum value on $\mathcal{X}$. Thus, there exists $x_{i}^{*} \in \mathcal{X}$ that maximizes the expected utility. When $x_{i}^{*}(\beta, s) \in \text{Int}(\mathcal{X})$, $x_{i}^{*}(\beta, s)$ must be the unique maximizer of the expected utility on $\mathcal{X}$ (Lemma \ref{lemm:unique_maximizer}).

\end{subsubsection}


\begin{subsubsection}{Differentiability}
Fix $\beta$. We aim to show that if a best response $x_{i}^{*}(\beta, s) \in \text{Int}(\mathcal{X}),$ then $x_{i}^{*}$ is continuously differentiable in $s$ by applying the Implicit Function Theorem (Theorem \ref{theo:implicit_function_theorem}). 

By Lemma \ref{lemm:diff_concavity_utility}, if a best response $x_{i}^{*}(\beta, s) \in \text{Int}(\mathcal{X})$, then it can be expressed implicitly as the solution to $\nabla_{x} \EE[\epsilon]{U_{i}(x;\beta, s)} = 0.$ By Lemma \ref{lemm:diff_concavity_utility}, the conditions of the Implicit Function Theorem are satisfied: 1) $\nabla_{x} \EE[\epsilon]{U_{i}(x;\beta, s)}$ is continuously differentiable in its arguments, 2) at any point $(x_{0}, s_{0})$, we have $\text{det}(\nabla_{x}^{2} \EE[\epsilon]{U_{i}(x_{0};\beta, s_{0})}) \neq 0$. As a result, $x_{i}^{*}$ is continuously differentiable in $s$. An analogous proof can be used to show that the best response $x_{i}^{*}$ is continuously differentiable in $\beta$.

\end{subsubsection}

\begin{subsubsection}{Contraction}
To show that $\omega_{i}(s; \beta)$ is a contraction in $s$ with Lipschitz constant less than $\bar{\kappa}$, it is sufficient to show that $|\nabla_{s} \omega_{i}(s; \beta)| < \bar{\kappa}$. Since $\sigma^{2} > \frac{1}{\alpha_{i} \sqrt{2\pi e}} \cdot \frac{\bar{\kappa} + 1}{\bar{\kappa}}$ for some $\bar{\kappa} \in (0, 1],$ we have that the form of $\nabla_{s} \omega_{i}(s; \beta)$ is given by \eqref{eq:score_gradient}.

 First, we show that the denominator of the expression on the right of \eqref{eq:score_gradient} must be greater than $0.$
 \begin{align}
&1 + \phi'_{\sigma}(s - \omega_{i}(s; \beta)) \beta^{T}\bfH^{-1}\beta \\
&\geq 1 + \inf_{y \in \mathbb{R}} \phi'_{\sigma}(y) \cdot \sup_{\beta} \beta^{T} \bfH^{-1} \beta \label{eq:C1} \\
&\geq 1 - \frac{1}{\sigma^{2} \sqrt{2\pi e}} \cdot \frac{1}{\alpha_{i}} \label{eq:C2} \\
& > 0 \label{eq:C3}
 \end{align}
\eqref{eq:C1} follows from the observation that $\phi'_{\sigma}(y)$ may take negative values while $\bfH^{-1}$ is positive definite. \eqref{eq:C2} holds because $|\phi'_{\sigma}(y)| \leq \frac{1}{\sigma^{2} \sqrt{2\pi e}}$ and
\[ \sup_{\beta' \in \mathcal{B}} \beta'^{T}\bfH^{-1}\beta' \leq \frac{1}{\alpha_{i}}.\] In \eqref{eq:C3}, we use our assumption that $\sigma^{2} > \frac{1}{\alpha_{i}\sqrt{2 \pi e}}.$ 

Now, we show that the numerator of the expression on the right of \eqref{eq:score_gradient} satisfies the following bounds
\begin{equation}\label{eq:bounds} -\frac{\bar{\kappa}}{\bar{\kappa} + 1} < \phi'_{\sigma}(s - \omega_{i}(s; \beta)) \beta^{T}\bfH^{-1}\beta < \frac{\bar{\kappa}}{1 - \bar{\kappa}}. \end{equation}

We first show the upper bound.
\begin{align*}
&\phi'_{\sigma}(s - \omega_{i}(s; \beta)) \beta^{T}\bfH^{-1}\beta \\
&\leq \sup_{y \in \mathbb{R}} \phi'_{\sigma}(y) \cdot \sup_{\beta} \beta^{T} \bfH^{-1} \beta  \\
&= \frac{1}{\sigma^{2} \sqrt{2\pi e}} \cdot \frac{1}{\alpha_{i}} \\
&\leq \frac{\bar{\kappa}}{1 - \bar{\kappa}},
\end{align*}
where the last line follows from the fact that $\sigma^{2} > \frac{1}{\alpha_{i}\sqrt{2\pi e}} \cdot \frac{1 - \bar{\kappa}}{\bar{\kappa}}$ because $\bar{\kappa} \in (0, 1).$

Now, we show the lower bound.
\begin{align*}
&\phi'_{\sigma}(s - \omega_{i}(s; \beta)) \beta^{T}\bfH^{-1}\beta \\
&\geq \inf_{y \in \mathbb{R}} \phi'_{\sigma}(y) \cdot \sup_{\beta} \beta^{T} \bfH^{-1} \beta  \\
&= -\frac{1}{\sigma^{2} \sqrt{2\pi e}} \cdot \frac{1}{\alpha_{i}} \\
&\geq -\frac{\bar{\kappa}}{\bar{\kappa} + 1},
\end{align*}
where the last line follows from the fact that $\sigma^{2} > \frac{1}{\alpha_{i}\sqrt{2\pi e}} \cdot \frac{\bar{\kappa} + 1}{\bar{\kappa}}.$

Using the result that the denominator is positive and the numerator satisfies the bounds in \eqref{eq:bounds} yields that $|\nabla_{s} \omega_{i}(s; \beta)| < \bar{\kappa}.$ We can see this by realizing that $\nabla_{s} \omega_{i}(s; \beta)$ has the form of a function $f(s) = \frac{\tilde{f}(s)}{\tilde{f}(s) + 1}.$ Since we can guarantee that $\tilde{f}(s) + 1 > 0$ and $-\frac{\bar{\kappa}}{\bar{\kappa} + 1} <\tilde{f}(s) <  \frac{\bar{\kappa}}{1 - \bar{\kappa}}$, we can rearrange these expressions to find that they imply that
\[ -\bar{\kappa} < \frac{\tilde{f}(s)}{\tilde{f}(s) + 1} < \bar{\kappa},\]
which yields our desired result. Thus, $|\nabla_{s} \omega_{i}(s; \beta)| < \bar{\kappa},$ so $\omega_{i}(s; \beta)$ must be a contraction in $s$ with Lipschitz constant less than $\bar{\kappa}$.

\end{subsubsection}
\end{subsection}
\end{section}

\begin{section}{Proofs of Mean-Field Results}
\label{sec:eq_proofs}
We state  technical lemmas that will be used in many of our results. The proofs of these lemmas can be found in Appendix \ref{sec:technical_proofs}.

\begin{lemm}
\label{lemm:cdf}
The distribution $P(\beta, s)$ is given by 
\begin{equation}
\label{eq:cdf}
    P(\beta, s)(r) = \int \Phi_{\sigma}(r - \beta^{T} x^{*}_{i}(\beta,s)) dF. \\
\end{equation} 
Under  Assumptions \ref{assumption:strong_convexity}, \ref{assumption:finite_types}, and \ref{assumption:br_interior}, if $\sigma^{2} > \frac{1}{\alpha_{*}\sqrt{2\pi e}}$, then $P(\beta, s)(r)$ is a well-defined function. Furthermore, it is strictly increasing in $r$, continuously differentiable in $\beta, s, r$, and has a unique continuous inverse distribution function. \hyperref[subsec:cdf]{Proof in Appendix \ref{subsec:cdf}}.
\end{lemm}

\begin{lemm}
\label{lemm:dq_ds}
Fix $\beta \in \mathcal{B}, \bar{\kappa} \in (0, 1].$ Suppose  Assumptions \ref{assumption:strong_convexity}, \ref{assumption:finite_types}, and \ref{assumption:br_interior} hold. If $\sigma^{2} > \frac{1}{\alpha_{*}\sqrt{2\pi e}}$, then $\frac{\partial q(P(\beta, s))}{\partial s} < 1$. If $\sigma^{2} > \frac{1}{\alpha_{*}\sqrt{2\pi e}} \cdot \frac{\bar{\kappa} + 1}{\bar{\kappa}}$, then $\Big|\frac{\partial q(P(\beta, s))}{\partial s}\Big| < \kappa.$ \hyperref[subsec:dq_ds]{Proof in Appendix \ref{subsec:dq_ds}}.
\end{lemm}

\begin{lemm}
\label{lemm:dist_limits}
Let $\beta \in \mathcal{B}.$ Let $P$ be the distribution over $\beta^{T}(Z_{i} + \epsilon_{i})$ where $(Z_{i}, c_{i}, \epsilon_{i}) \sim F$. Under Assumption \ref{assumption:strong_convexity}, \ref{assumption:finite_types}, \ref{assumption:br_interior}, if $\sigma^{2} > \frac{1}{\alpha_{*}\sqrt{2\pi e} },$ then $P(\beta, s)$ converges pointwise to $P$ as $s \rightarrow \infty$ and as $s \rightarrow -\infty.$ In addition,
\begin{align*}
    \lim_{s \rightarrow \infty} q(P(\beta, s)) &= q(P), \\
    \lim_{s \rightarrow - \infty} q(P(\beta, s)) &= q(P).
\end{align*}
\hyperref[subsec:dist_limits]{Proof in Appendix \ref{subsec:dist_limits}}.
\end{lemm}

\begin{subsection}{Proof of Theorem \ref{theo:equilibrium}}
This proof has several parts. First, we establish that $q(P(\beta, s))$ is continuously differentiable in $(\beta, s)$. Second, we verify that $q(P(\beta, \cdot))$ has at least one fixed point. Third, we show that the fixed point is unique. Fourth, we demonstrate that the function $s(\beta)$, that maps $\beta$ to the fixed point of $q(P(\beta, s))$ is a continuously differentiable function of $\beta$. Fifth, we check that when $\sigma^{2} > \frac{2}{\alpha_{*} \sqrt{2\pi e}},$ $q(P(\beta, \cdot))$ is a contraction and fixed point iteration will converge.

\begin{subsubsection}{Differentiability}
We show that $q(P(\beta, s))$ is continuously differentiable in $(\beta, s).$ First, we express $q(P(\beta, s))$ implicitly as the value of $r$ that solves
\begin{equation}
\label{eq:implicit_q}
    \psi_{1}(s, r) = P(\beta, s)(r) -q = 0.
\end{equation}
because $P(\beta, s)$ has a unique inverse distribution function (Lemma \ref{lemm:cdf}). Second, we can apply the Implicit Function Theorem (Theorem \ref{theo:implicit_function_theorem}) can be applied to the above equation to show that $r = q(P(\beta, s))$ is continuously differentiable in $s$. By Lemma \ref{lemm:cdf}, the conditions of the Implicit Function Theorem hold because 1) $\psi_{1}(s, r)$ is continuously differentiable and 2) $\frac{\partial \psi_{1}}{\partial r} \neq 0$ because  $\frac{\partial P(\beta, s)(r)}{\partial r}$ is strictly increasing.  Thus, $q(P(\beta, s))$ is a continuously differentiable function of $s$. An analogous proof can be used to show that $q(P(\beta, s))$ is continuously differentiable in $\beta$.
\end{subsubsection}

\begin{subsubsection}{Existence}
We demonstrate that $q(P(\beta, \cdot))$ must have at least one fixed point. We apply the Intermediate Value Theorem to the function $g(s) = s - q(P(\beta, s))$. We will show that there exist $s_{l}$ and $s_{h}$ such that $g(s_{l}) < 0$ and $g(s_{h}) > 0.$ So, there must be $s$ such that $g(s) = 0$, demonstrating that $q(P(\beta, \cdot))$ has at least one fixed point. 

To define $s_{h}$ and $s_{l}$, we recall that by Lemma \ref{lemm:dist_limits}, we have that 
\begin{align*}
    \lim_{s \rightarrow \infty} q(P(\beta, s)) =\lim_{s \rightarrow -\infty} q(P(\beta, s)) = q(P).
\end{align*} Let $\delta > 0.$ There exists $S_{1} < \infty$ so that for $s \geq S_{1}$, $|q(P(\beta, s)) - q(P)| < \delta$, and $S_{2} > -\infty$ so that for $s \leq S_{2},$ $|q(P(\beta, s)) - q(P)| < \delta$. Let $s_{h} = \max(q(P) + \delta, S_{1}).$ Then we have for $s \geq s_{h}$,
\begin{align*}
    g(s) &=s - q(P(\beta, s)) \\
    &> s - q(P) - \delta  \\
    &\geq q(P) + \delta - q(P) - \delta \\
    &= 0.
\end{align*}
Similarly, we can define $s_{l} = \min(q(P) - \delta, S_{2})$ and observe that for $s \leq s_{l},$ $g(s) < 0.$ Thus, $q(P(\beta, s))$ must have at least one fixed point.
\end{subsubsection}

\begin{subsubsection}{Uniqueness}
If a fixed point exists, it must satisfy $P(\beta, s)(s) = q$ because $P(\beta, s)$ has a unique inverse (Lemma \ref{lemm:cdf}). We find that $P(\beta, s)(s)$ is a continuous and strictly increasing function of $s$. A continuous and strictly increasing function can intersect a horizontal line in at most one point, so the fixed point must be unique.

Applying \eqref{eq:cdf} from Lemma \ref{lemm:cdf}, we have
\begin{align*}
    P(\beta, s)(s) &= \int \Phi_{\sigma}(s - \beta^{T} x_{i}^{*}(\beta,  s)) dF \\
    &= \int \Phi_{\sigma}(h_{i}(s; \beta)) dF, 
\end{align*}   
where $h_{i}(s; \beta) = s - \beta^{T} x_{i}^{*}(\beta, s)$. 

From Lemma \ref{lemm:score_gradient}, we have that $h_{i}(s; \beta)$  is strictly increasing in $s$ for any agent with unobservables in the support of $F$. Since $\Phi_{\sigma}$ is a strictly increasing CDF, so we have that $\Phi_{\sigma}(h_{i}(s; \beta))$ is also strictly increasing. Finally, the sum of strictly increasing functions is strictly increasing, which gives that the integral is also a strictly increasing function of $s$. Since $P(\beta, s)(s)$ is continuous and strictly increasing in $s$, there is at most one point where it can equal $q$, guaranteeing uniqueness of the fixed point.
\end{subsubsection}

\begin{subsubsection}{Differentiability of Fixed Point}
Let $s:\mathcal{B} \rightarrow \mathbb{R}$ be a mapping from $\beta$ to the unique fixed point of $q(P(\beta, \cdot))$. We can verify that $s$ is continuously differentiable in $\beta.$

Note that we can implicitly represent $s(\beta)$ by $s$ in the following equation
\[ \psi_{2}(\beta, s) = s - q(P(\beta, s)) = 0.\]
We verify that the conditions of the Implicit Function Theorem are satisfied because 1) $\psi_{2}( \beta, s)$ must be continuously differentiable in its arguments because $q(P(\beta, s))$ is continuously differentiable and 2) $\frac{\partial \psi_{2}(\beta, s)}{\partial s}(\beta_{0}, s_{0}) \neq 0$ because
\begin{align*}
    \frac{\partial \psi_{2}(\beta, s)}{\partial s} = 1 - \frac{\partial q(P(\beta, s))}{\partial s}.\\
\end{align*}
From Lemma \ref{lemm:dq_ds}, we have that $\frac{\partial q(P(\beta, s))}{\partial s} < 1$, so  $\frac{\partial \psi_{2}(\beta, s)}{\partial s} > 0$. As a result, we can apply the Implicit Function Theorem and find that $s(\beta)$ is a continuously differentiable function of $\beta$.

\end{subsubsection}

\begin{subsubsection}{Contraction and Fixed Point Iteration}
If $\sigma^{2} > \frac{1}{\alpha_{*} \sqrt{2\pi e}} \cdot \frac{\kappa + 1}{\kappa}$, we can apply the second part of Lemma \ref{lemm:dq_ds} to conclude that $|\frac{\partial q(P(\beta, s))}{\partial s}|< \kappa$. By Lemma \ref{lemm:contraction_derivative}, $q(P(\beta, s))$ is a contraction in $s$. As a consequence of Banach's Fixed Point Theorem (Theorem \ref{theo:banach_fixed_point}), we can conclude that fixed point iteration \eqref{eq:fpi} converges to the unique fixed point of $q(P(\beta, s)).$

\end{subsubsection}
\end{subsection}

\end{section}

\begin{section}{Proofs of Finite-Sample Approximation Results}
\label{sec:finite_model_proofs}
\begin{lemm}
\label{lemm:stoch_dom_little}
Suppose the conditions of Theorem \ref{theo:emp_fpi_convergence}. Let $\{z^{t}\}$ be a sequence of random variables where 
\[ z^{t} = \begin{cases} \lambda_{g} & \text{w.p. } p_{n}(\lambda_{g}) \\
C_{k} &\text{w.p. } \frac{1 - p_{n}(\lambda_{g})}{2^{k}}, k \geq 1,\\
\end{cases} \]
where $p_{n}(\lambda_{g})$ is the bound from Lemma \ref{lemm:constant_in_quantile_bound} and 
\begin{equation} 
\label{eq:s_k}
C_{k} = \sqrt{\frac{1}{2nM_{\lambda_{g}}^{2}} \cdot \log\Big(\frac{2^{k+1}}{ 1 - p_{n}(\lambda_{g})}\Big)}.\end{equation}
For any $s \in \mathbb{R}$, $z^{t}$ stochastically dominates $|q(P^{n}(\beta, s)) - q(P(\beta, s))|$. \hyperref[subsec:stoch_dom_little]{Proof in Appendix \ref{subsec:stoch_dom_little}}.
\end{lemm}

\begin{lemm}
\label{lemm:stoch_dom_big}
Suppose the conditions of Theorem \ref{theo:emp_fpi_convergence} hold. Let $\{\hat{S}^{t}_{n}\}_{t \geq 0}$ be a stochastic process generated via \eqref{eq:emp_fpi}. Let $C$ be as defined in Theorem \ref{theo:emp_fpi_convergence}. Let $\{z^{t}\}_{t \geq 0}$ be a sequence of random variables where 
\[ z^{t} = \begin{cases} \lambda_{g} & \text{w.p. } p_{n}(\lambda_{g}) \\
C_{k} &\text{w.p. } \frac{1 - p_{n}(\lambda_{g})}{2^{k}}, k \geq 1,\\
\end{cases} \]
where $p_{n}(\lambda_{g})$ is the bound from Lemma \ref{lemm:constant_in_quantile_bound} and $C_{k}$ is as defined in Lemma \ref{lemm:stoch_dom_little}. Let $\kappa$ be the Lipschitz constant of $q(P(\beta, s)).$ Then $\sum_{i=1}^{t} z^{t-i} \kappa^{i} + \kappa^{t} \cdot C$ stochastically dominates $|\hat{S}^{t}_{n} - s(\beta)|.$  \hyperref[subsec:stoch_dom_big]{Proof in Appendix \ref{subsec:stoch_dom_big}}.
\end{lemm}

\begin{subsection}{Proof of Lemma \ref{lemm:constant_in_quantile_bound}}
\label{subsec:constant_in_quantile_bound}
We define notation that will be used in the rest of the proof. Let $s_{i}^{*} = \argmax_{s \in \mathbb{R}} \omega_{i}^{*}(\beta, s).$ Note that $s_{i}^{*}$ is unique by Lemma \ref{lemm:limits_br}.
Let $\mathcal{A}$ be the set of $s_{i}^{*}$ for each agent $i$ with type in support of $F$, i.e.
\[ \mathcal{A} = \{(Z_{i}, c_{i}) \in \text{supp}(F) \mid s_{i}^{*}\}.\]
Under Assumption \ref{assumption:finite_types}, $\mathcal{A}$ is bounded, so we can write
\begin{align*}
s_{L} := \inf \mathcal{A}\,\quad s_{H} := \sup \mathcal{A}.
\end{align*}
We also define
\begin{align*}
    f_{1}(s) &:= P(\beta, s)(q(P(\beta, s)) + \lambda ) - q \\
    f_{2}(s) &:= q - P(\beta, s)(q(P(\beta, s) - \lambda).
\end{align*}
Recall that
\[ M_{\lambda} = \inf_{s \in \mathbb{R}} \min \{f_{1}(s), f_{2}(s)\},\]
and we aim to show that $M_{\lambda} > 0.$ We note that 
\begin{align*}
    M_{\lambda} &= \min \{ \inf_{s \in \mathbb{R}} f_{1}(s),  \inf_{s \in \mathbb{R} } f_{2}(s) \}.
\end{align*}

We note that $\inf_{s \in \mathbb{R}} f_{i}(s) = \min\{ \inf_{s < s_{L}} f_{i}(s), \inf_{s > s_{H}} f_{i}(s), \inf_{s \in [s_{L}, s_{H}]} f_{i}(s) \}$ for $i= 1, 2.$ 

The differentiability of $P(\beta, s)$ and $q(P(\beta, s))$ in $s$ is given by Lemma \ref{lemm:cdf} and Theorem \ref{theo:equilibrium}, respectively. Thus, we can write that
\begin{align*}
    \frac{df_{1}}{ds} &= p_{\beta, s}(q(P(\beta, s)) + \lambda) \cdot \frac{\partial q(P(\beta, s))}{\partial s}.
\end{align*}
By Lemma \ref{lemm:limits_br}, the expected score $\omega_{i}(s; \beta)$ for each agent $i$ is increasing on $s \leq s_{L}$ and decreasing on $s \geq s_{H}.$ By Lemma \ref{lemm:dq_ds}, $\frac{\partial q(P(\beta, s))}{\partial s}$ is a convex combination of $\beta^{T}\nabla_{s} x^{*}_{i}$. Thus, $\frac{\partial q(P(\beta, s))}{\partial s}$ is positive for $s < s_{L}$  and negative on $s > s_{H}.$ This implies that $\frac{df_{1}}{ds}$ is positive for $s < s_{L}$ and negative on $s > s_{H}.$ Thus, $f_{1}(s)$ is increasing on $(-\infty, s_{L})$ and $f_{1}(s)$ is decreasing on $(s_{H}, \infty).$ So,
\begin{align*}
    \inf_{s < s_{L}} f_{1}(s) &= \lim_{s \rightarrow -\infty} f_{1}(s) \\
    &= \lim_{s \rightarrow -\infty}  P(\beta, s)(q(P(\beta, s)) + \lambda ) - q \\
    &= P(q(P) + \lambda) - q,
\end{align*}
where the last line follows from Lemma \ref{lemm:dist_limits} and $P$ is the distribution defined that lemma. 
Similarly, we can show that
\[\inf_{s > s_{H}} f_{1}(s) = \lim_{s \rightarrow \infty} f_{1}(s) = P(q(P) + \lambda) - q.\]
Finally, because $[s_{L}, s_{H}]$ is a compact set, there is some $s_{1} \in [s_{L}, s_{H}]$ for which $f_{1}(s)$ achieves its infimum on the interval. Thus, $\inf_{s \in \mathbb{R}} f_{1}(s) = \min\{ P(q(P) + \lambda) - q, P(\beta, s_{1})(q(P(\beta, s_{1})) + \lambda) - q \}.$
We conclude that $\inf_{s \in \mathbb{R}} f_{1}(s) > 0$.

By a similar analysis, we can show that 
\[ \inf_{s \in \mathbb{R}} f_{2}(s) = \min_{s \in \{s_{L}, s_{H}, s_{2}\}} f_{2}(s),\]
so $\inf_{s \in \mathbb{R}} f_{2}(s) > 0.$ Thus, $M_{\lambda} > 0.$

Second, we show that $M_{\lambda}$ is approximately linear in $\lambda$ as $\lambda$ approaches $0$. Recall that agent $i$ with type $(Z_{i}, c_{i})$ has expected score $\omega_{i}^{*}(s; \beta).$ By Lemma 31 (see appendix), $\omega_{i}^{*}(s; \beta)$ has unique maximizer $s_{i}^{*}$. Let $\mathcal{A}$ be the set of all $s_{i}^{*}$ of the best response for all agents with type in the support of $F$, i.e.
    \[\mathcal{A} = \{ s_{i}^{*} \mid (Z_{i}, c_{i}) \in \text{supp}(F)\}.\]
    Note that $\mathcal{A}$ must be bounded because $F$ has finite support. The proof of Lemma 3 demonstrates that
    {
    \begin{align} \label{eq:m_lambda} \begin{split}M_{\lambda} = \min &\Big\{ \{ P(q(P) + \lambda) - q, P(\beta, s_{1})(q(P(\beta, s_{1}) + \lambda)) - q \} \\
    &\cup \{q - P(\beta, s)(q(P(\beta, s)) - \lambda)\}_{s \in [\inf(\mathcal{A}), \sup(\mathcal{A}), s_{2}]} \Big\}.
    \end{split}
    \end{align}
    }%
    for particular choices of $s_{1}, s_{2} \in [\inf(\mathcal{A}), \sup(\mathcal{A})]\}.$
    We observe that $M_{\lambda}$ scales approximately linearly in $\lambda$ by applying a Taylor expansion to any of the terms on the right side of \eqref{eq:m_lambda}, e.g.
    \begin{align*} 
    &P(q(P) + \lambda) -q  \\
    &= P(q(P)) - q + \lambda \cdot (p(q(P))) + o(|\lambda|^{2}).
    \end{align*}

Now, we proceed to show the third component of the lemma. From Theorem \ref{theo:concentration_quantiles}, we have that 
$P(|q(P(\beta, s)) - q(P^{n}(\beta, s))| < \lambda ) \geq1- 4 e^{-2n M_{\lambda, s}^{2}},$
where $M_{\lambda, s} = \min \{f_{1}(s), f_{2}(s)\}.$ We can obtain a bound that is uniform over $s$ by realizing that 
$M_{\lambda} = \inf_{s \in \mathbb{R}} \min \{f_{1}(s), f_{2}(s)\}$ and $M_{\lambda} > 0.$ So, we have that
\[ P(|q(P(\beta, s)) - q(P^{n}(\beta, s))| < \lambda ) \geq 1- 4 e^{-2n M_{\lambda}^{2}}.\]
\end{subsection}

\begin{subsection}{Proof of Theorem \ref{theo:emp_fpi_convergence}}
\label{subsec:emp_fpi_convergence}
Let $\{z^{t}\}_{t \geq 1}$ be a sequence of random variables where 
\[ z^{t} = \begin{cases} \lambda_{g} & \text{w.p. } p_{n}(\lambda_{g}) \\
C_{k} &\text{w.p. } \frac{1 - p_{n}(\lambda_{g})}{2^{k}}, k\geq 1\\
\end{cases},\] 
where $\lambda_{g} = \frac{\lambda
 (1 - \kappa)}{2}$ and $p_{n}(\lambda_{g})$ is the bound from Lemma \ref{lemm:constant_in_quantile_bound}. From Lemma \ref{lemm:stoch_dom_big}, we have that
\begin{align}
    |\hat{S}^{t}_{n} - s(\beta)| &\preceq_{\text{SD}} \sum_{i=1}^{t} z^{t-i} \kappa^{i} + \kappa^{t} \cdot C. \label{eq:sd_2}
\end{align}
We note that 
\begin{align*}
  \sum_{i=1}^{t} \lambda_{g} \kappa^{i} < \sum_{i=1}^{\infty} \lambda_{g} \kappa^{i} &= \frac{\lambda_{g}}{1- \kappa} = \frac{\lambda (1- \kappa)}{2} \cdot \frac{1}{1 - \kappa} = \frac{\lambda}{2}.
\end{align*}
In addition, let $t \geq \Big\lceil \frac{\log(\frac{\lambda}{2C}) }{\log \kappa} \Big\rceil $. For such $t$, we have that 
\[  t \geq \frac{\log(\frac{\lambda}{2C}) }{\log \kappa}.\]
Rearranging the above inequality gives
\[ \kappa^{t}C \leq \frac{\lambda}{2}.\]
As a result, we have that 
\[ \sum_{i=1}^{t} \lambda_{g} \kappa^{i} + \kappa^{t}C < \frac{\lambda}{2} + \frac{\lambda}{2} = \lambda\]
By \eqref{eq:sd_2} and the definition of stochastic dominance, we have that
\begin{align*}
    &P(|\hat{S}^{t}_{n} - s(\beta)| \leq \lambda ) \\
    &\geq P(\sum_{i=1}^{t} z^{t-i} \kappa^{i} + \kappa^{t} C \leq \lambda ) \\
    &\geq P( z^{t-i} = \lambda_{g} \text{ for } i = 0, 1 \dots t) \geq (p_{n}(\lambda_{g}))^{t+1}
\end{align*}
If we have that 
\begin{equation}
\label{eq:n}
n \geq \frac{1}{2 M_{\lambda_{g}}^{2}} \log \Big(\frac{4t}{\delta}\Big),
\end{equation} then we can show that $p_{n}(\lambda_{g}) \geq 1 - \frac{\delta}{t}.$ We can rearrange \eqref{eq:n}
\[  e^{-2nM_{\lambda_{g}}^{2}} \leq \frac{\delta}{4t}.\]
Rearranging again,
\[ 1 - 4e^{-2nM_{\lambda_{g}}^{2}} \geq  1- \frac{\delta}{t}.\]
Thus, we have that
\[ p_{n}(\lambda_{g}) \geq 1 - \frac{\delta}{t}.\]

So, $(p_{n}(\lambda_{g}))^{t} \geq (1 - \frac{\delta}{t+1})^{t}.$ Applying Theorem \ref{theo:bernoulli_inequality} gives that $(p_{n}(\lambda_{g}))^{t} \geq 1 - \delta.$ Therefore, we conclude that if $t \geq  \Big\lceil \frac{\log(\frac{\lambda}{2C}) }{\log \kappa} \Big\rceil$ and $n \geq \frac{1}{2 M_{\lambda_{g}}^{2}} \log \Big(\frac{4t}{\delta}\Big)$, then 
\[P(|\hat{S}^{t}_{n} - s(\beta)| \leq \lambda ) \geq 1 - \delta, \]
as desired.

\end{subsection}

\begin{subsection}{Proof of Corollary \ref{coro:convergence_in_prob_threshold}}
\label{subsec:convergence_in_prob_threshold}
To show that $\hat{S}_{n}^{t_{n}} \xrightarrow{p} s(\beta)$, we must show that
\[\lim_{n \rightarrow \infty} P(|\hat{S}_{n}^{t_{n}} - s(\beta)| > \lambda) = 0.\]
It is sufficient to show that for any $\delta > 0$, there exists $N$ such that for $n \geq N$, \[P(|\hat{S}_{n}^{t_{n}} - s(\beta)| > \lambda) \leq \delta.\] As in the statement of Theorem \ref{theo:emp_fpi_convergence}, let $C = |\hat{S}^{1}_{n} - s(\beta)|.$ Let $N_{1} \in \mathbb{N}$ be the smallest value of $n$ such that $t_{n} \geq \lceil \frac{\log(\frac{\lambda}{2C}) }{\log \kappa} \rceil$. 

We have that $t_{n} \prec \exp(n)$. So, there exists $N_{2} \in \mathbb{N}$ such that for $n \geq N_{2},$
\[ t_{n} \leq \frac{\delta}{4}\exp \Big(2 n M_{\lambda_{g}}^{2}\Big).\]
Rearranging this equation, we have that for $n \geq N_{2},$
\[ \exp(2nM_{\lambda_{g}}^{2}) \geq \frac{4t_{n}}{\delta}.\]
Taking log of both sides yields for $n \geq N_{2}$
\[ 2nM_{\lambda_{g}}^{2} \geq \log(\frac{4t_{n}}{\delta}).\]
So, for $n \geq N_{2},$ we have that
\[ n \geq \frac{1}{2M_{\lambda_{g}}^{2}}\log(\frac{4t_{n}}{\delta}).\]
We can take $N = \max\{N_{1}, N_{2}\}.$ By Theorem \ref{theo:emp_fpi_convergence}, we have that for $n \geq N, P(|\hat{S}^{t_{n}}_{n} - s(\beta)| > \lambda) < \delta.$
Thus, we have that $\hat{S}^{t_{n}}_{n} \xrightarrow{p} s(\beta).$
\end{subsection}
\end{section}

\begin{section}{Proofs of Learning Results}
\label{sec:learning_proofs}
We state technical lemmas that will be used in many of our learning results. 

\begin{lemm}
\label{lemm:truncation}
Let $\beta \in \mathcal{B}.$ Let $s(\beta)$ be the mean-field equilibrium threshold. Define a truncated stochastic fixed point iteration process
\begin{equation}
    \hat{S}^{t}_{n} = \min( \max(q(P_{\beta, \hat{S}^{t-1}_{n}}^{n}), -D), D).
\end{equation}
Under Assumptions \ref{assumption:strong_convexity}, \ref{assumption:finite_types}, \ref{assumption:br_interior}, if $\sigma^{2} > \frac{2}{\alpha_{*}\sqrt{2 \pi e}},$ then for any sequence $\{t_{n}\}$ such that $t_{n} \uparrow \infty$ as $n \rightarrow \infty$ and $t_{n} \prec \exp(n)$, we have that $\hat{S}^{t_{n}}_{n} \xrightarrow{p} s(\beta).$ \hyperref[subsec:truncation]{Proof in Appendix \ref{subsec:truncation}}.
\end{lemm}

\begin{lemm}
\label{lemm:nice_loss_and_pi}
Recall that under Assumption \ref{assumption:nice_loss}, $m(z, c; w) = \EE[]{Y_{i}(w) \mid Z_{i}=z, c_{i}=c}.$ Let $\tilde{j}, \tilde{y}, \tilde{k}$ be functions $\text{supp}(F_{Z, c, \epsilon}) \times \mathcal{B} \times \mathcal{S} \times \mathcal{S} \rightarrow \mathbb{R}$ such that 
\begin{align*}
&\tilde{j}(Z_{i}, c_{i}, \epsilon_{i}, \beta, s, r) \\
&\indent= \pi(X_{i}(\beta, s); \beta, r), \\
&\tilde{y}(Z_{i}, c_{i}, \epsilon_{i}, \beta, s, r) \\
&\indent= m(Z_{i}, c_{i}; \pi(X_{i}(\beta, s); \beta, r)) \\
&\tilde{k}(Z_{i}, c_{i}, \epsilon_{i}, \beta, s, r) \\
&\indent= \mathbb{I}\Big( \frac{r - \beta^{T}X_{i}(\beta, s)}{h} \in \Big[-\frac{1}{2}, \frac{1}{2}\Big) \Big).
\end{align*}
Under the conditions of Theorem \ref{theo:direct_effect}, $\tilde{y}, \tilde{j},$ and $\tilde{k}$ satisfy the requirements on the function $a(u; \theta)$ from Lemma \ref{lemm:uniform_convergence_sum}, where the data $u$ is given by $(Z_{i}, c_{i}, \epsilon_{i}) \sim F_{Z, c, \epsilon}$  and the parameter $\theta$ is given by $(\beta, s, r)$.
 \hyperref[subsec:nice_loss_and_pi]{Proof in Appendix \ref{subsec:nice_loss_and_pi}}.
\end{lemm}

\begin{lemm}
\label{lemm:find_equivalent_agent_main}
 Let $\beta \in \mathcal{B}, s \in \mathcal{S}, \zeta \in \{-1, 1\}^{d}, \xi \in \{-1, 1\},$ and $b > 0$. Suppose Assumption \ref{assumption:strong_convexity} holds for all $c_{i} \sim F.$ Define $T$ to be a mapping from an agent $i$ with unobservables $(Z_{i}, c_{i}) \in \text{supp}(F)$ to an agent $i'$ with unobservables $Z_{i'; b, \zeta, \xi}, c_{i'}$. Let
\begin{align*}
    x_{1} &:= x^{*}_{i}(\beta + b\zeta, s + b\xi) \\
    r &:= (\beta + b\zeta)^{T}x_{1} - b\xi.
\end{align*}
Define $\tilde{Z}_{i';b, \zeta, \xi}$ and $c_{i'}$ as follows.
\begin{align}
     Z_{i'; b, \zeta, \xi} &:= Z_{i} + \beta \cdot b \cdot (\zeta^{T}x_{1} - \xi) \label{eq:eta_prime} \\
     c_{i'}(y) &:=c_{i}(y) - \phi_{\sigma}(s-r) \beta^{T}y \label{eq:c_nu_prime}.
\end{align}
If $Z_{i}, x_{1} \in \text{Int}(\mathcal{X})$ and $b$ sufficiently small, then $Z_{i'; b, \zeta, \xi} \in \mathcal{X}$, $c_{i'}$ is $\alpha_{i}$-strongly convex,  
\[ x^{*}_{i'}(\beta, s) = x_{1} + b \cdot \beta(\zeta^{T}x_{1} - \xi), \]
$x^{*}_{i'}(\beta, s) \in \text{Int}(\mathcal{X}),$ and $\beta^{T}x_{i'; b, \zeta, \xi}^{*}(\beta, s)=r.$
In other words, when the agent $i'$ with unobservables given by $T(i)$ best responds to the unperturbed model $\beta$ and threshold $s$, they obtain the same raw score (without noise) as the agent $i$ who responds to a perturbed model $\beta + b\zeta$ and threshold $s + b\xi$. \hyperref[subsec:find_equivalent_agent_main]{Proof in Appendix \ref{subsec:find_equivalent_agent_main}}.
\end{lemm}

\begin{lemm}
\label{lemm:transformed_distribution}
Suppose the conditions of Theorem \ref{theo:direct_effect} hold. Fix $\beta \in \mathcal{B}, s \in \mathcal{S}$. For sufficiently small $b$, there exists a distribution over unobservables $\tilde{F}$ such that when agents with unobservables $(Z_{i'}, c_{i'}) \sim \tilde{F}$ best respond to the unperturbed model $\beta$ and threshold $s$ the induced score distribution is equal to $P(\beta, s, b).$ Furthermore, the support of $\text{supp}(\tilde{F}_{Z}) \subset \mathcal{X}$, each $c_{i'} \sim \tilde{F}$ satisfies Assumption \ref{assumption:strong_convexity} and lies in $\mathcal{C}$, $\tilde{F}$ has a finite number of agent types, $\alpha_{*}(\tilde{F}) = \alpha_{*}(F)$, and for any agent $i'$ with unobservables $(Z_{i'}, c_{i'}) \sim \tilde{F}$,  we have $x_{i'}^{*}(\beta, s) \in \text{Int}(\mathcal{X})$.
\hyperref[subsec:transformed_distribution]{Proof in Appendix \ref{subsec:transformed_distribution}}.
\end{lemm}

\begin{lemm}
\label{lemm:transformed_distribution_continuity_fp}
Fix $\beta \in \mathcal{B}.$ Suppose the conditions of Theorem \ref{theo:direct_effect} hold. If $b$ is sufficiently small, then $q(P(\beta, s, b))$ has a unique fixed point $s(\beta, b)$. As $b \rightarrow 0$, $s(\beta, b) \rightarrow s(\beta)$, where $s(\beta)$ is the unique fixed point of $q(P(\beta, s)).$
\hyperref[subsec:transformed_distribution_continuity_fp]{Proof in Appendix \ref{subsec:transformed_distribution_continuity_fp}}.
\end{lemm}

\begin{lemm}
\label{lemm:quantile_perturbed_stoch_process}
Fix $\beta \in \mathcal{B}$. Let $\{t_{n}\}$ be a sequence such that $t_{n} \uparrow \infty$ as $n \rightarrow \infty.$ Under the conditions of Theorem \ref{theo:direct_effect}, if $b$ sufficiently small, 
\[
    \hat{S}^{t_{n}}_{b, n} \xrightarrow{p} s(\beta, b), \quad \hat{S}^{t_{n} - 1}_{b, n} \xrightarrow{p} s(\beta, b)\] where $s(\beta, b)$ is the unique fixed point of $q(P(\beta, s, b)).$
\hyperref[subsec:quantile_perturbed_stoch_process]{Proof in Appendix \ref{subsec:quantile_perturbed_stoch_process}}.
\end{lemm}


\begin{subsection}{Proof of Lemma \ref{lemm:loss_deriv}}
\label{subsec:loss_deriv}
Let $\Delta_{i} = Y_{i}(1) - Y_{i}(0).$ We have that

\begin{align*}
    &V_{\text{eq}}(\beta)  \\
    =& V(\beta, s(\beta), s(\beta)) \\
    =& \EE[F] { Y_{i} ( \mathbb{I}(\beta^{T} X_{i}(\beta, s(\beta))  \geq s(\beta)))} \\
    =& \EE[] { \EE[ \epsilon | \cdot ] {Y_{i}(1) \cdot \mathbb{I}(\beta^{T}X_{i}(\beta, s(\beta))  \geq s(\beta))}}  \\
    &+ \EE[F] { \EE[ \epsilon | \cdot] {Y_{i}(0) \cdot \mathbb{I}(\beta^{T}X_{i}(\beta, s(\beta))  < s(\beta))} } \\
    = &\EE[F] { Y_{i}(1) (1 - \Phi_{\sigma}(s(\beta) - \omega_{i}(s(\beta); \beta))} \\
    &+ \EE[F ] {Y_{i}(0)\Phi_{\sigma}(s(\beta) - \omega_{i}(s(\beta); \beta))  } \\
    = & \EE[F ] { Y_{i}(1) - \Delta_{i}  \cdot \Phi_{\sigma}(s(\beta) - \omega_{i}(s(\beta); \beta))}.
\end{align*}
Proposition \ref{prop:br_all} and Theorem \ref{theo:equilibrium} imply that $V_{\text{eq}}(\beta)$ continuously differentiable in $\beta$.
\end{subsection}


\begin{subsection}{Proof of Theorem \ref{theo:direct_effect}}
\label{subsec:direct_effect}
Let $s(\beta)$ be the equilibrium threshold induced by $\beta$. We introduce the following quantities.
\begin{align*}
    \tilde{Y}_{i}(\beta, s, r) &= Y_{i}(\pi(X_{i}(\beta, s);\beta, r)) \\
    \hat{V}_{n}(\beta, s, r) &:= \frac{1}{n} \sum_{i=1}^{n} \tilde{Y}_{i}(\beta, s, r).
\end{align*}

By Lemma \ref{lemm:loss_deriv}, we recall that $V(\beta, s, r)$ is continuously differentiable in $s$ and $r.$ 

The model gradient estimator $\hat{\tau}_{\text{MG}, n}^{t_{n}}(\beta, s, r)$ is the regression coefficient obtained by running OLS of $Y$ on $\mathbf{M}_{\beta}$. The regression coefficient must have the following form.
\begin{equation} 
\label{eq:ols_decomp}
\hat{\tau}_{\text{MG}, b_{n}, n}^{t_{n}}(\beta) = (\mathbf{S}_{zz}^{n})^{-1}\mathbf{s}_{zy}^{n},
\end{equation}
where $\mathbf{S}_{zz}^{n} := \frac{1}{b_{n}^{2} n}\mathbf{M}_{\beta}^{T}\mathbf{M}_{\beta}, \, \quad \mathbf{s}_{zy}^{n} := \frac{1}{b_{n}^{2} n} \mathbf{M}^{T}Y.$

In this proof, we establish convergence in probability of the two terms above separately. The bulk of the proof is the first step, which entails showing that
\[ \mathbf{s}_{zy}^{n} \xrightarrow{p} \frac{\partial V}{\partial \beta}(\beta, s(\beta), s(\beta)).\]
Due to $Y'$s dependence on the stochastic processes $\{ \hat{S}^{t_{n}-1}_{b_{n}, n} \}$ and $\{\hat{S}^{t_{n} }_{b_{n}, n} \}$, the main workhorse of this result is Lemma \ref{lemm:stochastic_equicontinuity_convergence_in_prob}. To apply this lemma, we must establish stochastic equicontinuity for the collection of stochastic processes $\{\hat{V}_{n}(\beta, s, r)\}$. Second, through a straightforward application of the Weak Law of Large Numbers, we show that
\[ \mathbf{S}_{zz} \xrightarrow{p} \mathbf{I}_{d}.\] Finally, we use Slutsky's Theorem to establish the convergence of the model gradient estimator.

We proceed with the first step of establishing convergence of $\mathbf{s}_{zy}^{n}.$ We have that
\begin{align*}
    \mathbf{s}_{zy}^{n} &= \frac{1}{b_{n}^{2} n} \mathbf{M}_{\beta}^{T}Y\\
    &= \frac{1}{b_{n}^{2} n} \sum_{i=1}^{n} b_{n}\zeta_{i}Y_{i} \\
    &= \frac{1}{b_{n}} \cdot \frac{1}{n} \sum_{i=1}^{n} \zeta_{i} Y_{i}.
\end{align*}
We fix $j$ and $b_{n}=b$ where $b >0$ and is small enough to satisfy the hypothesis of Lemma \ref{lemm:quantile_perturbed_stoch_process}. For each $\zeta \in \{-1, 1\}^{d}$ and $\xi \in \{-1, 1\}$, let \begin{align*}
    n_{\zeta, \xi} &= \sum_{i=1}^{n} \mathbb{I}(\zeta_{i}=\zeta, \xi_{i} = \xi).
\end{align*}

Let $z(\zeta)$ map a perturbation $\zeta \in \{-1, 1\}^{d}$ to the identical vector $\zeta$, except with $j$-th entry set to 0. So, if the $j$-th entry of $\zeta$ is 1, then $\zeta = \mathbf{e}_{j} + z(\zeta).$ If the $j$-th entry of $\zeta$ is $-1$, then $\zeta = -\mathbf{e}_{j} + z(\zeta).$ So, we have that
{
\begin{align*}
    Y_{i} &= \tilde{Y}_{i}( \beta + b \zeta_{i}, \hat{S}^{t_{n}-1}_{n} + b \xi_{i}, \, \hat{S}^{t_{n}}_{b, n} + b\xi_{i}) \\
    &=\tilde{Y}_{i}( \beta +  b\zeta_{i, j} \mathbf{e}_{j} + b\cdot z(\zeta_{i}), \, \hat{S}^{t_{n}-1}_{n} + b \xi_{i}, \hat{S}^{t_{n}}_{b, n} + b\xi_{i}).
\end{align*}
}%
As a result, we have that
{
\begin{align}
    &\frac{1}{n}\sum_{i=1}^{n}\zeta_{i, j} Y_{i} \\
    &= \frac{1}{n} \sum_{i=1}^{n} \zeta_{i, j} \cdot \tilde{Y}_{i}( \beta +  b\zeta_{i, j} \mathbf{e}_{j} + b\cdot z({\zeta}_{i}), \, \hat{S}^{t_{n}-1}_{b, n} + b \xi_{i}, \hat{S}^{t_{n}}_{b, n} + b\xi_{i})  \\
    =& \sum_{\substack{\zeta \in \{-1, 1\}^{d} \text{ s.t. } \zeta_{j} = 1 \\ \xi \in \{-1, 1\}}} \frac{n_{\zeta, \xi}}{n} \sum_{i=1}^{n_{\zeta, \xi}} \tilde{Y}_{i}(\beta + b\mathbf{e}_{j} + b \cdot z(\zeta), \,  \hat{S}^{t_{n}-1}_{b, n} + b \xi, \hat{S}^{t_{n}}_{b, n} + b\xi) \label{eq:s_zy_part_a} \\
    &- \sum_{\substack{\zeta \in \{-1, 1\}^{d} \text{ s.t. } \zeta_{j} = -1 \\ \xi \in \{-1, 1\}}} \frac{n_{\zeta, \xi}}{n} \sum_{i=1}^{n_{\zeta, \xi}} \tilde{Y}_{i}(\beta - b\mathbf{e}_{j} + b\cdot z(\zeta), \,   \hat{S}^{t_{n}-1}_{b, n} + b \xi, \hat{S}^{t_{n}}_{b, n} + b\xi) \label{eq:s_zy_part_b}
\end{align}
}%


To establish convergence properties of each term in the double sum in \eqref{eq:s_zy_part_a} and \eqref{eq:s_zy_part_b}, we must establish stochastic equicontinuity of the collection of stochastic processes $\{ \hat{V}_{n}(\beta, s, r) \}$ indexed by $(s, r) \in \mathcal{S} \times \mathcal{S}$. Because $\mathcal{S} \times \mathcal{S}$ compact and $V(\beta, s, r)$ is continuous in $s$ and $r$, we can show that $\{\hat{V}_{n}(\beta, s, r)\}$ is stochastically equicontinuous by establishing that $\hat{V}_{n}(\beta, s, r)$ converges uniformly in probability (with respect to $(s, r)$) to $V(\beta, s, r)$ (Lemma \ref{lemm:uniform_convergence_implies_stochastic_equicontinuity}). 

We show that $\hat{V}_{n}(\beta, s, r)$ converges uniformly (with respect to $(s, r)$) in probability to $V(\beta, s, r)$. We realize that
{
\begin{align}
 &\sup_{(s, r) \times \mathcal{S} \times \mathcal{S}} |\hat{V}_{n}(\beta, s, r) - V(\beta, s, r)| \\
 &= \sup_{(s, r) \times \mathcal{S} \times \mathcal{S}}  \Big| \frac{1}{n} \sum_{i=1}^{n} \tilde{Y}_{i}(\beta, s, r) - V(\beta, s, r) \Big| \\
 &= \sup_{(s, r) \times \mathcal{S} \times \mathcal{S}}  \Big| \frac{1}{n} \sum_{i=1}^{n} m(Z_{i}, c_{i}; \pi(X_{i}(\beta, s);\beta, r)) + \rho_{i}  - V(\beta, s, r) \Big| \label{eq:y_def} \\
 &\leq \sup_{(s, r) \times \mathcal{S} \times \mathcal{S}}  \Big| \frac{1}{n} \sum_{i=1}^{n} m(Z_{i}, c_{i}; \pi(X_{i}(\beta, s);\beta, r))  - V(\beta, s, r) \Big| + \sup_{(s, r) \times \mathcal{S} \times \mathcal{S}} |\frac{1}{n} \sum_{i=1}^{n} \rho_{i}|.
 \end{align}
 }%
 \eqref{eq:y_def} follows from Assumption \ref{assumption:nice_loss}. By Lemma \ref{lemm:nice_loss_and_pi}, we have that $m$ satisfies the conditions of Lemma \ref{lemm:uniform_convergence_sum}. So, we have that 
 { 
\begin{align*}
&\sup_{(s, r) \in \mathcal{S} \times \mathcal{S}} \Big|\frac{1}{n} \sum_{i=1}^{n} m(Z_{i}, c_{i}; \pi(X_{i}(\beta, s);\beta, r))  - \EE[F]{m(Z_{i}, c_{i}; \pi(X_{i}(\beta, s);\beta, r))}\Big|\\
&= \sup_{(s, r) \in \mathcal{S} \times \mathcal{S}} \Big|\frac{1}{n} \sum_{i=1}^{n} m(Z_{i}, c_{i}; \pi(X_{i}(\beta, s);\beta, r))  -  V(\beta, s, r) \Big| \xrightarrow{p} 0.
\end{align*}
}%
Furthermore, under Assumption \ref{assumption:nice_loss}, $\rho_{i}$ is mean-zero, so by the Weak Law of Large Numbers, $\frac{1}{n}\sum_{i=1}^{n} \rho_{i} \xrightarrow{p} 0$. Because $\rho_{i}$ does not depend on $(s, r)$, this convergence also holds uniformly in $(s, r)$. Combining these results yields that $\hat{V}_{n}(\beta, s, r)$ converges uniformly (with respect to $(s, r)$) in probability to $V(\beta, s, r).$

As a consequence, the collection of stochastic processes $\{\hat{V}_{n}(\beta, s, r)\}$ is stochastically equicontinuous. In particular, $\hat{V}_{n}(\beta, s, r)$ is stochastically equicontinuous at $(s(\beta, b), s(\beta, b))$, where $s(\beta, b)$ is the unique fixed point of $q(P(\beta, s, b))$ (see Lemma \ref{lemm:transformed_distribution_continuity_fp}). By Lemma \ref{lemm:quantile_perturbed_stoch_process}, we have that
\begin{align*}
    \hat{S}^{t_{n}-1}_{b, n} &\xrightarrow{p} s(\beta, b) \\
    \hat{S}^{t_{n}}_{b, n} &\xrightarrow{p} s(\beta, b).
\end{align*}
Now, we can apply Lemma \ref{lemm:stochastic_equicontinuity_convergence_in_prob} to establish convergence in probability for each term in the double sum of \eqref{eq:s_zy_part_a}, \eqref{eq:s_zy_part_b}. As an example, for a perturbation $\zeta \in \{-1, 1\}^{d}$ with $j$-th entry equal to $1$ and arbitrary $\xi \in \{-1, 1\}$, Lemma \ref{lemm:stochastic_equicontinuity_convergence_in_prob} gives that
{
\begin{align*}
    &\hat{V}_{n_{\zeta, \xi}}(\beta + b \mathbf{e}_{j} + b \cdot z(\zeta),\, \hat{S}^{t_{n}-1}_{b_{n}, n} + b\xi,\, \hat{S}^{t_{n}}_{b, n} + b\xi)\\ &\xrightarrow{p} \hat{V}_{n_{\zeta, \xi}}(\beta + b \mathbf{e}_{j} + b \cdot z(\zeta),\, s(\beta, b) +b\xi, s(\beta, b)+b\xi),
\end{align*}
}%
and by the Weak Law of Large Numbers, we have that
{
\begin{align*}
    &\hat{V}_{n_{\zeta, \xi}}(\beta + b \mathbf{e}_{j} + b\cdot z(\zeta),\, s(\beta, b)+b\xi, s(\beta, b)+b\xi)\\
    &\xrightarrow{p} V(\beta + b\mathbf{e}_{j} + b\cdot z(\zeta),\, s(\beta, b)+b\xi, s(\beta, b)+b\xi).
\end{align*}
}%
Analogous statements hold for the remaining terms in \eqref{eq:s_zy_part_a} and \eqref{eq:s_zy_part_b}. Also,
\[     \frac{n_{\zeta, \xi}}{n} \xrightarrow{p} \frac{1}{2^{d+1}}, \quad \zeta \in \{-1, 1\}^{d}, \xi \in \{-1, 1\}.\]  
By Slutsky's Theorem, when any $j$ and $b$ fixed, we have
{
\begin{align}
    &\mathbf{s}_{zy, j}^{n} \\
    &\xrightarrow{p} \sum_{\substack{\zeta \in \{-1, 1\}^{d} \text{ s.t. } \zeta_{j} = 1 \\ \xi \in \{-1, 1\}}} \frac{V(\beta + b \mathbf{e}_{j} + b \cdot z(\zeta),s(\beta, b)+b\xi, s(\beta, b)+b\xi) }{2^{d+1} \cdot b} \label{eq:r_b1}\\
    &\indent- \sum_{\substack{\zeta \in \{-1, 1\}^{d} \text{ s.t. } \zeta_{j} = -1 \\ \xi \in \{-1, 1\}}} \frac{V(\beta - b \mathbf{e}_{j} + b \cdot z(\zeta),s(\beta, b)+b\xi, s(\beta, b)+b\xi) }{2^{d+1} \cdot b}. \label{eq:r_b2}
\end{align}
}
Let $R_{b}$ denote the expression on the right side of the above equation. If there is a sequence $\{b_{n}\}$ such that $b_{n} \rightarrow 0$, then by Lemma \ref{lemm:transformed_distribution_continuity_fp},  $s(\beta, b_{n}) \rightarrow s(\beta)$, where $s(\beta)$ is the unique fixed point of $q(P(\beta, s)).$ Furthermore, by the continuity of $V$, we have that
\[R_{b_{n}} \rightarrow \frac{\partial V}{\partial \beta_{j}}(\beta, s(\beta), s(\beta)).\]
Using the definition of convergence in probability, we show that there exists such a sequence $\{b_{n}\}.$ From \eqref{eq:r_b1} and \eqref{eq:r_b2}, we have that for each $\epsilon, \delta > 0$ and $b> 0$ and sufficiently small, there exists $n(\epsilon, \delta, b)$ such that for $n \geq n(\epsilon, \delta, b)$
\[ P( | \mathbf{s}_{zy, j}^{n} - R_{b}| \leq \epsilon) \geq  1-\delta.\]
So, we can fix $\delta >0.$ For $k= 1, 2, \dots,$ let $N(k) = n(\frac{1}{k}, \delta, \frac{1}{k})$. Then, we can define a sequence such that $b_{n} = \epsilon_{n} = \frac{1}{k}$ for all $N(k) \leq n \leq N(k+1).$ So, we have that $\epsilon_{n} \rightarrow 0$ and $b_{n} \rightarrow 0.$ Finally, this gives that 
\[ \mathbf{s}_{zy, j}^{n} \xrightarrow{p} \frac{\partial V}{\partial \beta_{j}}(\beta, s(\beta), s(\beta)).\]
Considering all indices $j$,
\[ \mathbf{s}_{zy}^{n} \xrightarrow{p} \frac{\partial V}{\partial \beta}(\beta, s(\beta), s(\beta)).\]
It remains to establish convergence in probability for $\mathbf{S}_{zz}^{n}.$ We have that
\begin{align*}
    &\mathbf{S}^{n}_{zz} \\
    &= \frac{1}{b_{n}^{2}n}\mathbf{M}_{\beta}^{T}\mathbf{M}_{\beta} \\
    &= \frac{1}{b_{n}^{2} n} \sum_{i=1}^{n} (b_{n}\zeta_{i})^{T} (b_{n}\zeta_{i}) =\frac{1}{n}\sum_{i=1}^{n} \zeta_{i}^{T} \zeta_{i}.
\end{align*}
We note that 
\[\EE[\zeta_{i} \sim R^{d}]{\zeta_{i, j} \zeta_{i, k}} = \begin{cases} 1 \text{ if } j=k \\ 0 \text{ if } j \neq k \end{cases}\] because $\zeta_{i}$ is a vector of independent Rademacher random variables. So, $\EE{\zeta_{i}^{T}\zeta_{i}} = \mathbf{I}_{d}.$ By the Weak Law of Large Numbers, we have that 
\[ \mathbf{S}_{zz}^{n} \xrightarrow{p} \mathbf{I}_{d}.\]
Finally, we can use Slutsky's Theorem to show that
\begin{align*}
\hat{\tau}_{\text{MG},b_{n}, n}^{t_{n}}(\beta) &= (\mathbf{S}_{zz}^{n})^{-1}\mathbf{s}_{zy}^{n} \\
&\xrightarrow{p} (\mathbf{I}_{d})^{-1}\frac{\partial V}{\partial \beta}(\beta, s(\beta), s(\beta)) \\
&= \frac{\partial V}{\partial \beta}(\beta, s(\beta), s(\beta)) \\
&= \tau_{\text{MG}}(\beta).
\end{align*}
\end{subsection}

\begin{subsection}{Proof of Theorem \ref{theo:indirect_effect}}
\label{subsec:indirect_effect}
Let $s(\beta) = s(\beta)$. Let $\hat{\Gamma}^{b, n}_{Y, s, r}, \hat{\Gamma}^{b, n}_{\Pi, s}, \hat{\Gamma}^{b, n}_{\Pi, \beta}$ be the regression coefficients defined in Section \ref{sec:results_learning}. Let $p^{n}(\beta, s, b)(s)$ be the density estimate defined in Section \ref{sec:results_learning}. In this proof, we rely on the results on the following convergence results for these estimators.
\begin{coro}
\label{coro:dl_ds}
 Let $\{t_{n}\}$ be a sequence such that $t_{n} \uparrow \infty$ as $n \rightarrow \infty.$ Let $\mathbf{M}_{s}, Y, \hat{\Gamma}^{b, n}_{Y, s, r}(\beta, s, r)$ be as defined in Section \ref{sec:results_learning}. Under the conditions of Theorem \ref{theo:direct_effect}, there exists a sequence $\{b_{n}\}$ such that $b_{n} \rightarrow 0$ so that
\begin{align} \label{eq:dL_ds} \begin{split} &\hat{\Gamma}^{b_{n}, n}_{Y, s, r}(\beta, \hat{S}^{t_{n}-1}_{b_{n}, n}, \hat{S}^{t_{n}}_{b_{n}, n}) \\
&\xrightarrow{p} \frac{\partial V}{\partial s}(\beta, s(\beta), s(\beta)) + \frac{\partial V}{\partial r}(\beta, s(\beta), s(\beta)). \end{split} \end{align} \hyperref[subsec:dl_ds]{Proof in Appendix \ref{subsec:dl_ds}}.
\end{coro}

\begin{lemm}
\label{lemm:dpi_dbeta}
Let $\{t_{n}\}$ be a sequence such that $t_{n} \uparrow \infty$ as $n \rightarrow \infty.$ Let $\mathbf{M}_{\beta}, I, \hat{\Gamma}^{b, n}_{\Pi, \beta}(\beta, s, r)$ be as defined Section \ref{sec:results_learning}. Under the conditions of Theorem \ref{theo:direct_effect}, there exists a sequence $\{b_{n}\}$ such that $b_{n} \rightarrow 0$ so that
\begin{align} \label{eq:dpi_dbeta}\begin{split} &\hat{\Gamma}^{b_{n}, n}_{\Pi, \beta}(\beta, \hat{S}^{t_{n}}_{b_{n}, n}, \hat{S}^{t_{n}}_{b_{n}, n}) \\
&\xrightarrow{p} \frac{\partial \Pi}{\partial \beta}(\beta, s(\beta); s(\beta)). \end{split}\end{align} 
\hyperref[subsec:dpi_dbeta]{Proof in Appendix \ref{subsec:dpi_dbeta}}.
\end{lemm}

\begin{coro}
\label{coro:dpi_ds}
Let $\{t_{n}\}$ be a sequence such that $t_{n} \uparrow \infty$ as $n \rightarrow \infty.$ Let $\mathbf{M}_{s}, I, \hat{\Gamma}^{b, n}_{\Pi, s}(\beta, s, r)$ be as defined in Section \ref{sec:results_learning}. Under the conditions of Theorem \ref{theo:direct_effect}, there exists a sequence $\{b_{n}\}$ such that $b_{n} \rightarrow 0$ so that
\begin{equation} \hat{\Gamma}^{b_{n}, n}_{\Pi, s}(\beta, \hat{S}^{t_{n}}_{b_{n}, n}, \hat{S}^{t_{n}}_{b_{n}, n}) \xrightarrow{p} \frac{\partial \Pi}{\partial s}(\beta, s(\beta); s(\beta)). \label{eq:dpi_ds} \end{equation} \hyperref[subsec:dpi_ds]{Proof in Appendix \ref{subsec:dpi_ds}}.
\end{coro}

\begin{lemm}
\label{lemm:density_converges}
Fix $\beta \in \mathcal{B}$. Let $\{h_{n}\}$ be a sequence such that $h_{n} \rightarrow 0$ and $nh_{n} \rightarrow \infty$. Let $p^{n}(\beta, s, b)(r)$ denote a kernel density estimate of $p(\beta, s, b)(r)$ with kernel function $k(z) = \mathbb{I}(z \in [- \frac{1}{2}, \frac{1}{2})$ and bandwidth $h_{n}.$ Let $\{t_{n}\}$ be a sequence such that $t_{n} \uparrow \infty$ as $n \rightarrow \infty.$ Under the conditions of Theorem \ref{theo:direct_effect}, there exists a sequence $\{b_{n}\}$ such that $b_{n} \rightarrow 0$ so that
\begin{equation}
\label{eq:density_converges}
p^{n}_{\beta, \hat{S}^{t_{n}}_{b_{n}, n}, b_{n}}(\hat{S}^{t_{n}}_{b_{n}, n}) \xrightarrow{p} p(\beta, s(\beta))(s(\beta)),\end{equation} where $s(\beta)$ is the unique fixed point of $q(P(\beta, s)).$ \hyperref[subsec:density_converges]{Proof in Appendix \ref{subsec:density_converges}}.
\end{lemm}

Finally, we use the following lemma to show that we recover the equilibrium gradient. 

\begin{lemm}
\label{lemm:ds_dbeta}
Under Assumptions \ref{assumption:strong_convexity}, 
\ref{assumption:finite_types}, and
\ref{assumption:br_interior}, if $\sigma^{2} > \frac{1}{\alpha_{*} \sqrt{2\pi e}}$ then
{
\begin{align} 
\label{eq:ds_db}
\begin{split}
\frac{\partial s}{\partial \beta} = \frac{1}{p(\beta, s(\beta))(s(\beta)) - \frac{\partial \Pi}{\partial s}(\beta, s(\beta); s(\beta))} \cdot \frac{\partial \Pi}{\partial \beta}(\beta, s(\beta); s(\beta)),\end{split}\end{align}
}%
where $s(\beta) = s(\beta)$, the unique fixed point induced by the model $\beta$. \hyperref[subsec:ds_dbeta]{Proof in Appendix \ref{subsec:ds_dbeta}}.
\end{lemm}

We proceed with the main proof. The equilibrium gradient estimator in \eqref{eq:ie_estimator} consists of two terms. We see that the convergence of the first term is immediately given by \eqref{eq:dL_ds} above. It remains to show that the second term converges in probability to $\frac{\partial s}{\partial \beta}(\beta)$. We have that
{
\begin{align}
    &\frac{\hat{\Gamma}^{b_{n}, n}_{\Pi, \beta}(\beta, \hat{S}^{t_{n}}_{b_{n}, n}; \hat{S}^{t_{n}}_{b_{n}, n})}{p^{n}_{\beta, \hat{S}^{t_{n}}_{b_{n}, n}, b_{n}}(\hat{S}^{t_{n}}_{b_{n}, n}) - \hat{\Gamma}^{b_{n}, n}_{\Pi, s}(\beta, \hat{S}^{t_{n}}_{n}; \hat{S}^{t_{n}}_{b_{n}, n}) } \\
    &\xrightarrow{p} \frac{1}{p(\beta, s(\beta))(s(\beta)) - \frac{\partial \Pi}{\partial s}(\beta, s(\beta); s(\beta))} \cdot \frac{\partial \Pi}{\partial \beta}(\beta, s(\beta); s(\beta)) \label{eq:ds_db_1}\\
    &= \frac{\partial s}{\partial \beta}(\beta). \label{eq:ds_db_2}
\end{align}
}%

\eqref{eq:ds_db_1} follows by Slutsky's Theorem given  \eqref{eq:dpi_ds}, \eqref{eq:dpi_dbeta}, and \eqref{eq:density_converges}. \eqref{eq:ds_db_2} follows from Lemma \eqref{lemm:ds_dbeta}. Combining \eqref{eq:dL_ds} and \eqref{eq:ds_db_2} using Slutsky's Theorem, yields
{
\begin{align*}
&\hat{\tau}_{\text{EG}, n}^{t_{n}}(\beta) \\
&= \hat{\Gamma}^{b_{n}, n}_{Y, s, r}(\beta, \hat{S}^{t_{n}-1}_{b_{n}, n},  \hat{S}^{t_{n}}_{b_{n}, n}) \cdot  \frac{\hat{\Gamma}^{b_{n}, n}_{\Pi, \beta}(\beta, \hat{S}^{t_{n}}_{b_{n}, n}; \hat{S}^{t_{n}}_{n})}{p^{n}_{\beta, \hat{S}^{t_{n}}_{b_{n}, n}, b_{n}}(\hat{S}^{t_{n}}_{b_{n}, n}) - \hat{\Gamma}^{b_{n}, n}_{\Pi, s}(\beta, \hat{S}^{t_{n}}_{b_{n}, n}; \hat{S}^{t_{n}}_{b_{n}, n}) }  \\
&\xrightarrow{p} \Big(\frac{\partial V}{\partial s}(\beta, s(\beta), s(\beta))  + \frac{\partial V}{\partial r}(\beta, s(\beta), s(\beta)\Big)\cdot \frac{\partial s}{\partial \beta}(\beta) \\
&= \tau_{\text{EG}}(\beta).
\end{align*}
}%
\end{subsection}

\begin{subsection}{Proof of Corollary \ref{coro:total}}
\label{subsec:total}
This result follows from applying Slutsky's Theorem to the results of Theorem \ref{theo:direct_effect} and Theorem \ref{theo:indirect_effect}.
\end{subsection}
\end{section}

\begin{section}{Proofs of Technical Results}
\label{sec:technical_proofs}
\begin{subsection}{Proof of Lemma \ref{lemm:uniform_convergence_root}}
\label{subsec:uniform_convergence_root}
Since $f_{n} \rightarrow f$ uniformly and $f_{n}$'s are defined on a compact domain, then $f$ must be continuous. By assumption, $f$ has only one zero $x_{i}^{*}$ in $[a, b]$. We can choose \[\epsilon = \inf \{ |f(x)| \mid |x- x_{i}^{*}| > \delta \}.\]  By uniform convergence, there exists $N$ such that for $n \geq N$, $\sup_{x \in \mathcal{X}} |f_{n}(x) - f(x)| < \frac{\epsilon}{2}.$ By the triangle inequality we have that
\begin{align*}
    |f(x) | &= |f(x) - f_{n}(x) + f_{n}(x)| \\
    &\leq |f(x) - f_{n}(x)| + |f_{n}(x)|\\
    |f_{n}(x)| &\geq |f(x)| - |f(x) - f_{n}(x)|.
\end{align*}
For $n \geq N$ and $x$ such that $|x - x_{i}^{*}| > \delta$, we realize that $|f_{n}(x) | > \frac{\epsilon}{2}.$ So $x$ cannot be a fixed point of $f_{n}.$ Thus, if $x$ is a fixed point of $f_{n}$, then we have that $|x_{n} - x_{i}^{*}| < \delta.$ This implies that $x_{n} \rightarrow x_{i}^{*}.$
\end{subsection}

\begin{subsection}{Proof of Lemma \ref{lemm:uniform_convergence_maximizer}}
\label{subsec:uniform_convergence_maximizer}
By uniform convergence, we have that for any $\epsilon > 0$, for every $x \in \mathcal{X}$, there is $n \geq N$ so that
\[ f(x) - \epsilon < f_{n}(x) < f(x) + \epsilon. \]
So, for all $x \in \mathcal{X}$, we have that $f_{n}(x) < f(x) + \epsilon < x_{i}^{*} + \epsilon.$ In addition, for all $x \in \mathcal{X}$, we have that $f(x) - \epsilon < x_{n}$. We realize that this implies that $x_{n} \leq x + \epsilon$ and $x_{i}^{*} - \epsilon \leq x_{n}.$ Thus, we have that $|x_{n} - x| < \epsilon.$ So, we have that $x_{n} \rightarrow x_{i}^{*}.$
\end{subsection}

\begin{subsection}{Proof of Lemma \ref{lemm:diff_concavity_utility}}
\label{subsec:diff_concavity_utility}
From Assumption \ref{assumption:strong_convexity}, we have that $c_{i}$ is twice continuously differentiable. Since $\Phi_{\sigma}$ is the Normal CDF, we have that it is twice continuously differentiable. Since the composition and sum of twice continuously differentiable functions is also twice continuously differentiable, we have that $\EE[\epsilon]{U_{i}(x; \beta, s)}$ is twice continuously differentiable in $x, \beta, s$.

In addition, we can show that when the noise has sufficiently high variance, $\nabla^{2} \EE[\epsilon]{U_{i}(x; \beta, s)}$ is strictly concave in $x$. We have that
{\small
\begin{align*}
    \nabla_{x} \EE[\epsilon]{U_{i}(x; \beta, s)} &= - \nabla c_{i}(x - Z_{i}) + \phi_{\sigma}(s - \beta^{T}x) \beta^{T}. \\
    \nabla_{x}^{2} \EE[\epsilon]{U_{i}(x; \beta, s)} &= - \nabla^{2} c_{i}(x- Z_{i}) - \beta \phi_{\sigma}'( s- \beta^{T}x)  \beta^{T}.
\end{align*}
}

We can show that $-\nabla^{2}_{x} \EE[\epsilon]{U_{i}(x; \beta, s)}$ is positive definite at any point. Let $z \in \mathbb{S}^{d-1},$
\begin{align}
    &z^{T}(- \nabla^{2}_{x} \EE[\epsilon]{U_{i}(x; \beta, s)} ) z \\
    &= z^{T} [\nabla^{2} c_{i}( x - Z_{i}) + \beta \phi_{\sigma}'(s - \beta^{T}x) \beta^{T} ] z \\
    &= z^{T} \nabla^{2} c_{i}( x - Z_{i}) z + \phi_{\sigma}'(s - \beta^{T}x) \cdot z^{T}  \beta \beta^{T}  z \label{eq:pd_1} \\
    &\geq \inf_{y \in \mathbb{R}^{d}} z^{T} \nabla^{2} c_{i}(y) z + \inf_{y \in \mathbb{R}} \phi_{\sigma}'( y) \cdot  z^{T} \beta \beta^{T} z \label{eq:pd_2} \\ 
    &\geq \alpha_{i} + (-\frac{1}{\sigma^{2} \sqrt{2\pi e}}) \cdot z^{T} \beta \beta^{T} z \label{eq:pd_3} \\
    &\geq \alpha_{i} + (-\alpha_{i}) \cdot z^{T} \beta \beta^{T} z \label{eq:pd_4} \\
    &> 0  \label{eq:pd_5}
\end{align}

We check the above inequality as follows. By Assumption \ref{assumption:strong_convexity}, $c_{i}$ is $\alpha_{i}$-strongly convex and twice differentiable. So, we can lower bound the first term in \eqref{eq:pd_1}. \eqref{eq:pd_3} holds because $- \frac{1}{\sigma^{2} \sqrt{2\pi e}} \leq \phi_{\sigma}'(y) \leq \frac{1}{\sigma^{2} \sqrt{2\pi e}}$. The assumption that $\sigma^{2} > \frac{1}{\alpha_{i} \sqrt{2\pi e}}$ yields \eqref{eq:pd_4}. Finally, $z^{T}\beta\beta^{T} z = (z^{T}\beta)^{2}$, and the dot product of two unit vectors, $\beta$ and $z$, must be between -1 and 1, so $0 \leq z^{T}\beta\beta^{T}z \leq 1.$ Thus, $-\nabla^{2}_{x} \EE[\epsilon]{U_{i}(x; \beta, s)}$ is positive definite. So, $\EE[\epsilon]{U_{i}(x; \beta, s)}$ is strictly concave in $x$.

Finally, given that $\EE[\epsilon]{U_{i}(x; \beta, s)}$ is twice differentiable and strictly concave, we can apply \ref{lemm:maximizer_characterization} to establish the last claim.
\end{subsection}

\begin{subsection}{Proof of Lemma \ref{lemm:score_gradient}}
\label{subsec:score_gradient}
Without loss of generality, we fix $\beta$. We use the following abbreviations \begin{align*}
    \EE[\epsilon]{U_{i}(x; s)} &:= \EE[\epsilon]{U_{i}(x; \beta, s)} \\
    x_{i}^{*}(s)&:=x_{i}^{*}(\beta, s) \\
    h_{i}(s)&:=h_{i}(s; \beta). 
\end{align*}
We state an additional lemma that will be used in the proof of Lemma \ref{lemm:score_gradient}. 
\begin{lemm}
\label{lemm:apply_sherman}
Let $x \in \mathcal{X}, \beta \in \mathcal{B}, s \in \mathbb{R}$. For an agent $i$ with unobservables $(Z_{i}, c_{i})\sim F$, define $\bfH = \nabla^{2} c_{i}(x - Z_{i})$. Under Assumption \ref{assumption:strong_convexity}, we have that 
\begin{align*} 
&(\bfH + \phi_{\sigma}'(s - \beta^{T} x) \beta \beta^{T})^{-1} \\
&= \bfH^{-1} - \frac{ \phi_{\sigma}'(s - \beta^{T} x)  \beta \beta^{T} \bfH^{-1}}{1 + \phi_{\sigma}'(s - \beta^{T} x) \beta^{T} \bfH^{-1} \beta}. \end{align*}
\hyperref[subsec:apply_sherman]{Proof in Appendix \ref{subsec:apply_sherman}}.
\end{lemm}

Now, we proceed with the main proof. We compute $\nabla_{s} x_{i}^{*}$ by using the implicit expression for $x_{i}^{*}(s)$ given by the first-order condition in Lemma \ref{lemm:diff_concavity_utility}; we note that $x_{i}^{*}(s)$ must satisfy $\nabla_{x} \EE[\epsilon]{U_{i}(x;s)} = 0$, so

\[ \nabla_{x} \EE[\epsilon]{U_{i}(x;s)} = - \nabla c_{i}(x - Z_{i}) + \phi_{\sigma}( s - \beta^{T} x) \beta^{T}.\]

So, the best response $x_{i}^{*}(s)$ satisfies
\[ - \nabla c_{i}(x_{i}^{*}(s) - Z_{i}) + \phi_{\sigma}( s - \beta^{T} x_{i}^{*}(s)) \beta^{T} = 0.\]

From Proposition \ref{prop:br_all}, we have that the best response $x_{i}^{*}(s)$ is continuously differentiable in $s$, so we can differentiate the above equation with respect to $s$. This yields the following equation

\begin{align*} &(\nabla^{2} c_{i}( x_{i}^{*}(s) - Z_{i}) +  \phi_{\sigma}'(s - \beta^{T}x_{i}^{*}(s)) \beta \beta^{T}) \nabla_{s} x_{i}^{*} \\
&= \phi_{\sigma}'(s - \beta^{T}x_{i}^{*}(s)) \beta.
\end{align*}

Let $\bfH = \nabla^{2} c_{i}( x_{i}^{*}(s) - Z_{i}).$ The above equation can be rewritten as
\begin{align*} 
&(\bfH + \phi_{\sigma}'(s - \beta^{T}x_{i}^{*}(s)) \beta \beta^{T}) \nabla_{s} x_{i}^{*} \\
&= \phi_{\sigma}'(s - \beta^{T}x_{i}^{*}(s)) \beta. \end{align*}

From Lemma \ref{lemm:apply_sherman}, we realize that the matrix term on the left side of the equation is invertible. We multiply both sides of the equation by the inverse of the matrix to compute $\nabla_{s} x_{i}^{*}$. 
{
\begin{align*} 
&\nabla_{s} x_{i}^{*} \\
&= (\bfH + \phi_{\sigma}'(s - \beta^{T}x_{i}^{*}(s)) \beta \beta^{T})^{-1} \phi_{\sigma}'(s - \beta^{T}x_{i}^{*}(s)) \beta. \end{align*}
}%

We can substitute the expression for $(\bfH + \phi_{\sigma}'(s - \beta^{T}x_{i}^{*}(s)) \beta \beta^{T})^{-1}$ from Lemma \ref{lemm:apply_sherman} into the above equation. 

{
\begin{align*}&\nabla_{s} x_{i}^{*} \\
&= \Big(\bfH^{-1} - \frac{\phi_{\sigma}'(s -\beta^{T}x_{i}^{*}(s)) \bfH^{-1} \beta \beta^{T} \bfH^{-1} }{1 + \phi_{\sigma}'(s - \beta^{T} x_{i}^{*}(s)) \beta^{T} \bfH^{-1} \beta }\Big) \phi_{\sigma}'(s - \beta^{T}x_{i}^{*}(s)) \beta. \end{align*}}%

This gives us that
{
\begin{align}
    &\beta^{T} \nabla_{s} x_{i}^{*} \\
    &= \beta^{T}\Big(\bfH^{-1} - \frac{\phi_{\sigma}'(s -\beta^{T}x_{i}^{*}(s)) \bfH^{-1} \beta \beta^{T} \bfH^{-1} }{1 + \phi_{\sigma}'(s - \beta^{T} x_{i}^{*}(s)) \beta^{T} \bfH^{-1} \beta }\Big) \phi_{\sigma}'(s - \beta^{T}x_{i}^{*}(s)) \beta \\
    &= \phi_{\sigma}'(s - \beta^{T}x_{i}^{*}(s)) \beta^{T}\bfH^{-1}\beta - \Big( \frac{(\phi_{\sigma}'(s - \beta^{T} x_{i}^{*}(s)) \beta^{T} \bfH^{-1} \beta)^{2} }{1 + \phi_{\sigma}'(s - \beta^{T} x_{i}^{*}(s)) \beta^{T} \bfH^{-1} \beta} \Big) \\
    &= \frac{\phi_{\sigma}'(s - \beta^{T} x_{i}^{*}(s)) \beta^{T} \bfH^{-1} \beta }{1 + \phi_{\sigma}'(s - \beta^{T} x_{i}^{*}(s)) \beta^{T} \bfH^{-1} \beta}.
\end{align}
}%
as desired.

Finally, we can show that $h_{i}(s; \beta)$ is strictly increasing by showing that its derivative is always positive. We have that $h_{i}'(s; \beta) = 1 - \beta^{T}\nabla_{s} x_{i}^{*}.$ 

We can view $\beta^{T} \nabla_{s} x_{i}^{*}$ as a function of the form $f(s) = \frac{\tilde{f}(s)}{\tilde{f}(s) + 1}$. For any choice of $s$ and $\tilde{f}$, $f(s) < 1.$ Thus, we can conclude $\beta^{T} \nabla_{s} x_{i}^{*} < 1$, so $h_{i}(s; \beta)$ is strictly increasing.

\end{subsection}

\begin{subsection}{Proof of Lemma \ref{lemm:limits_br}}
\label{subsec:limits_br}
Define $\bar{u}_{i}(x; \beta) = \lim_{s \rightarrow \infty} \EE[\epsilon]{U_{i}(x; \beta, s)}.$ Note that
\[ \bar{u}_{i}(x; \beta) = -c_{i}(x-Z_{i}) + 1.\]
We realize that 
\[ \argmax_{x \in \mathcal{X}} \bar{u}_{i}(x; \beta) = Z_{i}.\]

To show that \eqref{eq:up} holds, we establish that $\EE[\epsilon]{U_{i}(x; \beta, s)} \rightarrow \bar{u}(x; \beta)$ uniformly in $x$ for $x \in \mathcal{X}$ as $s \rightarrow \infty.$ Then, we show that the maximizer of $\EE[\epsilon]{U_{i}(x; \beta, s)}$ must converge to the maximizer of $\bar{u}_{i}(x; \beta)$ as $s \rightarrow \infty,$ which gives the desired result.

First, we verify the conditions of Lemma \ref{lemm:uniform_convergence_concave} to establish the uniform convergence of $\EE[\epsilon]{U_{i}(x; \beta, s)}$. We note that $\mathcal{X}$ is compact. In addition, for every $s$, we have that the expected utility is continuous and strictly concave (Lemma \ref{lemm:diff_concavity_utility}). Also, $\bar{u}(x; \beta)$ is continuous and concave. We note that $\EE[\epsilon]{U_{i}(x; \beta, s)} \rightarrow \bar{u}(x; \beta)$ pointwise in $x$ as $s \rightarrow \infty.$ Thus, Lemma \ref{lemm:uniform_convergence_concave} implies that $\EE[\epsilon]{U_{i}(x; \beta, s)} \rightarrow \bar{u}(x; \beta)$ converges uniformly in $x$ for $x \in \mathcal{X}$ as $s \rightarrow \infty.$

Second, we verify the conditions of  Lemma \ref{lemm:uniform_convergence_maximizer} to show that \begin{equation}\label{eq:maximizer_converge} \lim_{s \rightarrow \infty} x_{i}^{*}(\beta, s) \rightarrow Z_{i}. \end{equation}  We note that $\EE[\epsilon]{U_{i}(x; \beta, s)}$ has a unique maximizer $x_{i}^{*}(\beta, s)$ for every $s$ (Proposition \ref{prop:br_all}), and $\bar{u}_{i}(x; \beta)$ is uniquely maximized at $Z_{i}.$ As shown in the previous part, $\EE[\epsilon]{U_{i}(x; \beta, s)}$ converges uniformly in $x$ as $s \rightarrow \infty.$ So, we can apply Lemma \ref{lemm:uniform_convergence_maximizer} to conclude that \eqref{eq:maximizer_converge}. This implies \eqref{eq:up}. An identical argument implies \eqref{eq:down}.

Next, we can demonstrate that $\omega_{i}(s; \beta)$ is maximized at some $s^{*}$ and is increasing for $s < s^{*}$ and decreasing for $s > s^{*}.$ We use the following lemma.

\begin{lemm}
\label{lemm:br_fixed_point}
Under Assumption \ref{assumption:strong_convexity}, if $\sigma^{2} > \frac{1}{\alpha_{i}\sqrt{2\pi e}}$ and $x_{i}^{*}(\beta, s) \in \text{Int}(\mathcal{X})$, $\omega_{i}(s; \beta)$ has a unique fixed point.
\hyperref[subsec:br_fixed_point]{Proof in Appendix \ref{subsec:br_fixed_point}}.
\end{lemm}

By Lemma \ref{lemm:br_fixed_point}, $\omega_{i}(s; \beta)$ has a unique fixed point. Let the unique fixed point be $s^{*}.$ We show that $\nabla_{s} \omega_{i}(s^{*}; \beta) = 0$ by an application of Lemma \ref{lemm:score_gradient}. Let $\bfH = \nabla^{2} c(x_{i}^{*}(\beta, s^{*}) - Z_{i}).$
\begin{align*}
    \nabla_{s} \omega_{i}(\beta, s^{*}) &= \beta^{T}\nabla_{s} x_{i}^{*}(\beta, s^{*}) \\
    &=  \frac{\phi_{\sigma}'(s^{*} - \omega_{i}(s^{*}; \beta)) \beta^{T} \bfH^{-1} \beta }{1 + \phi_{\sigma}'(s^{*} - \omega_{i}(s^{*}; \beta)) \beta^{T} \bfH^{-1} \beta}  \\
    &=\frac{\phi_{\sigma}'(0)\beta^{T}\bfH^{-1}\beta)}{1 + (\phi_{\sigma}'(0))\beta^{T}\bfH^{-1}\beta} \\
    &= 0.
\end{align*}
The last line follows from the fact that $\phi_{\sigma}'(0) = 0$. Thus, we have that $\nabla_{s}\omega_{i}(s^{*}; \beta) = 0.$

When $s < s^{*},$ we have that $h_{i}(s; \beta) < h_{i}(s^{*}; \beta)$ because by Lemma \ref{lemm:score_gradient}, $h_{i}(s; \beta)$ is strictly increasing in $s.$ Since $h_{i}(s^{*}; \beta) = 0,$ this implies that $h_{i}(s; \beta) < 0.$ Thus, we have that 
\begin{align*}
    \nabla_{s}\omega_{i}(s; \beta) &= \beta^{T}\nabla_{s}x_{i}^{*}(s) \\
    &= \frac{\phi_{\sigma}'(h_{i}(s))\beta^{T}\bfH^{-1}\beta}{1 + \phi_{\sigma}'(h_{i}(s))\beta^{T}\bfH^{-1}\beta} \\
    &> 0.
\end{align*}
The last line follows because $\phi_{\sigma}'(h_{i}(s)) > 0$ when $h_{i}(s) < 0.$ Thus, $\omega_{i}(s; \beta)$ is increasing when $s < s^{*}.$

When $s>s^{*}$, we have that $h_{i}(s^{*}; \beta) < h_{i}(s; \beta),$ again because $h_{i}$ is strictly increasing. This implies that $h_{i}(s; \beta) > 0$. So, when $s >s^{*},$ $\phi_{\sigma}'(h_{i}(s)) < 0$. Meanwhile,
\begin{align*}
    &1 + \phi_{\sigma}'(h_{i}(s))\beta^{T}\bfH^{-1}\beta \\
    &\geq 1 + \inf_{y \in \mathbb{R}} \phi_{\sigma}'(y) \cdot \sup_{\beta' \in \mathcal{B}} \beta'^{T}\bfH \beta' \\
    &\geq 1 + (- \frac{1}{\sigma^{2}\sqrt{2\pi e}}) \cdot \frac{1}{\alpha_{i}} \\
    &\geq 1 + (-\alpha_{i}) \cdot \frac{1}{\alpha_{i}} \\
    &> 0.
\end{align*}
This means that for $s> s^{*},$ we have that $\nabla_{s} \omega_{i}(s; \beta) < 0.$ So, $\omega_{i}(s; \beta)$ is decreasing when $s > s^{*}.$ 

Since $\omega_{i}(s; \beta)$ is increasing when $s < s^{*}$ and is decreasing when $s > s^{*},$ then $\omega_{i}(s; \beta)$ is maximized when $s = s^{*}.$
\end{subsection}

\begin{subsection}{Proof of Lemma \ref{lemm:cdf}}
\label{subsec:cdf}
Recall that $P(\beta, s)$ denotes the distribution over scores, and the score for an agent $i$ is denoted by \[\beta^{T} X_{i}(\beta, s) = \beta^{T}x_{i}^{*}(\beta, s) + \beta^{T}\epsilon_{i},\] where the randomness in the score comes from the unobservables $(Z_{i}, c_{i}, \epsilon_{i}) \sim F$. Note that $\beta^{T}\epsilon_{i} \sim \Phi_{\sigma}$ because $\epsilon_{i} \sim N(0, \sigma^{2}\mathbf{I}_{d}).$ We have that 
\begin{align*}
    P(\beta, s)(r) &= P(\beta^{T} x_{i}^{*}(\beta, s) + \beta^{T}\epsilon_{i} \leq r) \\
    &=  \int P(\beta^{T}\epsilon_{i} \leq r - \beta^{T} x_{i}^{*}(\beta, s)) dF_{Z, c} &  \\
    &= \int \Phi_{\sigma}(r - \beta^{T} x_{i}^{*}(\beta,s)) dF_{Z, c}.
\end{align*}
Thus, $P(\beta, s)(r)$ has the form given in \eqref{eq:cdf}. Under our conditions, the best response for each agent type exists and is unique via Proposition \ref{prop:br_all}, so $P(\beta, s)(r)$ is a well-defined function.

First, we establish that $P(\beta, s)$ is strictly increasing because we know that $\Phi_{\sigma}$ is strictly increasing, and the sum of strictly increasing functions is also strictly increasing. 

Second, we establish that $P(\beta, s)$ is continuously differentiable in $r$ because $\Phi_{\sigma}$ is continuously differentiable. $P(\beta, s)$ is continuously differentiable in $\beta, s$ because $\Phi_{\sigma}$ is continuously differentiable and the best response mappings $x_{i}^{*}(\beta, s)$ are continuously differentiable (Proposition \ref{prop:br_all}).

The combination of the above two properties is sufficient for showing that $P(\beta, s)$ has a continuous inverse distribution function.

\end{subsection}

\begin{subsection}{Proof of Lemma \ref{lemm:dq_ds}}
\label{subsec:dq_ds}
First, we compute $\frac{\partial q(P(\beta, s))}{\partial s}$  via implicit differentiation. We note that our expression for $\frac{\partial q(P(\beta, s))}{\partial s}$ consists of a convex combination of terms of the form $\beta^{T}\nabla_{s}x_{i}^{*}(\beta, s)$. Finally, we can bound each term in the convex combination and bound $\frac{\partial q(P(\beta, s))}{\partial s}$.

From Lemma \ref{lemm:cdf}, we have that $P(\beta, s)$ has an inverse distribution function, so $q(P(\beta, s))$ is uniquely defined. Thus, $P(\beta, s)( q(P(\beta, s))) = q$ implicitly defines $q(P(\beta, s))$. Using the expression for $P(\beta, s)(r)$ from \eqref{eq:cdf}, we have
\begin{equation}
\label{eq:quantile_equality}
    \int \Phi_{\sigma}(q(P(\beta, s)) - \beta^{T} x_{i}^{*}(\beta, s))dF = q \\
\end{equation}
From Theorem \ref{theo:equilibrium}, $q(P(\beta, s))$ is differentiable in $s$. So, we can differentiate both sides of \eqref{eq:quantile_equality} with respect to $s$.
{
\begin{align*}
    &\frac{\partial}{\partial s} \int \Phi_{\sigma}(q(P(\beta, s)) - \beta^{T} x_{i}^{*}(\beta, s)dF_{Z, c}\\
    &= \int \frac{\partial}{\partial s} \Big( \Phi_{\sigma}(q(P(\beta, s)) - \beta^{T} x_{i}^{*}(\beta, s) \Big) dF_{Z, c} \\
    &= \int \phi_{\sigma}(q(P(\beta, s)) - \beta^{T} x_{i}^{*}(\beta, s)) \cdot \Big(\frac{\partial q(P(\beta, s))}{\partial s} - \beta^{T} \nabla_{s} x_{i}^{*}(\beta, s) \Big) dF_{Z, c}\\
    &= 0.
\end{align*}
}%
Rearranging the last two lines to solve for $\frac{\partial q(P(\beta, s))}{\partial s}$ yields
{
\begin{align*}
    &\frac{\partial q(P(\beta, s))}{\partial s} \\
    &=  \int \beta^{T} \nabla_{s} x_{i}^{*}(\beta, s) \cdot \frac{\phi_{\sigma}(q(P(\beta, s)) - \beta^{T} x_{i}^{*}(\beta, s))}{\int \phi_{\sigma}(q(P(\beta, s)) - \beta^{T} x_{i}^{*}(\beta, s)) dF_{Z, c}} dF_{Z, c}. 
\end{align*}
}%
We can define 
\[ \lambda_{i}(\beta, s) = \frac{\phi_{\sigma}(q(P(\beta, s)) - \beta^{T} x_{i}^{*}(\beta, s))}{\int \phi_{\sigma}(q(P(\beta, s)) - \beta^{T} x_{i}^{*}(\beta, s)) dF_{Z, c}}.\]
where $0 \leq \lambda_{i}(\beta, s) \leq 1$ and $\int \lambda_{i}(\beta, s) dF_{Z, c} = 1.$ As a result, $\frac{\partial q(P(\beta, s))}{\partial s}$ is a convex combination of $\beta^{T}\nabla_{s}x_{i}^{*}(\beta, s)$ terms:

\[ \frac{\partial q(P(\beta, s))}{\partial s} = \int \beta^{T} \nabla_{s} x_{i}^{*}(\beta, s) \cdot \lambda_{i}(\beta, s) dF_{Z, c}.  \]

We can upper bound each term $\beta^{T} \nabla_{s} x_{i}^{*}(\beta, s)$. When $\sigma^{2} > \frac{1}{\alpha_{*}\sqrt{2\pi e}}$ Lemma \ref{lemm:score_gradient} gives us that for any agent type $(Z_{i}, c_{i}) \in \text{supp}(F)$, the function $h_{i}(s; \beta) = s - \beta^{T}x_{i}^{*}(\beta, s)$ is strictly increasing. Since $h_{i}(s; \beta)$ is strictly increasing and differentiable, we have that \[\frac{dh_{i}}{ds} = 1 - \beta^{T} \nabla_{s} x_{i}^{*}(\beta, s) > 0.\] As a result, each term satisfies $\beta^{T}x_{i}^{*}(\beta, s) < 1$. Since $\frac{\partial q(P(\beta, s))}{\partial s} $ is a convex combination of such terms, we also have that $\frac{\partial q(P(\beta, s))}{\partial  s} < 1.$

When $\sigma^{2} > \frac{2}{\alpha_{*}\sqrt{2\pi e}} \cdot \frac{\bar{\kappa} + 1}{\bar{\kappa}}$, Propostion \ref{prop:br_all} gives us that for any agent $i$ with unobservables sampled from $F$, the expected score $\omega_{i}^{*}(s; \beta)$ is a contraction in $s$ with Lipschitz constant at most $\kappa$, so $|\beta^{T}\nabla_{s}x_{i}^{*}| < \kappa$. As a result, since $\frac{\partial q(P(\beta, s))}{\partial s} $ is a convex combination of such terms $\beta^{T}\nabla_{s} x_{i}^{*}$, then $|\frac{\partial q(P(\beta, s))}{\partial  s}| < \kappa.$
\end{subsection}

\begin{subsection}{Proof of Lemma \ref{lemm:dist_limits}}
\label{subsec:dist_limits}
The cdf of $P$ must be 
\begin{equation}
\label{eq:P}
    P(r) = \int \Phi_{\sigma}(r - \beta^{T}Z_{i}) dF_{Z}.
\end{equation}
We have that
\begin{align*}
    &\lim_{s \rightarrow \infty} P(\beta, s)(r) \\
    &= \lim_{s \rightarrow \infty} \int \Phi_{\sigma}(r - \beta^{T} x_{i}^{*}(\beta,s)) dF_{Z, c} \\
    &= \int\Phi_{\sigma}(r - \beta^{T}Z_{i}) dF_{Z, c} \\
    &= P(r).
\end{align*}
The first line follows from the definition of $P(\beta, s)$ from \eqref{eq:cdf}. The second line follows from Lemma \ref{lemm:limits_br}. The third line follows from \eqref{eq:P}. An identical proof can be used to show that 
\[  \lim_{s \rightarrow -\infty} P(\beta, s)(r) = P(r).\]
Since $P(\beta, s)(r) \rightarrow P(r)$ pointwise in $r$ as $s \rightarrow \infty$, $P(\beta, s)$ and $P$ are continuous and invertible and have continuous inverses (Lemma \ref{lemm:cdf}), then we also have that 
\[ \lim_{s \rightarrow \infty} q(P(\beta, s)) = q(P).\]
Similarly, we also have that
\[ \lim_{s \rightarrow -\infty} q(P(\beta, s)) = q(P).\]
\end{subsection}

\begin{subsection}{Proof of Lemma \ref{lemm:stoch_dom_little}}
\label{subsec:stoch_dom_little}
First, we verify that $z^{t}$ is a valid random variable. We note that
\begin{align*}
    &P(z^{t} = \epsilon_{g}) + \sum_{k=1}^{\infty} P(z^{t}= C_{k}) \\
    &= p_{n}(\epsilon_{g}) + \sum_{k=1}^{\infty} \frac{(1- p_{n}(\epsilon_{g}))}{2^{k}} \\
    &= p_{n}(\epsilon_{g}) + \frac{1}{1 - \frac{1}{2}} \cdot \frac{1 - p_{n}(\epsilon_{g}))}{2} \\
    &= p_{n}(\epsilon_{g}) + (1 - p_{n}(\epsilon_{g})) \\
    &= 1.
\end{align*}
    
Now, we show that $z^{t}$ stochastically dominates $|q(P^{n}(\beta, s)) - q(P(\beta, s)))|$. So, for $b \in \mathbb{R},$ we show that 
\begin{equation}
\label{eq:stoch_dom}
    P(|q(P^{n}(\beta, s)) - q(P(\beta, s)))| \geq b) \leq P(z^{t} \geq b),\end{equation}
which is equivalent to the condition that $z^{t}$ stochastically dominates $|q(P^{n}(\beta, s)) - q(P(\beta, s))|.$

From Lemma \ref{lemm:constant_in_quantile_bound} we realize that for $b \in \mathbb{R},$
\[
    P(|q(P^{n}(\beta, s)) - q(P(\beta, s)))| \geq b) \leq 1 - p_{n}(b).
\]
In addition, we have that
{\small
\[
    P(z^{t} \geq b) = \begin{cases}
        1 & \text{ if } b \leq \epsilon_{g}  \\
        1 - p_{n}(\epsilon_{g}) & \text{ if } \epsilon_{g} < b \leq C_{1} \\
        \frac{1- p_{n}(\epsilon_{g})}{2^{k-1}} & \text{ if } C_{k-1} < b \leq C_{k}, k\geq 2.\\ 
    \end{cases}
\]
}
We show that \eqref{eq:stoch_dom} holds for the three cases 1) $b\leq \epsilon_{g}$, 2) $\epsilon_{g} < b \leq C_{1}$, and 3) $C_{k-1} < b \leq C_{k}$ for $k \geq 2$. When $b \leq \epsilon_{g}$, we have that
\begin{align*} 
&P(|q(P^{n}(\beta, s)) - q(P(\beta, s)))| \geq b) \\
&\leq 1 - p_{n}(b) \\
&\leq 1 \\
&= P(z^{t} \geq b).\end{align*}
When $ \epsilon_{g} < b \leq S_{1}$, we have that $p_{n}(b) \geq p_{n}(\epsilon_{g})$ because $p_{n}(y)$ is increasing in $y$. So, we note that $1 - p_{n}(b) \leq 1 - p_{n}(\epsilon_{g}).$ This yields
\begin{align*} 
&P(|q(P^{n}(\beta, s)) - q(P(\beta, s)))| \geq b) \\
&\leq 1 - p_{n}(b) \\
&\leq 1  - p_{n}(\epsilon_{g}) \\
&= P(z^{t} \geq b).\end{align*}

To prove the result in the case where $C_{k-1} < b \leq C_{k}, k \geq 2$, we first show that the definition of $C_{k}$ in \eqref{eq:s_k} implies that 
\begin{equation}
\label{eq:bound}
    1 - p_{n}(C_{k-1}) = \frac{1 - p_{n}(\epsilon_{g})}{2^{k-1}},
\end{equation} as follows. First, we consider the definition of $C_{k-1}$ below
\[ C_{k-1} = \sqrt{ \frac{1}{2nM_{\epsilon_{g}}^{2}} \cdot \log\Big(\frac{2^{k}}{1-p_{n}(\epsilon_{g})}\Big)},\]
and we square both sides:
\[ C_{k-1}^{2} = \frac{1}{2nM_{\epsilon_{g}}^{2}} \cdot \log\Big(\frac{2^{k}}{1 - p_{n}(\epsilon_{g})}\Big).\]
Multiplying by $-2nM_{\epsilon_{g}}^{2}$ and exponentiating both sides yields
\begin{equation}
\label{eq:interm}
\exp(-2nM_{\epsilon_{g}}^{2}C_{k-1}^{2}) = \frac{1 - p_{n}(\epsilon_{g})}{2^{k}}. \end{equation}
Finally, we realize that \eqref{eq:bound} holds because
\begin{align*}
    1 - p_{n}(C_{k-1}) &= 2\exp(-2nM_{\epsilon_{g}}^{2}C_{k-1}^{2}) \\
    &= 2 \cdot  \frac{1 - p_{n}(\epsilon_{g})}{2^{k}} \\
    & = \frac{1 - p_{n}(\epsilon_{g})}{2^{k-1}},
\end{align*}
and the second line follows by \eqref{eq:interm}. Using \eqref{eq:bound}, we observe that
\begin{align*}
    &P(|q(P^{n}(\beta, s)) - q(P(\beta, s)))| \geq b) \\
    &= 1 - p_{n}(b) \leq  1 - p_{n}(C_{k-1}) \\
    &= \frac{1 - p_{n}(\epsilon_{g})}{2^{k-1}} \\
    &= P(z^{t} \geq C_{k}) \\
    &= P(z^{t} \geq b).
\end{align*}
Thus, we conclude that \eqref{eq:stoch_dom} holds, yielding the desired result.
\end{subsection}

\begin{subsection}{Proof of Lemma \ref{lemm:stoch_dom_big}}
\label{subsec:stoch_dom_big}
We observe that
{\small
\begin{align}
    &|\hat{S}^{t}_{n} - s(\beta) | \\
    &= | \hat{S}^{t}_{n} - q(P(\beta, \hat{S}^{t-1}_{n})) + q(P(\beta, \hat{S}^{t-1}_{n})) - s(\beta) | \label{eq:cont_1} \\
    &=| q(P^{n}(\beta, \hat{S}^{t-1}_{n})) - q(P(\beta, \hat{S}^{t-1}_{n})) + q(P(\beta, \hat{S}^{t-1}_{n})) - s(\beta) | \label{eq:cont_2} \\
    &\leq |q(P^{n}(\beta, \hat{S}^{t-1}_{n})) - q(P(\beta, \hat{S}^{t-1}_{n}))  | \\
    &\indent+  |q(P(\beta, \hat{S}^{t-1}_{n})) - s(\beta) | \label{eq:cont_3} \\
    &\leq  |q(P^{n}(\beta, \hat{S}^{t-1}_{n})) - q(P(\beta, \hat{S}^{t-1}_{n}))  | \\
    &\indent+ \kappa |\hat{S}^{t-1}_{n} - s(\beta)| \label{eq:cont_4} \\
    &\preceq_{\text{SD}} z^{t} + \kappa|\hat{S}^{t-1}_{n} - s(\beta)|. \label{eq:cont_5}
\end{align}
}%
We have \eqref{eq:cont_2} because $\hat{S}^{t}_{n}$ is generated via \eqref{eq:emp_fpi}. \eqref{eq:cont_4} holds because $q(P(\beta, s))$ is a contraction mapping in $s$ and $s(\beta)$ is the unique fixed point of $q(P(\beta, s))$. \eqref{eq:cont_5} follows from Lemma \ref{lemm:stoch_dom_little}.

Using recursion, we find that 
\begin{align*}
|\hat{S}^{t}_{n} - s(\beta)| &\preceq_{\text{SD}} \sum_{i=1}^{t} z^{t-i} \kappa^{i} \\
&\indent+ \kappa^{t} |\hat{S}^{0}_{n} - s(\beta)| \\
&= \sum_{i=1}^{t} z^{t-i} \kappa^{i} + \kappa^{t} \cdot C. \end{align*}

\end{subsection}

\begin{subsection}{Proof of Lemma \ref{lemm:truncation}}
\label{subsec:truncation}
Let $\{z^{t}\}_{t \geq 1}$ be a sequence of random variables where 
\[ z^{t} = \begin{cases} \epsilon_{g} & \text{w.p. } p_{n}(\epsilon_{g}) \\
C_{k} &\text{w.p. } \frac{1 - p_{n}(\epsilon_{g})}{2^{k}}.\\
\end{cases},\] 
where $\epsilon_{g} = \frac{\epsilon (1 - \kappa)}{2}$ and $p_{n}(\epsilon_{g})$ is the bound from Lemma \ref{lemm:constant_in_quantile_bound}.

Since $s(\beta) \in \mathcal{S},$ we have that
\begin{align}
    &|\hat{S}^{t}_{n} - s(\beta)| \\
    &\leq |q(P^{n}(\beta, \hat{S}^{t-1}_{n})) - s(\beta)| \label{eq:trunc_1} \\
    &\leq |q(P^{n}(\beta, \hat{S}^{t-1}_{n})) - q(P(\beta,  \hat{S}^{t-1}_{n})) | \\
    &\indent+ |q(P_{\beta,  \hat{S}^{t-1}_{n}}) - s(\beta)| \label{eq:trunc_2}\\
    &\preceq_{\text{SD}} z^{t} + \kappa|\hat{S}^{t-1}_{n} - s(\beta)| \label{eq:trunc_3}.
\end{align}
where \eqref{eq:trunc_1} holds because the truncation is contraction map to the equilibrium threshold $s(\beta)$ and \eqref{eq:trunc_3} holds because $q(P(\beta, s))$ is a contraction in $s$ and $s(\beta)$ is equilibrium threshold. So, as in Lemma \ref{lemm:stoch_dom_big}, we can show that 
\[ |\hat{S}^{t}_{n} - s(\beta)| \preceq_{\text{SD}} \sum_{i=1}^{k} z^{t-i}\kappa^{i} + \kappa^{k}|\hat{S}^{t-k}_{n} - s(\beta)|.\]
Let $C = |\hat{S}_{0}^{n} - s(\beta)|.$ Thus, an identical argument as the proof of Theorem \ref{theo:emp_fpi_convergence} can be used to show that if \[t \geq \Big\lceil \frac{\log(\frac{\epsilon}{2C}) }{\log \kappa} \Big\rceil, \quad n \geq \frac{1}{2M_{\epsilon_{g}}^{2}} \log (\frac{4(t + 1)}{\delta}),\] we have that
\[ P(|\hat{S}_{n}^{t} - s(\beta)| \geq \epsilon ) \leq \delta.\]
Using an identical argument as the proof of Corollary \ref{coro:convergence_in_prob_threshold}, it can be shown that for any sequence $\{t_{n}\}$ such that $t_{n} \uparrow \infty$ as $n\rightarrow \infty$ and $t_{n} \prec \exp(n),$ $\hat{S}^{t_{n}}_{n} \xrightarrow{p} s(\beta)$, as desired.
\end{subsection}

\begin{subsection}{Proof of Lemma \ref{lemm:nice_loss_and_pi}}
\label{subsec:nice_loss_and_pi}
 We can use the following abbreviations
\begin{align*}
    u&:= (Z_{i}, c_{i}, \epsilon_{i}) \\
    \theta&:= (\beta, s, r) \\
    \tilde{j}(u, \theta) &:= \tilde{j}(Z_{i}, c_{i}, \epsilon_{i}, \beta, s, r) \\
    \tilde{y}(u, \theta) &:= \tilde{y}(Z_{i}, c_{i}, \epsilon_{i}, \beta, s, r) \\
    \tilde{k}(u, \theta) &:= \tilde{k}(Z_{i}, c_{i}, \epsilon_{i}, \beta, s, r).
\end{align*}
The conditions on $a(u; \theta)$ in Lemma \ref{lemm:uniform_convergence_sum} include that 
\begin{enumerate}
    \item $\theta \in \Theta$, where $\Theta$ is compact.
    \item $a(u; \theta)$ is continuous with probability 1 for each $\theta \in \Theta.$
    \item $|a(u; \theta)| \leq d(u)$ and $\EE{d(u)} < \infty.$ 
\end{enumerate}

First, for all of $\tilde{y}, \tilde{j}, \tilde{k}$, we have that the parameter space $\mathcal{B} \times \mathcal{S} \times \mathcal{S}$ is compact.

Second, we can verify that for fixed parameters, $\tilde{y}, \tilde{j},$ and $\tilde{k}$ are continuous with probability 1. By Assumption \ref{assumption:nice_loss}, we have that for each $w \in \{0, 1\}$, $m(Z_{i}, c_{i}; w) + \delta_{i}$ is continuous w.r.t. to the data, so the only discontinuity of $\tilde{y}$ occurs at the threshold when $w$ flips from 0 to 1. Similarly, $\tilde{j}$ is an indicator function, so its only discontinuity occurs at the threshold. Thus, $\tilde{y}(\cdot, \bftheta)$ and $\tilde{j}(\cdot, \bftheta)$ are discontinuous on the set \[A=\{(Z_{i}, c_{i}, \epsilon_{i}): \beta^{T}X_{i}(\beta, s) = r\}\]
but are otherwise continuous. The probability that $(Z_{i}, c_{i}, \epsilon_{i}) \sim F_{Z, c, \epsilon}$ satisfies the condition of set $A$ is equal to the probability that a score $z \sim P(\beta, s)$ takes value exactly $r$. We note that a singleton subset $\{r\}$ will have measure 0, so the probability that a score takes value $r$ is 0. Thus, $A$ must also have measure 0. Since $\tilde{y}$ and $\tilde{j}$ are continuous except on a set of measure 0, $\tilde{y}$ and $\tilde{j}$ are continuous with probability 1.
We realize that $\tilde{k}$ is continuous except for on the following set $A'= \{(Z_{i}, c_{i}, \epsilon_{i}): \beta^{T}X_{i}(\beta, s) = r + \frac{h}{2} \} \cup \{(Z_{i}, c_{i}, \epsilon_{i}): \beta^{T}X_{i}(\beta, s) = r - \frac{h}{2} \}.$
The probability that $(Z_{i}, c_{i}, \epsilon_{i}) \sim F$ satisfies the condition of set $A'$ is equal to the probability that a score $z \sim P(\beta, s)$ takes value exactly $r + \frac{h}{2}$ or value $r - \frac{h}{2}.$ Since the sets $\{r + \frac{h}{2}\}$ and $\{r - \frac{h}{2}\}$ have measure zero and a countable union of measure zero sets has measure zero, then $A'$ has measure zero. Thus, $\tilde{k}$ is continuous with probability 1.

Third, we note that $\tilde{y}, \tilde{j},$ and $\tilde{k}$ are dominated. For $\tilde{y},$ Assumption \ref{assumption:nice_loss} gives us that $\tilde{y}$ is bounded, so any constant function $d(u) = a$ for $a \geq \sup_{(z, c) \in \text{supp}(F), w \in \{0, 1\}} m(z, c; w)$ dominates $\tilde{y}$. Since $\tilde{j}$ and $\tilde{k}$ are indicators, they only takes values $\{0, 1\}$, so any constant function $d(u) = a$ where $a> 1$ satisfies the required condition. Thus, $\tilde{y}$, $\tilde{j}$, and $\tilde{k}$ satisfy the conditions of Lemma \ref{lemm:uniform_convergence_sum}.
\end{subsection}

\begin{subsection}{Proof of Lemma \ref{lemm:find_equivalent_agent_main}}
\label{subsec:find_equivalent_agent_main}
In this proof, we first verify that the agent with unobservables $(Z_{i'; b, \zeta, \xi}, c_{i'})$ has $Z_{i'; b, \zeta, \xi} \in \mathcal{X}$ and has a strongly-convex cost function $c_{i'}$. Second, we verify the value of the best response for this agent and show that it lies in $\text{Int}(\mathcal{X})$. Lastly, we show that this agent's raw score (without noise) matches that of the agent with unobservables $(Z_{i}, c_{i})$ under the perturbed model.

By the following lemma, we realize that with $Z_{i'; b, \zeta, \xi}$ as defined in \eqref{eq:eta_prime}, $Z_{i'; b, \zeta, \xi} \in \mathcal{X}$ as long as the perturbation magnitude $b$ is sufficiently small. 
\begin{lemm}
\label{lemm:find_equivalent_agent}
Let $\zeta \in \{-1, 1\}^{d}$ and $\xi \in \{-1, 1\}$ denote perturbations. Let $b$ be the magnitude of the perturbations. Let $\beta \in \mathcal{B}$. If $y \in \text{Int}(\mathcal{X})$ and $x \in \mathcal{X}$, then for any $b$ sufficiently small and
\[ y' = y +  \beta \cdot b \cdot ( \zeta^{T} x - \xi),\]
we have that $y' \in Int(\mathcal{X}).$
\hyperref[subsec:find_equivalent_agent]{Proof in Appendix \ref{subsec:find_equivalent_agent}}.
\end{lemm}
With the above lemma, we define $b_{1}$ so that $Z_{i'; b, \zeta, \xi} \in \mathcal{X}$ for $b < b_{1}.$

We verify that $c_{i'}$ satisfies Assumption \ref{assumption:strong_convexity}. We note that $c_{i'}$ is twice continuously differentiable because it is the sum of twice continuously differentiable functions. Second, we show that $c_{i'}$ is strongly convex. Since $c_{i}$ satisfies Assumption \ref{assumption:strong_convexity}, then $c_{i}$ is $\alpha_{i}-$strongly convex for $\alpha_{i}> 0$ and twice continuously differentiable. In addition, $\phi_{\sigma}(s-r)\beta^{T}y$ is differentiable and convex in $y$. By the strong convexity of $c_{i}$ and the convexity of $\phi_{\sigma}(s-r)\beta^{T}y$ in $y$, we have that $c_{i'}(y)$ is $\alpha_{i}$-strongly convex, satisfying Assumption \ref{assumption:strong_convexity}. 

Let \[ x_{2} = x_{1} + b \cdot \beta(\zeta^{T}x_{1} - \xi).\] Note that by Lemma \ref{lemm:find_equivalent_agent}, for sufficiently small $b$, we have that $x_{2} \in \text{Int}(\mathcal{X}).$ Suppose that $x_{2} \in \text{Int}(\mathcal{X})$ for $b < b_{2}.$ We will show two useful facts about $x_{2}$ that will enable us to show that the best response of the agent with unobservables $(Z_{i'; b, \zeta, \xi}, c_{i'})$ to the model $\beta$ and threshold $s$ is given by $x_{2}$. For the first fact, we see that $x_{2} - Z_{i'; b, \zeta, \xi} = x_{1} - Z_{i}$.
\begin{align*}
    &x_{2} - Z_{i'; b, \zeta, \xi} \\
    &= x_{1} + \beta(b\zeta^{T}x_{1} - b\xi) - Z'_{i'; b, \zeta, \xi} \\
    &= x_{1} + \beta(b\zeta^{T}x_{1} - b\xi)-  Z_{i} - \beta \cdot b \cdot (\zeta^{T}x_{1} - \xi) \\
    &= x_{1} - Z_{i}.
\end{align*}
For the second fact, we have that $\beta^{T}x_{2} = r.$
\begin{align}
\beta^{T} x_{2} &= \beta^{T} \Big( \beta(b\zeta^{T}x_{1} - b\xi) + x_{1} \Big) \\
&=(\beta^{T}\beta) \cdot (b\zeta^{T}x_{1} - b\xi) + \beta^{T}x_{1} \\
&= (\beta + b\zeta)^{T}x_{1} - b\xi \\
&= r. \label{eq:score_unperturbed}
\end{align}
Now, we show that $x_{2} = x_{i}^{*}(\beta + b\zeta, s + b\xi)$. Since Assumption \ref{assumption:strong_convexity} holds, by Lemma \ref{lemm:diff_concavity_utility} it is sufficient to check $\nabla_{x} \EE[\epsilon]{U_{i'}(x_{2}; \beta, s)]} = 0$ to verify that $x_{2}$ is the best response:
{\small
\begin{align}
    &\nabla_{x} \EE[\epsilon]{U_{i'}(x_{2}; \beta, s)]} \\
    &= -\nabla c_{i'}(x_{2} - Z_{i'; b, \zeta, \xi}) + \phi_{\sigma}(s - \beta^{T}x_{2})\beta^{T} \\
    &= -\nabla c_{i}(x_{1} - Z_{i}) + \phi_{\sigma}(s - r)\beta^{T} \\
    &= - \nabla c_{i}(x_{1} - Z_{i}) + \phi_{\sigma}(s- r)b\zeta^{T} + \phi_{\sigma}(s - r)\beta^{T}. \label{eq:unperturbed_agent_intermediate}
\end{align}
}%

To further simplify the above equation, we have that $x_{1} = x_{i}^{*}(\beta + b\zeta, s + b\xi)$ and $x_{1} \in \text{Int}(\mathcal{X})$. By Lemma \ref{lemm:diff_concavity_utility}, this implies that $\nabla_{x} \EE[\epsilon]{U_{i}(x_{1}; \beta + b\zeta, s + b\xi)} = 0.$ This gives that
{\small
\begin{align*}
    &\nabla_{x} \EE[\epsilon]{U_{i}(x_{1}; \beta + b\zeta, s + b\xi)} \\
    &= -\nabla c_{i}(x_{1} - Z_{i}) + \phi_{\sigma}(s + b\xi - (\beta + b\zeta)^{T}x_{1})(\beta + b\zeta)^{T}  \\
    &= -\nabla c_{i}(x_{1} - Z_{i}) + \phi_{\sigma}(s - r)(\beta + b\zeta)^{T} \\
    &= 0.
\end{align*}
}
So, we have that 
\[\nabla c_{i}(x_{1} - Z_{i}) = \phi_{\sigma}(s - r)(\beta + b\zeta)^{T}.\]
Substituting this result into \eqref{eq:unperturbed_agent_intermediate} yields
{\small
\begin{align*}
     &\nabla_{x} \EE[\epsilon]{U_{i}(x_{2}; \beta, s)} \\
     &= - \nabla_{x} c_{i}(x_{1} - Z_{i}) + \phi_{\sigma}(s- r)b\zeta^{T} + \phi_{\sigma}(s - r)\beta^{T} \\
     &= - \phi_{\sigma}(s - r)(\beta + b\zeta)^{T}  + \phi_{\sigma}(s- r)b\zeta^{T} + \phi_{\sigma}(s - r)\beta^{T} \\
     &= 0.
\end{align*}
}
We note that if $b < \min(b_{1}, b_{2})$, then we have that $Z_{i';b, \zeta, \xi}, x_{2} \in \text{Int}(\mathcal{X})$. Under such conditions, we conclude that $x_{2} = x_{i}^{*}(\beta, s)$. The score obtained by the agent with type $(Z_{i'; b, \zeta, \xi}, c_{i'})$ under the model $\beta$ and threshold $s$ is $\beta^{T}x_{2}.$ As we showed earlier in \eqref{eq:score_unperturbed}, this quantity is equal to $r$. Thus, for sufficiently small perturbations, the agent with unobservables $(Z_{i}, c_{i})$  under perturbations obtains the same raw score as the agent with unobservables $(Z_{i';b, \zeta, \xi}, c_{i'})$ in the unperturbed setting.
\end{subsection}

\begin{subsection}{Proof of Lemma \ref{lemm:transformed_distribution}}
\label{subsec:transformed_distribution}
Note that by Assumption \ref{assumption:finite_types}, $F_{Z, c}$ has finite support. We denote elements of the support as $\nu:=(Z, c)$. Let $f$ be the probability mass function of $F_{Z, c}$, so $f(\nu)$ gives the probability of that an agent has unobservables $\nu$.

We construct the probability mass function $\tilde{f}_{b}$ of the distribution $\tilde{F}.$ Let $\nu_{i} \sim F_{Z, c}.$ Under Assumptions \ref{assumption:br_interior} and \ref{assumption:etas_in_interior},  we use the transformation $T$ to map an agent with unobservables $\nu_{i} = (Z_{i}, c_{i})$ to an agent with unobservables $\nu_{i';b, \zeta, \xi} = (Z_{i'; b, \zeta, \xi}, c_{i})$, as defined in Lemma \ref{lemm:find_equivalent_agent_main}, for each perturbation $\zeta \sim \{-1, 1\}^{d}$ and $\xi \sim \{-1, 1\}$ and unobservable  $\nu \sim F_{Z, c}$. 

We assign $\tilde{f}(\nu_{i';b, \zeta, \xi}') = \frac{1}{2^{d+1}}f(\nu).$ Since there are finitely many unobservables that occur with positive probability in $F_{Z, c}$ and finitely many perturbations ($2^{d+1}$ possible perturbations), there exists $b> 0$ such that this transformation is possible simultaneously for all types in the support of $F$ and all perturbations. Note that this is a valid probability mass function because $\sum_{\nu \sim \text{supp}(F)} f(\nu) = 1$, so
{\small
\begin{align*}
 &\sum_{\nu_{i'; b, \zeta, \xi} \in \text{supp}(\tilde{F})} \tilde{f}(\nu_{i'; b, \zeta, \xi}) \\
 &=  \sum_{\nu \sim \text{supp}(F)} \sum_{\substack{\zeta \in \{-1, 1\}^{d} \\ \xi \in \{-1, 1\}}} \frac{1}{2^{d+1}} f(\nu) \\
 &= 1.
 \end{align*}}%
In addition, note that the transformation yields
{\small
\[ \beta^{T}x_{i'; b, \zeta, \xi}^{*}(\beta, s) = (\beta + b\zeta)^{T}x_{i}^{*}(\beta + b\zeta, s + b\xi) - b\xi.\]}%

The transformation given in Lemma \ref{lemm:find_equivalent_agent_main} also provides other desirable properties. For instance, $Z_{i';b, \zeta, \xi}  \in \mathcal{X}$, which means that the support of $\tilde{F}_{Z}$ is contained in $\mathcal{X}$. The cost function of the $c_{i'}$ satisfies Assumption \ref{assumption:strong_convexity} with $\alpha_{i'} = \alpha_{i}$, which means that $\alpha_{*}(\tilde{F}) = \alpha_{*}(F)$. In addition, under Assumption \ref{assumption:etas_in_interior}, $c_{i}' \in \mathcal{C}.$ Lastly, for $\nu_{i'; b, \zeta, \xi} \sim \tilde{F}$, the best responses of the agents lie in $\text{Int}(\mathcal{X})$.

Additionally, we have that
{
\begin{align*} 
&P(\beta, s, b)(r) \\
&= \frac{1}{2^{d+1}} \sum_{\substack{\zeta \in \{-1, 1\}^{d} \\\xi \in \{-1, 1\} }} \int \Phi_{\sigma}\Big(r - (\beta + b\zeta)^{T}x_{i}^{*}(\beta + b\zeta ,s + b\xi) - b\xi\Big) dF_{Z, c} \\
&= \sum_{\nu_{b, \zeta, \xi}' \in \text{supp}(\tilde{F})} \Phi_{\sigma}(r - \beta^{T}x_{i'; b, \zeta, \xi}^{*}(\beta, s)) \tilde{f}_{b}(\nu_{i'; b, \zeta, \xi}') \\
&= \int \Phi_{\sigma}(r - \beta^{T}x_{i'; b, \zeta, \xi}^{*}(\beta, s)) d\tilde{F}_{Z,c}.\\
\end{align*}
}

The final line matches the form of the score distribution's CDF given in Lemma \ref{lemm:cdf} assuming that the unobservables are distributed according to $\tilde{F}.$
\end{subsection}

\begin{subsection}{Proof of Lemma \ref{lemm:transformed_distribution_continuity_fp}}
\label{subsec:transformed_distribution_continuity_fp}
We define the sequence of functions $\{h_{b}(s)\}$ where $h_{b}: \mathcal{S} \rightarrow \mathcal{S}.$ Let $h_{b}(s) := s - q(P(\beta, s, b))$ and $h(s) := s - q(P(\beta, s)).$

We aim to apply Lemma \ref{lemm:uniform_convergence_root} to this sequence of functions. We realize that the requirements on $h(s)$ are given by our results from Section \ref{sec:results_equilibrium}. Theorem \ref{theo:equilibrium} give us that $h(s)$ has a unique root, which is the unique fixed point of $q(P(\beta, s))$ called $s(\beta).$ Also, we note that $h_{b}(s)$ and $h(s)$ are defined on the compact set $\mathcal{S}.$ It remains to check that
\begin{enumerate}
    \item Each $h_{b}(s)$ is continuous,
    \item Each $h_{b}(s)$ has a unique root, which is the fixed point of $q(P(\beta, s, b))$ called $s(\beta, b)$,
    \item As $b \rightarrow 0$, $h_{b}(s) \rightarrow h(s)$ is uniformly.
\end{enumerate}

To verify the first two properties from the above list, we apply the transformation provided in Lemma \ref{lemm:transformed_distribution} to $P(\beta, s, b).$ This transformation enables us to apply the results from Section \ref{sec:results_equilibrium} directly to expressions involving $P(\beta, s, b).$ 

Since the transformation maintains all of our assumptions and \[\sigma^{2} > \frac{1}{ \alpha_{*}(F) \sqrt{2\pi e}} =\frac{1}{\alpha_{*}(\tilde{F}_{b})\sqrt{2\pi e}},\] we can apply Theorem \ref{theo:equilibrium} to see that $q(P(\beta, s, b))$ is continuous in $s$ and to see that $q(P(\beta, s, b))$ has a unique fixed point in $\mathcal{S}.$ We can call the fixed point $s(\beta, b)$, and $s(\beta, b)$ is also the unique root of $h_{b}(s).$

Finally, we must check the third point, which is uniform convergence of $h_{b}(s)$ to $h(s).$ We aim to apply Lemma \ref{lemm:uniform_convergence_monotonic}. First, we note that the continuity of $h(s)$ is given by Theorem \ref{theo:equilibrium}. Second, we check that each $h_{b}(s)$ is monotonically increasing. Under the transformation from Lemma \ref{lemm:transformed_distribution}, we can apply Lemma \ref{lemm:dq_ds} to observe that under our conditions, $\frac{\partial q(P(\beta, s, b))}{\partial s} < 1$, so $h_{b}(s)$ is strictly increasing. Third, we show that $h_{b}(s)$ converges pointwise to $h(s)$ as follows. 

To show $h_{b}(s) \rightarrow h(s)$ pointwise, we show $q(P(\beta, s, b)) \rightarrow q(P(\beta, s))$ pointwise. Note that by Lemma \ref{lemm:cdf}, $P(\beta, s)$ is strictly increasing, so we can let a lower bound on its density be $d$ for $s \in \mathcal{S}$, i.e.
\[ d = \inf_{r\in \mathcal{S}} p(\beta, s)(r). \]
Then, we realize that 
\begin{align*}
&|q(P(\beta, s, b)) - q(P(\beta, s))| \\
&\leq \frac{1}{d} \cdot \sup_{r \in \mathcal{S}}|P(\beta, s, b)(r) - P(\beta, s)(r)|.\end{align*} The following lemma gives us the required uniform convergence in $r$.
\begin{lemm}
\label{lemm:uniform_convergence_perturbed_cdf}
Under Assumptions \ref{assumption:strong_convexity}, 
\ref{assumption:finite_types},
\ref{assumption:br_interior},  and \ref{assumption:etas_in_interior}, if $\sigma^{2} > \frac{1}{\alpha_{*}(F)\sqrt{2\pi e}}$ then $P(\beta, s, b)(r) \rightarrow P(\beta, s)(r)$ uniformly in $r$ as $b \rightarrow 0.$\hyperref[subsec:uniform_convergence_perturbed_cdf]{Proof in Appendix \ref{subsec:uniform_convergence_perturbed_cdf}}.
\end{lemm}

So, we have that $q(P(\beta, s, b)) \rightarrow q(P(\beta, s))$ pointwise in $s$. This implies $h_{b}(s) \rightarrow h(s)$ pointwise. Thus, we have that $h_{b}(s)$ and $h(s)$ satisfy the conditions of Lemma \ref{lemm:uniform_convergence_monotonic}, which implies that $h_{b}(s) \rightarrow h(s)$ uniformly.

Thus, the conditions of Lemma \ref{lemm:uniform_convergence_root} are satisfied, so we have that $s(\beta, b) \rightarrow s(\beta)$ as $b \rightarrow 0.$
\end{subsection}

\begin{subsection}{Proof of Lemma \ref{lemm:quantile_perturbed_stoch_process}}
\label{subsec:quantile_perturbed_stoch_process}
For sufficiently small $b$, we can apply Lemma \ref{lemm:transformed_distribution} to show that $P(\beta, s, b)$ is equal to the score distribution generated when agents with unobservables $\nu_{b, \zeta, \xi}' = (Z_{i'; b, \zeta, \xi}, c_{i'} \sim \tilde{F}_{b}$ best respond to a model $\beta$ and threshold $s$. The conditions assumed when unobservables $\nu = (Z_{i}, c_{i}) \sim F_{Z, c}$ also hold when unobservables are distributed $\nu_{b, \zeta, \xi}' \sim \tilde{F}_{b}$. In particular, $\alpha_{*}(\tilde{F}^{b}) = \alpha_{*}(F),$ so we have that $\sigma^{2} > \frac{2}{ \alpha_{*}(\tilde{F}^{b}_{Z, c})\cdot \sqrt{2\pi e}}$.

As a result, the results from Section \ref{sec:results_equilibrium} and Section \ref{sec:finite_results} can be used to study $q(P(\beta, s, b))$ and the stochastic fixed point iteration process given by \eqref{eq:stochastic_fpi_with_perturbation}. First, we have that $\sigma^{2} > \frac{2}{ \alpha_{*}(\tilde{F}^{b})\cdot \sqrt{2\pi e}}$, so we have that $q(P(\beta, s, b))$ is a contraction in $s$ by Theorem \ref{theo:equilibrium}. Furthermore, the conditions of Lemma \ref{lemm:truncation} are satisfied by the assumed conditions, the results of Lemma \ref{lemm:transformed_distribution}, and the fact that $q(P(\beta, s, b))$ is a contraction. So, we have that
\[ \hat{S}^{t_{n}}_{b, n} \xrightarrow{p} s(\beta, b),\]
where $s(\beta, b)$ is the unique fixed point of $q(P(\beta, s, b)).$
In addition, since $\{t_{n}\}$ is a sequence such that $t_{n} \uparrow \infty$ as $n \rightarrow \infty$ and $t_{n} \prec \exp(n)$, we certainly have that $\{t_{n}-1\}$ is a sequence such that $t_{n}-1 \uparrow \infty$ as $n \rightarrow \infty$ and $t_{n}-1 \prec \exp(n)$, so again by Lemma \ref{lemm:truncation}, we have that \[ \hat{S}^{t_{n}-1}_{b, n} \xrightarrow{p} s(\beta, b).\]
\end{subsection}

\begin{subsection}{Proof of Corollary \ref{coro:dl_ds}}
\label{subsec:dl_ds}
The proof of this result is analogous to Theorem \ref{theo:direct_effect}.
\end{subsection}

\begin{subsection}{Proof of Lemma \ref{lemm:dpi_dbeta}}
\label{subsec:dpi_dbeta}
To simplify notation, we use the following abbreviations. Let $s(\beta)$ be the unique fixed point of $q(P(\beta, s)).$
\begin{align*}
    \tilde{I}_{i}(\beta, s, r) &:= \pi(\beta^{T}X_{i}(\beta, s); \beta, r) \\
    \hat{\Pi}_{n}(\beta, s, r) &:= \frac{1}{n} \sum_{i=1}^{n} \tilde{I}_{i}(\beta, s, r).
\end{align*}
The regression coefficient obtained by running OLS of $I$ on $\mathbf{M}_{\beta}$ is denoted by $\hat{\Gamma}^{b_{n}, n}_{\Pi, \beta}(\beta, \hat{S}^{t_{n}}_{b_{n}, n}, \hat{S}^{t_{n}}_{b_{n}, n})$. The regression coefficient must have the following form.
\begin{equation} 
\hat{\Gamma}^{b_{n}, n}_{\Pi, \beta}(\beta, \hat{S}^{t_{n}}_{b_{n}, n}, \hat{S}^{t_{n}}_{b_{n}, n}) = (\mathbf{S}_{zz}^{n})^{-1}\mathbf{s}_{zy}^{n},\end{equation} 
where $\mathbf{S}_{zz}^{n} := \frac{1}{b_{n}^{2} n}\mathbf{M}_{\beta}^{T}\mathbf{M}_{\beta},$ $\mathbf{s}_{zy}^{n} := \frac{1}{b_{n}^{2} n} \mathbf{M}_{\beta}^{T}I.$
In this proof, we establish convergence in probability of the two terms above separately. The bulk of the proof is the first step, which entails showing that
\[ \mathbf{s}_{zy}^{n} \xrightarrow{p} \frac{\partial \Pi}{\partial \beta}(\beta, s(\beta); s(\beta)).\]
Due to $\tilde{I}'$s dependence on the stochastic process $\{ \hat{S}^{t_{n}}_{b_{n}, n} \}$, the main workhorse of this result is Lemma \ref{lemm:stochastic_equicontinuity_convergence_in_prob}. To apply this lemma, we must establish stochastic equicontinuity for the collection of stochastic processes $\{\tilde{j}_{n}(\beta, s, r)\}$. Second, through a straightforward application of the Weak Law of Large Numbers, we show that
\[ \mathbf{S}_{zz}^{n} \xrightarrow{p} \mathbf{I}_{d}.\] Finally, we use Slutsky's Theorem to establish the convergence the regression coefficient.

We proceed with the first step of establishing convergence of $\mathbf{s}_{zy}.$ We have that
\begin{align*}
    \mathbf{s}_{zy}^{n} &= \frac{1}{b_{n}^{2} n} \mathbf{M}_{\beta}^{T}I\\
    &= \frac{1}{b_{n}^{2} n} \sum_{i=1}^{n} b_{n}\zeta_{i}I_{i} \\
    &= \frac{1}{b_{n}} \cdot \frac{1}{n} \sum_{i=1}^{n} \zeta_{i}I_{i}.
\end{align*}
We fix $j$ and $b_{n}=b$ where $b >0$ and is small enough to satisfy the hypothesis of Lemma \ref{lemm:quantile_perturbed_stoch_process}.
For each $\zeta \in \{-1, 1\}^{d}$ and $\xi \in \{-1, 1\}$, let \begin{align*}
    n_{\zeta, \xi} &= \sum_{i=1}^{n} \mathbb{I}(\zeta_{i}=\zeta, \xi_{i} = \xi).
\end{align*} 
Let $z(\zeta)$ map a perturbation $\zeta \in \{-1, 1\}^{d}$ to the identical vector $\zeta$, except with $j$-th entry set to 0. So, if the $j$-th entry of $\zeta$ is 1, then $\zeta = \mathbf{e}_{j} + z(\zeta).$ If the $j$-th entry of $\zeta$ is -1, then $\zeta = -\mathbf{e}_{j} + z(\zeta).$ So, we have that
\begin{align*}
    I_{i} &= \tilde{I}_{i}( \beta + b \zeta_{i}, \hat{S}^{t_{n}}_{b, n} + b \xi_{i}, \hat{S}^{t_{n}}_{b, n}) \\
    &=\tilde{I}_{i}( \beta +  b\zeta_{i, j} \mathbf{e}_{j} + b\cdot z(\zeta_{i}), \hat{S}^{t_{n}}_{b, n} + b \xi_{i}, \hat{S}^{t_{n}}_{b, n} ).
\end{align*}
As a result, we have that
{
\begin{align}
    &\frac{1}{n}\sum_{i=1}^{n}\zeta_{i, j} I_{i}\\
    =& \frac{1}{n} \sum_{i=1}^{n} \zeta_{i, j} \cdot \tilde{I}_{i}( \beta +  b\zeta_{i, j} \mathbf{e}_{j} + b\cdot z({\zeta}_{i}), \hat{S}^{t_{n}}_{b, n} + b \xi_{i}, \hat{S}^{t_{n}}_{b, n})  \\
    =& \sum_{\substack{\zeta \in \{-1, 1\}^{d} \text{ s.t. } \zeta_{j} = 1 \\ \xi \in \{-1, 1\}}} \frac{n_{\zeta, \xi}}{n} \sum_{i=1}^{n_{\zeta, \xi}} \tilde{I}_{i}(\beta + b\mathbf{e}_{j} + b \cdot z(\zeta),   \hat{S}^{t_{n}}_{b, n} + b \xi, \hat{S}^{t_{n}}_{b, n}) \label{eq:s_zy_part_2a} \\
    &- \sum_{\substack{\zeta \in \{-1, 1\}^{d} \text{ s.t. } \zeta_{j} = -1 \\ \xi \in \{-1, 1\}}} \frac{n_{\zeta, \xi}}{n} \sum_{i=1}^{n_{\zeta, \xi}} \tilde{I}_{i}(\beta - b\mathbf{e}_{j} + b\cdot z(\zeta),   \hat{S}^{t_{n}}_{b, n} + b \xi, \hat{S}^{t_{n}}_{b, n}) \label{eq:s_zy_part_2b}
\end{align}
}%

To establish convergence properties of terms in the double sums in \eqref{eq:s_zy_part_2a} and \eqref{eq:s_zy_part_2b}, we must establish stochastic equicontinuity of the collection of stochastic processes $\{ \hat{\Pi}_{n}(\beta, s, r) \}$ indexed by $(s, r) \in \mathcal{S} \times \mathcal{S}$. Because $\mathcal{S} \times \mathcal{S}$ is compact and $\Pi(\beta, s; r)$ is continuous in $(s, r)$, then we can show that $\{ \hat{\Pi}_{n}(\beta, s, r) \}$ by showing that $ \hat{\Pi}_{n}(\beta, s, r)$ converges uniformly in probability to $\Pi(\beta, s; r)$ (Lemma \ref{lemm:uniform_convergence_implies_stochastic_equicontinuity}). We can use Lemma \ref{lemm:uniform_convergence_sum} to show the necessary uniform convergence result.

By Lemma \ref{lemm:nice_loss_and_pi}, we have that $\tilde{I}$ satisfies the conditions of Lemma \ref{lemm:uniform_convergence_sum}. Thus, we can apply Lemma \ref{lemm:uniform_convergence_sum} to establish uniform convergence in probability of $\hat{\Pi}_{n}(\beta, s, r)$ with respect to $(s, r).$ As a consequence, the collection of stochastic processes $\{\hat{\Pi}_{n}(\beta, s, r)\}$ is stochastically equicontinuous. In particular, $\hat{\Pi}_{n}(\beta, s, r)$ is stochastically equicontinuous at $(s(\beta, b), s(\beta, b))$, where $s(\beta, b)$ is the unique fixed point of $q(P(\beta, s, b))$ (see Lemma \ref{lemm:transformed_distribution_continuity_fp}). By Lemma \ref{lemm:quantile_perturbed_stoch_process}, we have that
\begin{align*}
    \hat{S}^{t_{n}}_{b, n} &\xrightarrow{p} s(\beta, b).
\end{align*}
Now, we can apply Lemma \ref{lemm:stochastic_equicontinuity_convergence_in_prob} to establish convergence in probability for the terms in \eqref{eq:s_zy_part_2a} and \eqref{eq:s_zy_part_2b}. As an example, for a perturbation $\zeta \in \{-1, 1\}^{d}$ with $j$-th entry equal to $1$ and arbitrary $\xi \in \{-1, 1\}$, Lemma \ref{lemm:stochastic_equicontinuity_convergence_in_prob} gives that
\begin{align*}
    &\hat{\Pi}_{n_{\zeta, \xi}}(\beta + b \mathbf{e}_{j} + b \cdot z(\zeta), \hat{S}^{t_{n}}_{b, n} + b\xi, \hat{S}^{t_{n}}_{b, n} )\\
    &\xrightarrow{p} \hat{\Pi}_{n_{\zeta, \xi}}(\beta + b \mathbf{e}_{j} + b \cdot z(\zeta), s(\beta, b) +b\xi, s(\beta, b)),
\end{align*}
and by the Weak Law of Large Numbers, we have that
\begin{align*}
    &\hat{\Pi}_{n_{\zeta, \xi}}(\beta + b \mathbf{e}_{j} + b\cdot z(\zeta), s(\beta, b)+b\xi, s(\beta, b))\\
    &\xrightarrow{p} \Pi(\beta + b\mathbf{e}_{j} + b\cdot z(\zeta), s(\beta, b)+b\xi, s(\beta, b)).
\end{align*}
Analogous results for the remaining terms in \eqref{eq:s_zy_part_2a} and \eqref{eq:s_zy_part_2b}. Also,
\[     \frac{n_{\zeta, \xi}}{n} \xrightarrow{p} \frac{1}{2^{d+1}}, \quad \zeta \in \{-1, 1\}^{d}, \xi \in \{-1, 1\}.\]  
By Slutsky's Theorem, when any $j$ and $b$ fixed, we have
{
\begin{align}
    &\mathbf{s}_{zy, j}^{n} \\
    &\xrightarrow{p} \sum_{\substack{\zeta \in \{-1, 1\}^{d} \text{ s.t. } \zeta_{j} = 1 \\ \xi \in \{-1, 1\}}} \frac{\Pi(\beta + b \mathbf{e}_{j} + b \cdot z(\zeta),s(\beta, b)+b\xi, s(\beta, b)) }{2^{d+1} \cdot b} \label{eq:r_b_pi_1}\\
    &- \sum_{\substack{\zeta \in \{-1, 1\}^{d} \text{ s.t. } \zeta_{j} = -1 \\ \xi \in \{-1, 1\}}} \frac{\Pi(\beta - b \mathbf{e}_{j} + b \cdot z(\zeta),s(\beta, b)+b\xi, s(\beta, b)) }{2^{d+1} \cdot b}. \label{eq:r_b_pi_2}
\end{align}
}%

Let $R_{b}$ denote the expression on the right side of the above equation. If there is a sequence $\{b_{n}\}$ such that $b_{n} \rightarrow 0$, then by Lemma \ref{lemm:transformed_distribution_continuity_fp},  $s(\beta, b_{n}) \rightarrow s(\beta)$, where $s(\beta)$ is the unique fixed point of $q(P(\beta, s)).$ By continuity of $\Pi$,
\[R_{b_{n}} \rightarrow \frac{\partial \Pi}{\partial \beta_{j}}(\beta, s(\beta); s(\beta)).\]

Using the definition of convergence in probability, we show that there exists such a sequence $\{b_{n}\}.$ From \eqref{eq:r_b_pi_1} and \eqref{eq:r_b_pi_2}, we have that for each $\epsilon, \delta > 0$ and $b> 0$ and sufficiently small, there exists $n(\epsilon, \delta, b)$ such that for $n \geq n(\epsilon, \delta, b)$
\[ P( | \mathbf{s}_{zy, j}^{n} - R_{b}| \leq \epsilon) \geq  1-\delta.\]
So, we can fix $\delta >0.$ For $k= 1, 2, \dots,$ let $N(k) = n(\frac{1}{k}, \delta, \frac{1}{k})$. Then, we can define a sequence such that $b_{n} = \epsilon_{n} = \frac{1}{k}$ for all $N(k) \leq n \leq N(k+1).$ So, we have that $\epsilon_{n} \rightarrow 0$ and $b_{n} \rightarrow 0.$ Thus, we have that 
\[ \mathbf{s}_{zy, j}^{n} \xrightarrow{p} \frac{\partial \Pi}{\partial \beta_{j}}(\beta, s(\beta); s(\beta)).\]
Considering all indices $j$,
\[ \mathbf{s}_{zy}^{n} \xrightarrow{p} \frac{\partial \Pi}{\partial \beta}(\beta, s(\beta); s(\beta)).\]
It remains to establish convergence in probability for $\mathbf{S}_{zz}.$ We have that
\begin{align*}
    \mathbf{S}_{zz}^{n} &= \frac{1}{b_{n}^{2}n}\mathbf{M}_{\beta}^{T}\mathbf{M}_{\beta} \\
    &= \frac{1}{b_{n}^{2} n} \sum_{i=1}^{n} (b_{n}\zeta_{i})^{T} (b_{n}\zeta_{i}).\\
    &=\frac{1}{n}\sum_{i=1}^{n} \zeta_{i}^{T} \zeta_{i}.
\end{align*}
We note that 
\[\EE[\zeta_{i} \sim R^{d}]{\zeta_{i, j} \zeta_{i, k}} = \begin{cases} 1 \text{ if } j=k \\ 0 \text{ if } j \neq k \end{cases}\] because $\zeta_{i}$ is a vector of independent Rademacher random variables. So, $\EE{\zeta_{i}^{T}\zeta_{i}} = \mathbf{I}_{d}.$ By the Weak Law of Large Numbers, we have that 
\[ \mathbf{S}_{zz}^{n} \xrightarrow{p} \mathbf{I}_{d}.\]
Finally, we can use Slutsky's Theorem to show that
\begin{align*}
&\hat{\Gamma}^{b_{n}, n}_{\Pi, \beta}(\beta, \hat{S}^{t_{n}}_{b_{n}, n}; \hat{S}^{t_{n}}_{b_{n}, n}) \\
&= (\mathbf{S}_{zz}^{n})^{-1}\mathbf{s}_{zy}^{n} \\
&\xrightarrow{p} (\mathbf{I}_{d})^{-1}\frac{\partial \Pi}{\partial \beta}(\beta, s(\beta); s(\beta)) \\
&= \frac{\partial \Pi}{\partial \beta}(\beta, s(\beta); s(\beta)).\end{align*}

\end{subsection}

\begin{subsection}{Proof of Corollary \ref{coro:dpi_ds}}
\label{subsec:dpi_ds}
The proof of this result is analogous to Lemma \ref{lemm:dpi_dbeta}. 
\end{subsection}

\begin{subsection}{Proof of Lemma \ref{lemm:density_converges}}
\label{subsec:density_converges}
We study the convergence of the kernel density estimate $p^{n}(\beta, \hat{S}^{t_{n}}_{n}, b_{n})(\hat{S}^{t_{n}}_{b_{n}, n}).$ Let $p^{n}(\beta, s, b)(r)$ is a kernel density estimate of density of $P(\beta, s, b)$ at a point $r$. Let $\beta_{i} = \beta + b\zeta_{i},$ $s_{i} = s + b\xi_{i}$, where $\zeta \sim R^{d}$ and $\xi \sim R$. We can write the explicit form of $p^{n}(\beta, s, b)(r)$ as follows
\[ p^{n}(\beta, s, b)(r) = \frac{1}{h_{n}} \sum_{i=1}^{n} k \Big(\frac{r - \beta_{i}^{T}X_{i}(\beta_{i}, s_{i}) + b\xi_{i} }{h_{n}}\Big).\]

For sufficiently small $b$, we can apply Lemma \ref{lemm:transformed_distribution} to map the unobservables $\nu_{i}=(Z_{i}, c_{i}) \sim F$ and cost functions $c_{i}$ to types $\nu_{i';b, \zeta, \xi} = (Z_{i';b, \zeta, \xi}, c_{i'}) \sim \tilde{F}_{b}$, so that when the agent with unobservables $\nu_{i';b, \zeta, \xi}$ best responds to the unperturbed model and threshold, they obtain the same raw score (without noise) as the agent with unobservables $\nu$ who responds to a perturbed model and threshold. So, we can write
\begin{align*}
&p^{n}(\beta, s, b)(r) \\
&= \frac{1}{h_{n}} \sum_{i=1}^{n} k \Big(\frac{r - \beta^{T}X_{i}(\beta, s) }{h_{n}}\Big) \\
&= \frac{1}{h_{n}} \sum_{i=1}^{n} k \Big(\frac{r - \beta^{T}x_{i}^{*}(\beta, s_{i}) - \beta^{T}\epsilon_{i} }{h_{n}}\Big).
\end{align*}

We make the following abbreviations.
{
\begin{align*}
    \tilde{k}(\nu_{i';b, \zeta, \xi}, \epsilon_{i}, \beta, s, r; h) &:= k \Big(\frac{r - \beta^{T}X_{i';b, \zeta, \xi}(\beta, s)}{h}\Big) \\
    \tilde{k}_{n}(\beta, s, r; h) &:= \frac{1}{n} \sum_{i=1}^{n} \tilde{k}(\nu_{i';b, \zeta, \xi}, \epsilon_{i}, \beta, s, r, ; h)\\
    K(\beta, s, r; h) &:= \EE[\tilde{F}_{b}]{ \tilde{k}(\nu_{i';b, \zeta, \xi}, \epsilon_{i}, \beta, s, r; h)}.
\end{align*}
}

We can write \[p^{n}(\beta, \hat{S}^{t_{n}}_{b, n}, b)(\hat{S}^{t_{n}}_{b, n}) = \frac{1}{h_{n}} \tilde{k}_{n}(\beta, \hat{S}^{t_{n}}_{b, n}, \hat{S}^{t_{n}}_{b, n}; h_{n}).\]

Due to the density estimate's dependence on the stochastic process $\hat{S}^{t_{n}}_{b, n}$, we first must establish the stochastic equicontinuity of the collection of stochastic processes $\{\tilde{k}_{n}(\beta, s, r)\}$ indexed by $(s, r) \in \mathcal{S} \times \mathcal{S}$. We show stochastic equicontinuity via uniform convergence in probability (Lemma \ref{lemm:uniform_convergence_sum}). The remainder of the proof follows by the Weak Law of Large Numbers and taking standard limits.

We fix $h_{n} = h$. Since $\tilde{k}$ satisfies the conditions of Lemma \ref{lemm:uniform_convergence_sum}, we can apply Lemma  \ref{lemm:uniform_convergence_sum} to realize that $\tilde{k}_{n}(\beta, s, r; h)$ converges uniformly in probability to $K(\beta, s, r; h)$ with respect to $(s, r) \in \mathcal{S} \times \mathcal{S}.$ As a result, the collection of stochastic processes $\{\tilde{k}_{n}(\beta, s, r; h)\}$ indexed by $(s, r) \in \mathcal{S} \times \mathcal{S}$ are stochastically equicontinuous. In particular, $\{\tilde{k}_{n}(\beta, s, r; h)\}$ is stochastically equicontinuous at $(s(\beta, b), s(\beta, b)).$
By Lemma \ref{lemm:quantile_perturbed_stoch_process}, we have that 
\[ \hat{S}^{t_{n}}_{b, n} \xrightarrow{p} s(\beta, b),\]
where $s(\beta, b)$ is the unique fixed point of $q(P(\beta, s, b)).$
we can apply Lemma \ref{lemm:stochastic_equicontinuity_convergence_in_prob} to see that
{\small \[ k_{n}(\beta, \hat{S}^{t_{n}}_{b, n}, \hat{S}^{t_{n}}_{b, n}; h) - k_{n}(\beta, s(\beta, b), s(\beta, b); h) \xrightarrow{p} 0.\]}%
Furthermore, by the Weak Law of Large Numbers, we have that
{\small
\[ k_{n}(\beta, s(\beta, b), s(\beta, b); h) \xrightarrow{p} K(\beta, s(\beta, b),s(\beta, b); h).\]
}%
Given our definition of the kernel function $k$ and for fixed $h$, we have that 
{
\begin{align*}
    &p^{n}(\beta, \hat{S}^{t_{n}}_{b, n}, b)(\hat{S}^{t_{n}}_{b, n}) \\
    &\xrightarrow{p} \frac{K(\beta, s(\beta, b), s(\beta, b); h)}{h}\\
    &= \frac{P(\beta, s(\beta, b), b)(s(\beta, b) + \frac{h}{2}) - P(\beta, s(\beta, b), b)(s(\beta, b) - \frac{h}{2}) }{h}.
\end{align*}
}%
Given that our sequence $h_{n} \rightarrow 0$ and $nh_{n} \rightarrow \infty$ and $k$ satisfies the assumptions of Theorem \ref{theo:parzen_kernel_density}, we can apply Theorem \ref{theo:parzen_kernel_density} to see that for each fixed $b$, we obtain a consistent density estimate.
{
\begin{align} 
&p^{n}(\beta, \hat{S}^{t_{n}}_{b, n}, b)(\hat{S}^{t_{n}}_{b, n}) \\
&\xrightarrow{p} \lim_{h_{n} \rightarrow 0} \frac{P(\beta, s(\beta, b), b)(s(\beta, b) + \frac{h_{n}}{2}) -  P(\beta, s(\beta, b), b)(s(\beta, b) - \frac{h_{n}}{2}) }{h_{n}} \\
&= p(\beta, s(\beta, b), b)(s(\beta, b)). \label{eq:r_b}
\end{align}
}%
Let $R_{b}$ denote the right side of the above equation. Suppose there exists a sequence such that $b_{n} \rightarrow 0$. By Lemma \ref{lemm:transformed_distribution_continuity_fp}, this gives us that $s(\beta, b_{n}) \rightarrow s(\beta)$, where $s(\beta)$ is the unique fixed point of $q(P(\beta, s))$. We can show that $R_{b_{n}} \rightarrow p(\beta, s(\beta))(s(\beta))$ as follows.
{
\begin{align*}
    &|R_{b_{n}} - p(\beta, s(\beta))(s(\beta)) | \\
    \leq& | p(\beta, s(\beta, b_{n}), b_{n})(s(\beta, b_{n})) - p(\beta, s(\beta, b_{n}))(s(\beta, b_{n}))| \\
    &+ |p(\beta, s(\beta, b_{n}))(s(\beta, b_{n})) - p(\beta, s(\beta))(s(\beta))| \\
    \leq& \sup_{s, r \in \mathcal{S}} |p(\beta, s, b_{n})(r) - p(\beta, s)(r)| \\
    &+ |p(\beta, s(\beta, b_{n}))(s(\beta, b_{n})) - p(\beta, s(\beta))(s(\beta))|.
\end{align*}
}%
Since $p(\beta, s)(r)$ is continuous in $s$ and $r$ (Lemma \ref{lemm:cdf}), there exists $N$ such that for $n \geq N$, the second term is less than $\epsilon.$ To bound the first term, we require the following lemma.
\begin{lemm}
\label{lemm:uniform_convergence_perturbed_density}
Under Assumptions \ref{assumption:strong_convexity}, 
\ref{assumption:finite_types},
\ref{assumption:br_interior},  
and \ref{assumption:etas_in_interior}, if $\sigma^{2} > \frac{2}{ \alpha_{*}(F)\cdot \sqrt{2\pi e}},$ then $p(\beta, s, b)(r) \rightarrow p(\beta, s)(r)$ uniformly in $s$ and in $r$ as $b \rightarrow 0.$
\hyperref[subsec:uniform_convergence_perturbed_density]{Proof in Appendix \ref{subsec:uniform_convergence_perturbed_density}}.
\end{lemm}
Due to the uniform convergence result, we have that if there exists a sequence $\{b_{n}\}$ such that $b_{n} \rightarrow 0$, then
\[ R_{b_{n}} \rightarrow p(\beta, s(\beta))(s(\beta)).\]
It remains to show that there exists such a sequence $\{b_{n}\}$ where $b_{n} \rightarrow 0.$ Using the definition of convergence in probability, we show that there exists such a sequence $\{b_{n}\}.$ From \eqref{eq:r_b}, we have that for each $\epsilon, \delta > 0$ and $b> 0$ and sufficiently small, there exists $n(\epsilon, \delta, b)$ such that for $n \geq n(\epsilon, \delta, b)$
\[ P( | p^{n}(\beta, \hat{S}_{b_{n}, n}^{t_{n}}, b_{n})(\hat{S}_{b_{n}, n}^{t_{n}}) - R_{b}| \leq \epsilon) \geq  1-\delta.\]

So, we can fix $\delta >0.$ For $k= 1, 2, \dots,$ let $N(k) = n(\frac{1}{k}, \delta, \frac{1}{k})$. Then, we can define a sequence such that $b_{n} = \epsilon_{n} = \frac{1}{k}$ for all $N(k) \leq n \leq N(k+1).$ So, we have that $\epsilon_{n} \rightarrow 0$ and $b_{n} \rightarrow 0.$ Finally, this gives that  
\[ p^{n}(\beta, \hat{S}^{t_{n}}_{n}, b_{n})(\hat{S}^{t_{n}}_{n}) \xrightarrow{p} p(\beta, s(\beta))(s(\beta)).\]

\end{subsection}

\begin{subsection}{Proof of Lemma \ref{lemm:ds_dbeta} }
\label{subsec:ds_dbeta}
First, we note that $s(\beta)$ is continuously differentiable in $\beta$ by Theorem \ref{theo:equilibrium}, so we can use implicit differentiation to compute the following expression for $\frac{\partial s}{\partial \beta}$
\begin{equation} 
\label{eq:ds_dbeta}
\frac{\partial s}{\partial \beta} = \frac{1}{1 - \frac{\partial q(P(\beta, s(\beta)))}{\partial s} } \cdot \frac{\partial q(P(\beta, s(\beta))) }{\partial \beta}.\end{equation}
After that, we apply the lemma below to express the partial derivatives of the quantile mapping $q(P(\beta, s))$ in terms of partial derivatives of the complementary CDF $\Pi(\beta, s; r)$.

\begin{lemm}
\label{lemm:dq_dtheta}
Let $\beta \in \mathcal{B}$, $s \in \mathcal{S}$. Under Assumption \ref{assumption:strong_convexity},  \ref{assumption:finite_types},
\ref{assumption:br_interior}, if $\sigma^{2} > \frac{2}{ \alpha_{*} \cdot \sqrt{2\pi e}}$ then for $\beta^{t}, s^{t}$ sufficiently close to $\beta, s,$ the derivative of $q(P(\beta, s))$  with respect to a one-dimensional parameter $\theta$ is given by 
\[ \frac{\partial q(P(\beta, s))}{\partial \theta} =  \frac{1}{p(\beta^{t}, s^{t})(r^{t})} \cdot \frac{\partial \Pi}{\partial \theta}(\beta, s; r^{t}),\]
where $r^{t} = q(P(\beta^{t}, s^{t})).$
\hyperref[subsec:dq_dtheta]{Proof in Appendix \ref{subsec:dq_dtheta}}.
\end{lemm}

Since $s(\beta)$ is the fixed point induced by $\beta$, we have that
\[ s(\beta) - q(P(\beta, s(\beta))) = 0.\]
From Theorem \ref{theo:equilibrium}, we have that $s(\beta)$ is continuously differentiable in $\beta$. Differentiating both sides of the above equation with respect to $\beta$ yields
{\small
\[ \frac{\partial s}{\partial \beta} - \Big(\frac{\partial q(P(\beta, s(\beta)))}{\partial \beta}  + \frac{\partial q(P(\beta, s(\beta)))}{\partial s} \cdot \frac{\partial s}{\partial \beta}\Big) = 0.\]
}%
Rearranging the above equation yields \eqref{eq:ds_db}, which shows that $\frac{\partial s}{\partial \beta}$ in terms of $\frac{\partial q(P(\beta, s))}{\partial s}$ and $\frac{\partial q(P(\beta, s))}{\partial \beta}.$ From Lemma \ref{lemm:dq_dtheta}, we have that for $\beta^{t}, s^{t}$ sufficiently close to $\beta, s$, we have that
{
\begin{align*}
    \frac{\partial q(P(\beta, s))}{\partial s} &= \frac{1}{p(\beta^{t}, s^{t})(r^{t})} \cdot \frac{\partial \Pi}{\partial s}(\beta, s; r^{t}) \\
    \frac{\partial q(P(\beta, s))}{\partial \beta} &= \frac{1}{p(\beta^{t}, s^{t})(r^{t})} \cdot \frac{\partial \Pi}{\partial \beta}(\beta, s; r^{t}), \\
\end{align*}
}%
where $r^{t} = q(P(\beta^{t}, s^{t})).$ Let $s(\beta) = s(\beta)$. Suppose that we aim to estimate the derivative when the model parameters are $\beta$ and the threshold is $s(\beta)$. If we consider $\beta^{t} = \beta$, $s^{t} = s(\beta)$, then $r^{t} = s(\beta).$ So, we have that
{
\begin{align}
\label{eq:quantile_deriv1}
    \frac{\partial q(P(\beta, s(\beta)))}{\partial s} &= -\frac{1}{p(\beta, s(\beta))(s(\beta))} \cdot \frac{\partial \Pi}{\partial s}(\beta, s(\beta); s(\beta)) \\
\label{eq:quantile_deriv2}
    \frac{\partial q(P(\beta, s(\beta)))}{\partial \beta} &= -\frac{1}{p(\beta, s(\beta))(s(\beta))} \cdot \frac{\partial \Pi}{\partial \beta}(\beta, s(\beta); s(\beta)). 
\end{align}
}%
Substituting \eqref{eq:quantile_deriv1} and \eqref{eq:quantile_deriv2} into \eqref{eq:ds_dbeta} yields
{
\begin{align*}
    \frac{\partial s}{\partial \beta}  
    &= \frac{1}{p(\beta, s(\beta))(s(\beta)) - \frac{\partial \Pi}{\partial s}(\beta, s(\beta); s(\beta))} \cdot \frac{\partial \Pi}{\partial \beta}(\beta, s(\beta); s(\beta)).\\
\end{align*}
}%

\end{subsection}

\begin{subsection}{Proof of Lemma \ref{lemm:apply_sherman}}
\label{subsec:apply_sherman}
$\bfH$ is positive definite and invertible, so we can apply the Sherman-Morrison Formula (Theorem \ref{theo:sherman}) to $\\ {(\bfH+ \phi_{\sigma}'(s - \beta^{T} x) \beta \beta^{T})^{-1}}$: let $\mathbf{A} = \bfH,$ $\mathbf{u} = \phi_{\sigma}'(s - \beta^{T} x) \beta,$ and $\mathbf{v} = \beta.$

\begin{align*}
    &(\bfH + \phi_{\sigma}'(s - \beta^{T} x) \beta \beta^{T})^{-1} \\
    &= \bfH^{-1} - \frac{\bfH^{-1} (\phi_{\sigma}'(s - \beta^{T} x) \beta) \beta^{T} \bfH^{-1}}{1 + \beta^{T}\bfH^{-1} (\phi_{\sigma}'(s - \beta^{T} x)\beta)}  \\
    &= \bfH^{-1} - \frac{\phi_{\sigma}'(s -\beta^{T}x) \bfH^{-1} \beta \beta^{T} \bfH^{-1} }{1 + \phi_{\sigma}'(s - \beta^{T} x) \beta^{T} \bfH^{-1} \beta }. \\
\end{align*}
 
\end{subsection}

\begin{subsection}{Proof of Lemma \ref{lemm:br_fixed_point}}
\label{subsec:br_fixed_point}
In the first part of the proof, we establish existence of a fixed point of $\omega_{i}(s; \beta)$. In the second part of the proof, we show that if a fixed point exists, then it must be unique.

First, we use the Intermediate Value Theorem (IVT) to show existence of a fixed point. We apply the IVT to the function $h_{i}(s)= s - \omega_{i}(s; \beta)$. We note that by Lemma \ref{prop:br_all} that $\omega_{i}(s; \beta)$ is continuous. It remains to show that there exists $s_{l}$ such that $h_{i}(s_{1}) < 0$ and there exists $s_{2}$ such that $s_{2}> s_{1}$ and $h_{i}(s_{2}) > 0$.  Then, by the Intermediate Value Theorem, there must be $s \in [s_{1}, s_{2}]$ for which $h_{i}(s) = 0$, which gives that $\omega_{i}(s; \beta)$ has at least one fixed point.

Let $\delta > 0.$ By Lemma \ref{lemm:limits_br}, we have that there exists $S_{l, 1}$ so that for all $s \leq S_{l}$, we have that 
\[ |\beta^{T}x_{i}^{*}(\beta, s) - \beta^{T}Z_{i} | < \delta.\]
Let $S_{l, 2} = \beta^{T}Z_{i} - \delta$. Let $s_{1} < \min(S_{l, 1}, S_{l, 2}).$ Then we have that
\begin{align*}
    h_{i}(s_{1}) &= s_{1} - \beta^{T}x_{i}^{*}(\beta, s_{1}) \\
    &\leq s_{1} - \beta^{T}Z_{i} + \delta \\
    &< (\beta^{T}Z_{i} - \delta)  - \beta^{T}Z_{i} + \delta.\\
    &< 0.\\
\end{align*}
Second, by Lemma \ref{lemm:limits_br}, we have that there exists $S_{h, 1}$ so that for all $s \geq S_{l}$, we have that 
\[ |\beta^{T}x_{i}^{*}(\beta, s) - \beta^{T}Z_{i} | < \delta.\]
Let $S_{h, 2} = \beta^{T}Z_{i} + \delta$. Let $s_{2} > \max(S_{h, 1}, S_{h, 2}).$ Then we have that
\begin{align*}
    h_{i}(s_{2}) &= s_{2} - \beta^{T}x_{i}^{*}(\beta, s_{2}) \\
    &\geq s_{2} - \beta^{T}Z_{i} - \delta \\
    &> 0.
\end{align*}
We have that $s_{1} < \beta^{T}Z_{i} - \delta < \beta^{T}Z_{i} + \delta < s_{2}.$ By the IVT, there must be some  $s \in [s_{1}, s_{2}]$  so that $h_{i}(s)=0.$
Second, we show that if a fixed point exists, then the fixed point must be unique. By Lemma \ref{lemm:score_gradient}, $h_{i}(s)$ is strictly increasing in $s$. There can be only one point at which $h_{i}(s) = 0.$ So, there is only one $s$ such that $s - \omega_{i}(s; \beta) = 0$. Thus, $\omega_{i}(s; \beta)$ has a unique fixed point.
\end{subsection}

\begin{subsection}{Proof of Lemma \ref{lemm:find_equivalent_agent}}
\label{subsec:find_equivalent_agent}
Since $y$ is in the interior of $\mathcal{X}$, then there exists some $\epsilon > 0$ such that the open ball of radius $\epsilon$ about $y$ is a subset of $\mathcal{X}.$ We note that
\begin{align*}
    |y' - y| &= \Big|  \beta \cdot (b \zeta^{T} x - b\xi) \Big| \\
    &\leq ||\beta|| \cdot  \Big| b \zeta^{T} x - b\xi \Big| \\
    &\leq | b \zeta^{T} x - b\xi| \\
    &\leq b |\zeta^{T} x - \xi |\\
    &\leq b (|\zeta||x| + |\xi|) \\
    &\leq b (\sqrt{d} \cdot  \sup_{x \in \mathcal{X}} |x| + 1)
\end{align*}
Since $\mathcal{X}$ is compact, we can say that the supremum in the above equation is achieved on $\mathcal{X}$ and we can call its value $m$. So, if $b < \frac{\epsilon}{(m\sqrt{d} + 1)}$, then $y' \in \text{Int}(\mathcal{X}).$
\end{subsection}

\begin{subsection}{Proof of Lemma \ref{lemm:uniform_convergence_perturbed_cdf}}
\label{subsec:uniform_convergence_perturbed_cdf}
We first show that $P(\beta, s, b)(r) \rightarrow P(\beta, s)(r)$ uniformly in $r$ as $b \rightarrow 0$. We aim to apply Lemma \ref{lemm:uniform_convergence_monotonic}. First, note that the continuity of $P(\beta, s)$ in $r$ is given by Lemma \ref{lemm:cdf}. We recall that 
{
\begin{align*}
&P(\beta, s, b)(r) \\
&= \frac{1}{2^{d+1}} \sum_{\substack{\zeta \in \{-1, 1\}^{d} \\\xi \in \{-1, 1\} }} \int \Phi_{\sigma}\Big(r - (\beta + b\zeta)^{T}x_{i}^{*}(\beta + b\zeta ,s + b\xi\Big) dF.\end{align*}
}%
$\Phi_{\sigma}$ is strictly increasing, so $P(\beta, s, b)(r)$ is strictly increasing because the sum of strictly increasing functions is also strictly increasing. By continuity of $x$ in $\beta$ and $s$ (Proposition \ref{prop:br_all}), we have that $P(\beta, s, b)(r) \rightarrow P(\beta, s)(r)$ pointwise in $r$. By Lemma \ref{lemm:uniform_convergence_monotonic}, as $b \rightarrow 0$, we have that 
\[ \sup_{r \in \mathcal{S}} |P(\beta, s, b)(r) - P(\beta, s)(r)| \rightarrow 0.\]
\end{subsection}

\begin{subsection}{Proof of Lemma \ref{lemm:uniform_convergence_perturbed_density}}
\label{subsec:uniform_convergence_perturbed_density}
We show that $p(\beta, s, b)(r) \rightarrow p(\beta, s)(r)$ uniformly in $s$ and $r$ as $b \rightarrow 0.$ We prove the claim in two steps. First, we rewrite $p(\beta, s)(r)$ and $p(\beta, s, b)(r)$ as a finite sum of terms that align by type and perturbation. Second, we can show uniform convergence for pairs of terms in the sums, which gives that the aggregate quantity $p(\beta, s, b)(r) \rightarrow p(\beta, s)(r)$ uniformly.

First, we rewrite $p(\beta, s)$ as follows
{
\begin{align}
&p(\beta, s)(r) \\
&= \int \phi_{\sigma}(r - \beta^{T}x_{i}^{*}(\beta, s))dF \\
&=  \sum_{\nu \in \text{supp}(F_{Z, c})} \sum_{\substack{\zeta \in \{-1, 1\}^{d} \\ \xi \in \{-1, 1\}}} \phi_{\sigma}(r - \beta^{T}x_{i}^{*}(\beta, s)) \frac{f(\nu)}{2^{d+1}} \label{eq:regular_score_pdf}.
\end{align}
}%
To rewrite $p(\beta, s, b)$, recall that for sufficiently small $b$, we can use Lemma \ref{lemm:transformed_distribution} to express $P(\beta, s, b)$ as the score distribution induced by agents with unobservables $(Z_{i'; b, \zeta, \xi}, c_{i'}) \sim \tilde{F}_{b}$ who best respond to a model $\beta$ and threshold $s$. The type and cost function is given by  transformation $T$ from Lemma \ref{lemm:find_equivalent_agent}. Recall that $T(i; b, \zeta, \xi)$ maps an agent $i$ with unobservables $\nu =(Z_{i}, c_{i}) \sim F$ to an agent $i'$ with unobservables $i'$ with unobservables $\nu_{i'; b, \zeta, \xi}=(Z_{i'; b, \zeta, \xi}, c_{i'})$. Since our assumed conditions also transfer to $\tilde{F}_{b},$ we have that that $P(\beta, s, b)(r)$ is continuously differentiable in $r$ with density $p(\beta, s, b)$ (Lemma \ref{lemm:cdf}).

Using the function $T_{1}$, we have that
{
\begin{align}
&p(\beta, s, b)(r) \\
&= \int \phi_{\sigma}(r - \beta^{T}x_{i'; b, \zeta, \xi}^{*}(\beta, s))d\tilde{F} \\
&= \sum_{\nu_{i'; b, \zeta, \xi} \in \text{supp}(\tilde{F}_{b})} \phi_{\sigma}(r - \beta^{T}x_{i'; b, \zeta, \xi}^{*}(\beta, s)) \cdot \tilde{f}_{b}(\nu_{i'; b, \zeta, \xi}) \\
&= \sum_{\nu \in \text{supp}(F)} \sum_{\substack{\zeta \in \{-1, 1\}^{d} \\ \xi \in \{-1, 1\}}}  \phi_{\sigma}(r - \beta^{T}x_{i'; b, \zeta, \xi}^{*}(\beta, s)) \cdot  \tilde{f}_{b}(\nu_{i'; b, \zeta, \xi}) \\
&=\sum_{\nu \in \text{supp}(F)} \sum_{\substack{\zeta \in \{-1, 1\}^{d} \\ \xi \in \{-1, 1\}}}  \phi_{\sigma}(r - \beta^{T}x_{i'; b, \zeta, \xi}^{*}(\beta, s)) \cdot \frac{f(\nu)}{2^{d+1}}. \label{eq:last_line}
\end{align}
}%
The last line follows from Lemma \ref{lemm:transformed_distribution}. Therefore, the terms of $p(\beta, s)$ in \eqref{eq:regular_score_pdf} align with the terms of $p(\beta, s, b)$ in \eqref{eq:last_line} by unobservable and perturbation. We have that 
{
\begin{align*}
    & |p(\beta, s, b)(r) - p(\beta, s)(r)| \\
    &= \Big| \sum_{\nu \in \text{supp}(F)} \sum_{\substack{\zeta \in \{-1, 1\}^{d} \\ \xi \in \{-1, 1\}}} ( \phi_{\sigma}(r - \beta^{T}x_{i'; b, \zeta, \xi}^{*}(\beta, s)) - \phi_{\sigma}(r - \beta^{T}x_{i}^{*}(\beta, s))   \cdot \frac{f(\nu)}{2^{d+1}} \Big| \\
    &\leq \sum_{\substack{\nu \in \text{supp}(F) \\ \zeta \in \{-1, 1\}^{d} \\ \xi \in \{-1, 1\}}} \Big| \phi_{\sigma}(r - \beta^{T}x_{i'; b, \zeta, \xi}^{*}(\beta, s)) - \phi_{\sigma}(r - \beta^{T}x_{i}^{*}(\beta, s)) \Big| \cdot \frac{f(\nu)}{2^{d+1}}.
\end{align*}
}%

Since the sum in the above inequality is finite, we can show $p(\beta, s, b)(r) \rightarrow p(\beta, s)(r)$ uniformly in $s, r$ if we can show that for every unobservable $\nu$ and perturbation $(\zeta, \xi)$, we have that
{
\[ \sup_{(s, r) \in \mathcal{S} \times \mathcal{S}} | \phi_{\sigma}(r - \beta^{T}x_{i'; b, \zeta, \xi}^{*}(\beta, s)) - \phi_{\sigma}(r - \beta^{T}x_{i}^{*}(\beta, s)) | \rightarrow 0.\]
}%

Now, we can use the following lemma to show uniform convergence (in $s$ and $r$) of the arguments to $G'$ in the above expression.
\begin{lemm}
\label{lemm:uniform_convergence_perturbed_score}
Suppose Assumption \ref{assumption:strong_convexity} and \ref{assumption:etas_in_interior} hold. Let agent $i$ be an agent with unobservables $\nu_{i} \sim F$ and cost function $c_{i}$. Let $T(i; b, \zeta, \xi)$ yield an agent $i'$ with unobservables $\nu_{i'; b, \zeta, \xi}$ and cost function $c_{i'}$, as defined in Lemma \ref{lemm:find_equivalent_agent_main} for any $\zeta \in \{-1, 1\}^{d}, \xi \in \{-1, 1\}$, and $b >0$ and sufficiently small. As $b \rightarrow 0$, $\beta^{T}x_{i'; b, \zeta, \xi}^{*}(\beta, s) \rightarrow \beta^{T}x_{i}^{*}(\beta, s)$ uniformly in $s$.
\hyperref[subsec:uniform_convergence_perturbed_score]{Proof in Appendix \ref{subsec:uniform_convergence_perturbed_score}}.
\end{lemm}

We observe that
\begin{align*}
     &\sup_{(s, r) \in \mathcal{S} \times \mathcal{S}} | (r- \beta^{T}x_{i'; b, \zeta, \xi}^{*}(\beta, s)) - (r - \beta^{T}x_{i}^{*}(\beta, s)| \\
     &=\sup_{s \in \mathcal{S} } |\beta^{T}x_{i'; b, \zeta, \xi}^{*}(\beta, s) - \beta^{T}x_{i}^{*}(\beta, s)| \\
     &\rightarrow 0
\end{align*}
where the uniform convergence in the last line follows from Lemma \ref{lemm:uniform_convergence_perturbed_score}. Since the argument to $\phi_{\sigma}$ in \eqref{eq:last_line} converges uniformly in $s$ and $r$, the argument to $\phi_{\sigma}$ is uniformly bounded. So, we can restrict the domain of $\phi_{\sigma}$ to an closed interval on which it is uniformly continuous. As a result, we also have that
{
\[ \sup_{s \in \mathcal{S}} |\phi_{\sigma}(r-\beta^{T}x_{i'; b, \zeta, \xi}^{*}(\beta, s) ) - \phi_{\sigma}(r-\beta^{T}x_{i}^{*}(\beta, s))| \rightarrow 0, \]
}%
which concludes the proof.
\end{subsection}

\begin{subsection}{Proof of Lemma \ref{lemm:dq_dtheta}}
\label{subsec:dq_dtheta}
For simplicity, we can write
\[
    r^{t} = q(P(\beta^{t}, s^{t})) = P(\beta^{t}, s^{t})^{-1}(q),\]
    and \[ r = q(P(\beta, s)) = P(\beta, s)^{-1}(q). \]
From Theorem \ref{theo:equilibrium}, we have the $q(P(\beta, s))$ is continuous in $\beta, s$. In addition,  we note that the density of the scores $p(\beta, s)(y)$ is continuous with respect to $\beta, s, y$ (Lemma \ref{lemm:cdf}). By the continuity of the density of the scores and the quantile mapping, we can choose $ \beta^{t}, s^{t}$ sufficiently close to $\beta, s$ such that $|r -  r^{t}| < \epsilon$ and $|p(\beta, s)(r^{t}) - p(\beta^{t}, s^{t})(r^{t}) | < \epsilon.$

From Lemma \ref{lemm:cdf}, we have that $P(\beta, s)$ and $P(\beta^{t}, s^{t})$ have unique inverses. So, the quantile mapping is uniquely defined, which means
\[
    P(\beta^{t}, s^{t})(r^{t}) = q, \quad 
    P(\beta, s)(r) = q. 
\]

As a result, we have that $P(\beta^{t}, s^{t})(r^{t}) = P(\beta, s)(r).$ Without loss of generality, suppose that $r > r^{t},$
{
\begin{align*}
    &P(\beta^{t}, s^{t})(r^{t}) - P(\beta, s)(r^{t}) \\
    &= P(\beta, s)(r) - P(\beta, s)(r) \\
    &= \int_{-\infty}^{r} p(\beta, s)(y) dy - \int_{-\infty}^{r^{t}} p(\beta, s)(y) dy \\
    &= \int_{r^{t}}^{r} p(\beta, s)(y) dy \\
    &= (r - r^{t}) p(\beta, s)(r^{t}) + o(|r^{t} - r|) \\
    &= (r - r^{t}) p(\beta^{t}, s^{t})(r^{t}) \\
    &\indent + o((r - r^{t}) | p(\beta, s) (r^{t}) - p(\beta^{t}, s^{t})(r^{t})|) + o(|r-r^{t} |) \\
    &= (q(P(\beta, s)) - q(P(\beta^{t}, s^{t}))) p(\beta^{t}, s^{t})(r^{t}) + o(\epsilon^{2}) + o(\epsilon) \\
\end{align*}
}%

We can differentiate both sides of the above equation with respect to a one-dimensional parameter $\theta$.
\[ - \frac{\partial P(\beta, s)(r^{t})}{\partial \theta} = \frac{\partial q(P(\beta, s))}{\partial \theta} \cdot p(\beta^{t}, s^{t})(r^{t}).\]
Since $\Pi$ is the complementary cdf of the score distribution, we observe that
\[\frac{\partial P(\beta, s)(r)}{\partial \theta} = -\frac{\partial \Pi}{\partial \theta}(\beta, s; r).\]
Solving for $\frac{\partial q(P(\beta, s))}{\partial \theta}$, we find that
\[ \frac{\partial q(P(\beta, s))}{\partial \theta} = \frac{1}{p(\beta^{t}, s^{t})(r^{t})} \cdot \frac{\partial \Pi}{\partial \theta}(\beta, s; r^{t}).\]

\end{subsection}

\begin{subsection}{Proof of Lemma \ref{lemm:uniform_convergence_perturbed_score}}
\label{subsec:uniform_convergence_perturbed_score}
Consider $b$ sufficiently small so that the transformation in Lemma \ref{lemm:find_equivalent_agent_main} is possible. Let $Z_{i';b, \zeta, \xi}$ be as defined in \eqref{eq:eta_prime}. Let $h_{b}: \mathcal{S} \rightarrow \mathbb{R},$ where $h_{b}(s) := s - \beta^{T}x_{i'; b, \zeta, \xi}^{*}(\beta, s)$ and $h(s) := s - \beta^{T}x_{i}^{*}(\beta, s)$. It is sufficient to show that $h_{b}(s) \rightarrow h(s)$ uniformly in $s$ because
\begin{align*}
    &\sup_{s \in \mathcal{S}}|\beta^{T}x_{i'; b, \zeta, \xi}^{*}(\beta, s) - \beta^{T}x_{i}^{*}(\beta, s)|\\
     &= \sup_{s \in \mathcal{S}}|s -\beta^{T}x_{i'; b, \zeta, \xi}^{*}(\beta, s) - s + \beta^{T}x_{i}^{*}(\beta, s)| \\
    &= \sup_{s \in \mathcal{S}}| h_{b}(s) - h(s)|.
\end{align*}
We aim to apply Lemma \ref{lemm:uniform_convergence_monotonic} to show $h_{b} \rightarrow h$ uniformly. We have that $\mathcal{S}$ compact. By Proposition \ref{prop:br_all}, we have that $h(s)$ is continuous. By Lemma \ref{lemm:score_gradient}, we have that each $h_{b}$ strictly increasing in $s$. In addition, we have the following pointwise convergence
{
\begin{align}
    &\lim_{b \rightarrow 0} h_{b}(s) \\
    &= \lim_{b \rightarrow 0} s - \beta^{T}x_{i'; b, \zeta, \xi}^{*}(\beta, s) \\
    &= \lim_{b \rightarrow 0}s - \beta^{T}\Big(x_{i}^{*}(\beta + b\zeta, s + b\xi)  + b \cdot \beta(\zeta^{T}x_{i}^{*}(\beta + b\zeta, s + b\xi) - \xi)\Big) \label{eq:score2} \\
    &=\lim_{b \rightarrow 0}s - \beta^{T}x_{i}^{*}(\beta + b\zeta, s + b\xi)  - b \cdot (\zeta^{T}x_{i}^{*}(\beta + b\zeta, s + b\xi) - \xi) \\
    &=s - \beta^{T}x_{i}^{*}(\beta, s) \label{eq:score3} \\
    &=h(s).
\end{align}
}%
\eqref{eq:score2} follows from Lemma \ref{lemm:find_equivalent_agent_main}, which gives an explicit expression for $x_{i';b, \zeta, \xi}^{*}(\beta, s)$.  \eqref{eq:score3} follows from continuity of the best response mapping in $\beta, s$ (Proposition \ref{prop:br_all}). Thus, $h_{b} \rightarrow h$ uniformly, so we have that $\beta^{T}x_{i'; b, \zeta, \xi}^{*}(\beta, s) \rightarrow \beta^{T}x_{i}^{*}(\beta, s)$ uniformly in $s$.
\end{subsection}

\end{section}

\end{document}